\documentclass[10pt,twocolumn,letterpaper]{article}

\usepackage[pagenumbers]{cvpr} %
\usepackage[accsupp]{axessibility}  %

\usepackage{graphicx}
\usepackage{amsmath}
\usepackage{amssymb}
\usepackage{booktabs}
\usepackage{tikz}
\usepackage{bbm}
\usepackage{bm}
\usepackage{pifont}
\usepackage{cuted}
\usepackage{colortbl}
\newcommand{\cmark}{\ding{51}}%

\DeclareMathOperator*{\argmin}{arg\,min}

\usepackage{multirow}
\usepackage[labelfont=bf,font=small,skip=3pt]{caption}

\usepackage[pagebackref,breaklinks,colorlinks,citecolor=blue]{hyperref}

\usepackage[capitalize]{cleveref}
\crefname{section}{Sec.}{Secs.}
\Crefname{section}{Section}{Sections}
\Crefname{table}{Table}{Tables}
\crefname{table}{Tab.}{Tabs.}

\def\cvprPaperID{8140} %
\def\confName{CVPR}
\def\confYear{2023}

\begin{document}
\setlength{\abovedisplayskip}{3pt}
\setlength{\belowdisplayskip}{3pt}
\setlength{\abovedisplayshortskip}{3pt}
\setlength{\belowdisplayshortskip}{3pt}

\newcommand{\tree}{\bm{\Gamma}}
\newcommand{\shenlong}[1]{\textcolor{magenta}{#1}}
\newcommand{\todocite}[1]{\textcolor{red}{\textit{Citation needed []}}}
\newcommand{\shenlongsay}[1]{\textcolor{blue}{[Shenlong: #1]}}
\newcommand{\sg}[1]{\textcolor{blue}{[Saurabh: #1]}}
\newcommand{\todo}[1]{\textcolor{red}{\textit{TODO: #1}}}
\newcommand{\shaowei}[1]{\textcolor{magenta}{[Shaowei: #1]}}

\newcommand{\imgtile}[2]{
    {\tikz{
    \node[draw=black, draw opacity=1.0, line width=.3mm, fill opacity=0.7,fill=white, inner sep=0pt](gt) at (0, 0) {\includegraphics[width=#2\linewidth]{#1}};
    \node[draw=black, draw opacity=1.0, line width=.3mm, fill opacity=0.7,fill=white, inner sep=0pt](gt) at (-0.1, 0.1) {\includegraphics[width=#2\linewidth]{#1}};
    \node[draw=black, draw opacity=1.0, line width=.3mm, fill opacity=0.7,fill=white, inner sep=0pt](gt) at (-0.2, 0.2) {\includegraphics[width=#2\linewidth]{#1}};
    \node[draw=black, draw opacity=1.0, line width=.3mm, fill opacity=0.7,fill=white, inner sep=0pt](gt) at (-0.3, 0.3) {\includegraphics[width=#2\linewidth]{#1}}; }}
}

\newcommand{\robotD}[0]{RoboArt\xspace}
\newcommand{\sapiensD}[0]{Sapiens\xspace}
\newcommand{\wim}[0]{WatchItMove\xspace}
\newcommand{\mbs}[0]{MultiBodySync\xspace}
\newcommand{\xpar}[1]{\noindent\textbf{#1}\ \ }
\newcommand{\vpar}[1]{\vspace{3mm}\noindent\textbf{#1}\ \ }

\newcommand{\sect}[1]{Section~\ref{#1}}
\newcommand{\sects}[1]{Sections~\ref{#1}}
\newcommand{\eqn}[1]{Equation~\ref{#1}}
\newcommand{\eqns}[1]{Equations~\ref{#1}}
\newcommand{\fig}[1]{Figure~\ref{#1}}
\newcommand{\figs}[1]{Figures~\ref{#1}}
\newcommand{\tab}[1]{Table~\ref{#1}}
\newcommand{\tabs}[1]{Tables~\ref{#1}}

\newcommand{\ignorethis}[1]{}
\newcommand{\norm}[1]{\lVert#1\rVert}
\newcommand{\fcseven}{$\mbox{fc}_7$}

\renewcommand*{\thefootnote}{\fnsymbol{footnote}}

\def\naive{na\"{\i}ve\xspace}
\def\Naive{Na\"{\i}ve\xspace}

\makeatletter
\DeclareRobustCommand\onedot{\futurelet\@let@token\@onedot}
\def\@onedot{\ifx\@let@token.\else.\null\fi\xspace}

\def\iid{\emph{i.i.d}\onedot}
\def\eg{\emph{e.g}\onedot} \def\Eg{\emph{E.g}\onedot}
\def\ie{\emph{i.e}\onedot} \def\Ie{\emph{I.e}\onedot}
\def\cf{\emph{c.f}\onedot} \def\Cf{\emph{C.f}\onedot}
\def\etc{\emph{etc}\onedot} \def\vs{\emph{vs}\onedot}
\def\wrt{w.r.t\onedot} \def\dof{d.o.f\onedot}
\def\etal{\emph{et al}\onedot}
\makeatother

\definecolor{citecolor}{RGB}{34,139,34}
\definecolor{mydarkblue}{rgb}{0,0.08,1}
\definecolor{mydarkgreen}{rgb}{0.02,0.6,0.02}
\definecolor{mydarkred}{rgb}{0.8,0.02,0.02}
\definecolor{mydarkorange}{rgb}{0.40,0.2,0.02}
\definecolor{mypurple}{RGB}{111,0,255}
\definecolor{myred}{rgb}{1.0,0.0,0.0}
\definecolor{mygold}{rgb}{0.75,0.6,0.12}
\definecolor{myblue}{rgb}{0,0.2,0.8}
\definecolor{mydarkgray}{rgb}{0.66,0.66,0.66}

\newcommand{\myparagraph}[1]{\vspace{-6pt}\paragraph{#1}}

\newcommand{\bbR}{{\mathbb{R}}}
\newcommand{\bK}{\mathbf{K}}
\newcommand{\bX}{\mathbf{X}}
\newcommand{\bY}{\mathbf{Y}}
\newcommand{\bk}{\mathbf{k}}
\newcommand{\bx}{\mathbf{x}}
\newcommand{\by}{\mathbf{y}}
\newcommand{\bhy}{\hat{\mathbf{y}}}
\newcommand{\bty}{\tilde{\mathbf{y}}}
\newcommand{\bG}{\mathbf{G}}
\newcommand{\bI}{\mathbf{I}}
\newcommand{\bg}{\mathbf{g}}
\newcommand{\bS}{\mathbf{S}}
\newcommand{\bs}{\mathbf{s}}
\newcommand{\bM}{\mathbf{M}}
\newcommand{\bw}{\mathbf{w}}
\newcommand{\eye}{\mathbf{I}}
\newcommand{\bU}{\mathbf{U}}
\newcommand{\bV}{\mathbf{V}}
\newcommand{\bW}{\mathbf{W}}
\newcommand{\bn}{\mathbf{n}}
\newcommand{\bv}{\mathbf{v}}
\newcommand{\bq}{\mathbf{q}}
\newcommand{\bR}{\mathbf{R}}
\newcommand{\bi}{\mathbf{i}}
\newcommand{\bj}{\mathbf{j}}
\newcommand{\bp}{\mathbf{p}}
\newcommand{\bt}{\mathbf{t}}
\newcommand{\bJ}{\mathbf{J}}
\newcommand{\bu}{\mathbf{u}}
\newcommand{\bB}{\mathbf{B}}
\newcommand{\bD}{\mathbf{D}}
\newcommand{\bz}{\mathbf{z}}
\newcommand{\bP}{\mathbf{P}}
\newcommand{\bC}{\mathbf{C}}
\newcommand{\bA}{\mathbf{A}}
\newcommand{\bZ}{\mathbf{Z}}
\newcommand{\bff}{\mathbf{f}}
\newcommand{\bF}{\mathbf{F}}
\newcommand{\bo}{\mathbf{o}}
\newcommand{\bc}{\mathbf{c}}
\newcommand{\bT}{\mathbf{T}}
\newcommand{\bQ}{\mathbf{Q}}
\newcommand{\bL}{\mathbf{L}}
\newcommand{\bl}{\mathbf{l}}
\newcommand{\ba}{\mathbf{a}}
\newcommand{\bE}{\mathbf{E}}
\newcommand{\bH}{\mathbf{H}}
\newcommand{\bd}{\mathbf{d}}
\newcommand{\br}{\mathbf{r}}
\newcommand{\bb}{\mathbf{b}}
\newcommand{\bh}{\mathbf{h}}

\newcommand{\btheta}{\bm{\theta}}

\newcommand{\bhh}{\hat{\mathbf{h}}}
\newcommand{\ci}{{\cal I}}
\newcommand{\ct}{{\cal T}}
\newcommand{\co}{{\cal O}}
\newcommand{\ck}{{\cal K}}
\newcommand{\cu}{{\cal U}}
\newcommand{\cS}{{\cal S}}
\newcommand{\cQ}{{\cal Q}}
\newcommand{\cT}{{\cal S}}
\newcommand{\cC}{{\cal C}}
\newcommand{\cE}{{\cal E}}
\newcommand{\cF}{{\cal F}}
\newcommand{\cL}{{\cal L}}
\newcommand{\X}{{\cal{X}}}
\newcommand{\Y}{{\cal Y}}
\newcommand{\cH}{{\cal H}}
\newcommand{\cP}{{\cal P}}
\newcommand{\cN}{{\cal N}}
\newcommand{\cU}{{\cal U}}
\newcommand{\cV}{{\cal V}}
\newcommand{\cX}{{\cal X}}
\newcommand{\cY}{{\cal Y}}
\newcommand{\graph}{{\cal H}}
\newcommand{\bayes}{{\cal B}}
\newcommand{\cx}{{\cal X}}
\newcommand{\cg}{{\cal G}}
\newcommand{\cm}{{\cal M}}
\newcommand{\cM}{{\cal M}}
\newcommand{\cG}{{\cal G}}
\newcommand{\cR}{\cal{R}}
\newcommand{\R}{\cal{R}}
\newcommand{\eig}{\mathrm{eig}}

\newcommand{\D}{{\cal D}}
\newcommand{\bfp}{{\bf p}}
\newcommand{\bfd}{{\bf d}}

\newcommand{\cv}{{\cal V}}
\newcommand{\ce}{{\cal E}}
\newcommand{\cy}{{\cal Y}}
\newcommand{\cz}{{\cal Z}}
\newcommand{\cb}{{\cal B}}
\newcommand{\cq}{{\cal Q}}
\newcommand{\cd}{{\cal D}}
\newcommand{\bcf}{{\cal F}}
\newcommand{\cI}{\mathcal{I}}

\newcommand{\ut}{^{(t)}}
\newcommand{\up}{^{(t-1)}}

\newcommand{\bpi}{\boldsymbol{\pi}}
\newcommand{\bphi}{\boldsymbol{\phi}}
\newcommand{\bPhi}{\boldsymbol{\Phi}}
\newcommand{\bmu}{\boldsymbol{\mu}}
\newcommand{\bSigma}{\boldsymbol{\Sigma}}
\newcommand{\bGamma}{\boldsymbol{\Gamma}}
\newcommand{\bbeta}{\boldsymbol{\beta}}
\newcommand{\bomega}{\boldsymbol{\omega}}
\newcommand{\blambda}{\boldsymbol{\lambda}}
\newcommand{\bkappa}{\boldsymbol{\kappa}}
\newcommand{\btau}{\boldsymbol{\tau}}
\newcommand{\balpha}{\boldsymbol{\alpha}}
\def\bgamma{\boldsymbol\gamma}

\newcommand{\prox}{{\mathrm{prox}}}

\newcommand{\pardev}[2]{\frac{\partial #1}{\partial #2}}
\newcommand{\dev}[2]{\frac{d #1}{d #2}}
\newcommand{\dw}{\delta\bw}
\newcommand{\lab}{\mathcal{L}}
\newcommand{\unlab}{\mathcal{U}}
\newcommand{\ind}{1{\hskip -2.5 pt}\hbox{I}}
\newcommand{\ff}[2]{   \cf_{\prec (#1 \rightarrow #2)}}
\newcommand{\vv}[2]{   \cv_{\prec (#1 \rightarrow #2)}}
\newcommand{\dd}[2]{   \delta_{#1 \rightarrow #2}}
\newcommand{\ld}[2]{   \lambda_{#1 \rightarrow #2}}
\newcommand{\en}[2]{  \bD(#1|| #2)}
\newcommand{\ex}[3]{  \bE_{#1 \sim #2}\left[ #3\right]} 
\newcommand{\exd}[2]{  \bE_{#1 }\left[ #2\right]}

\newcommand{\se}[1]{\mathfrak{se}(#1)}
\newcommand{\SE}[1]{\mathbb{SE}(#1)}
\newcommand{\so}[1]{\mathfrak{so}(#1)}
\newcommand{\SO}[1]{\mathbb{SO}(#1)}

\newcommand{\poselow}{\xi}
\newcommand{\pose}{\bm{\poselow}}
\newcommand{\linpose}{\pose^\ell}
\newcommand{\cbpose}{\pose^c}
\newcommand{\rateparam}{v_i}
\newcommand{\bapose}{\bm{\poselow}_i}
\newcommand{\trackingpose}{\bm{\poselow}}
\newcommand{\rotlow}{\omega}
\newcommand{\rot}{\bm{\rotlow}}
\newcommand{\translow}{v}
\newcommand{\trans}{\bm{\translow}}
\newcommand{\hnorm}[1]{\left\lVert#1\right\rVert_{\gamma}}
\newcommand{\lnorm}[1]{\left\lVert#1\right\rVert}
\newcommand{\barate}{v_i}
\newcommand{\trackingrate}{v}
\newcommand{\imgpt}{\mathbf{u}_{i,k,j}}
\newcommand{\mappt}{\mathbf{X}_{j}}
\newcommand{\timet}[1]{\bar{t}_{#1}}
\newcommand{\mf}[1]{\text{MF}_{#1}}
\newcommand{\kmf}[1]{\text{KMF}_{#1}}
\newcommand{\Exp}{\text{Exp}}
\newcommand{\Log}{\text{Log}}

\newcommand{\robotvis}[3]{
    \imgtile{figures/#1/input.png}{0.13} &
    \includegraphics[trim=#3, clip, width=#2\linewidth]{figures/#1/#1_pred_seg.png} &
    \includegraphics[trim=#3, clip, width=#2\linewidth]{figures/#1/pred_skeleton.png} &
    \includegraphics[trim=#3, clip, width=#2\linewidth]{figures/#1/pred_flow.png} &
    \includegraphics[trim=#3, clip, width=#2\linewidth]{figures/#1/#1_gt_seg.png} &
    \includegraphics[trim=#3, clip, width=#2\linewidth]{figures/#1/gt_skeleton.png} &
    \includegraphics[trim=#3, clip, width=#2\linewidth]{figures/#1/gt_flow.png} \\
}

\definecolor{lgray}{rgb}{0.7, 0.7, 0.7}
\definecolor{lgreen}{rgb}{0.75, 0.93, 0.686}
\definecolor{lred}{rgb}{0.93, 0.73, 0.686}

\newcommand{\reanimatecompare}[4]{
    \includegraphics[trim=#3, clip, width=#2\linewidth]{figures/#1/retarget_input_#4.png} & 
    \includegraphics[trim=#3, clip, width=#2\linewidth]{figures/#1/novel_#4_pred_base.png} & 
    \includegraphics[trim=#3, clip, width=#2\linewidth]{figures/#1/novel_#4_pred.png} & 
    \includegraphics[trim=#3, clip, width=#2\linewidth]{figures/#1/novel_#4_gt.png} \\
}

\newcommand{\reanimatevis}[4]{
    \includegraphics[trim=#3, clip, width=#2\linewidth]{figures/#1/retarget_input_#4.png} & 
    \includegraphics[trim=#3, clip, width=#2\linewidth]{figures/#1/novel_#4_pred.png} & 
    \includegraphics[trim=#3, clip, width=#2\linewidth]{figures/#1/novel_#4_gt.png} \\
}

\newcommand{\comparevis}[3]{
    \includegraphics[trim=#3, clip, width=#2\linewidth]{figures/comparison/#1/msync_#1_pred_seg.png} & 
    \includegraphics[trim=#3, clip, width=#2\linewidth]{figures/comparison/#1/wmove_#1_pred_seg.png} & 
    \includegraphics[trim=#3, clip, width=#2\linewidth]{figures/comparison/#1/#1_pred_seg.png} & 
    \includegraphics[trim=#3, clip, width=#2\linewidth]{figures/comparison/#1/#1_gt_seg.png} \\
    \includegraphics[trim=#3, clip, width=#2\linewidth]{figures/comparison/#1/msync_pred_skeleton.png} & 
    \includegraphics[trim=#3, clip, width=#2\linewidth]{figures/comparison/#1/wmove_pred_skeleton.png} & 
    \includegraphics[trim=#3, clip, width=#2\linewidth]{figures/comparison/#1/pred_skeleton.png} & 
    \includegraphics[trim=#3, clip, width=#2\linewidth]{figures/comparison/#1/gt_skeleton.png} \\
}

\newcommand{\sapienvis}[2]{
        \imgtile{figures/sapien/#1/#1_input_0.png}{0.25} &
        \includegraphics[trim=#2,clip,width=0.3\linewidth]{figures/sapien//#1/#1_pred_0.png} &
        \includegraphics[trim=#2,clip,width=0.3\linewidth]{figures/sapien/#1/#1_gt.png} 
}

\newcommand{\red}[1]{\textcolor{red}{#1}}
\newcommand{\blue}[1]{\textcolor{blue}{#1}}

\title{
Reanimating Arbitrary 3D Objects \\
Parsing Point Cloud Videos into Rearticulable 3D Models \\
Parsing Point Cloud Videos of Arbitrary Objects into Rearticulable 3D Models \\
Reart4D: Rearticulable Models from 4D Point Clouds \\
Care4D: Category Agnostic Re-articulable models from 4D Point Clouds \\
Re-Art4D: Rearticulable Models from 4D Point Clouds of Arbitrary Objects \\
}
\title{
Building Rearticulable Models for Arbitrary 3D Objects from 4D Point Clouds
}

\author{
Shaowei Liu$^1$\hspace{4mm}
Saurabh Gupta$^1${\thanks{equal advising, alphabetic order}}\hspace{4mm}
Shenlong Wang$^{1\ast}$\hspace{4mm} \\
$^1$University of Illinois Urbana-Champaign\\
{\normalsize \url{https://stevenlsw.github.io/reart/}}
}

\maketitle

\begin{strip}
\vspace{-1.7cm}
\centering
\includegraphics[trim={0 0 100 0}, clip, width=1.0\textwidth]{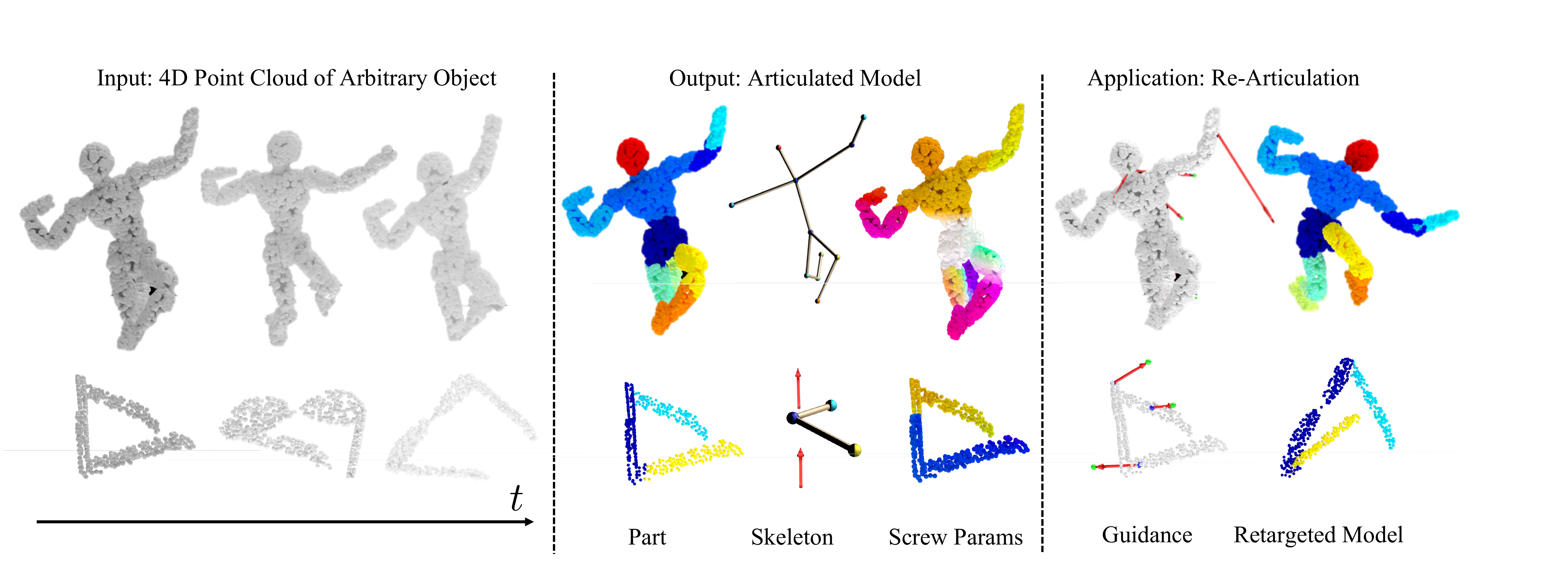}
\captionof{figure}{
Given a short point cloud sequence of arbitrary articulated object \textbf{(left)}, our method outputs an animatable 3D model \textbf{(middle)}, which can be retargeted to novel poses with only a few sparse point correspondences \textbf{(right)}.
}
\label{teaser}
\vspace{-.2cm}
\end{strip}

\begin{abstract}
    We build rearticulable models for arbitrary everyday man-made objects containing an arbitrary number of parts that are connected together in arbitrary ways via 1 degree-of-freedom joints. Given point cloud videos of such everyday objects, our method 
    identifies the distinct object parts, what parts are connected to what other parts, and the properties of the joints connecting each part pair.
    We do this by jointly optimizing the part segmentation, transformation, and kinematics using a novel energy minimization framework. Our inferred animatable models, enables retargeting to novel poses with sparse point correspondences guidance. We test our method on a new articulating robot dataset, and the Sapiens dataset with common daily objects, as well as real-world scans. Experiments show that our method outperforms two leading prior works on various metrics.
\end{abstract}

\begin{figure}
\centering
\includegraphics[trim=0cm 12cm 0cm 42cm, clip, width=1\linewidth]{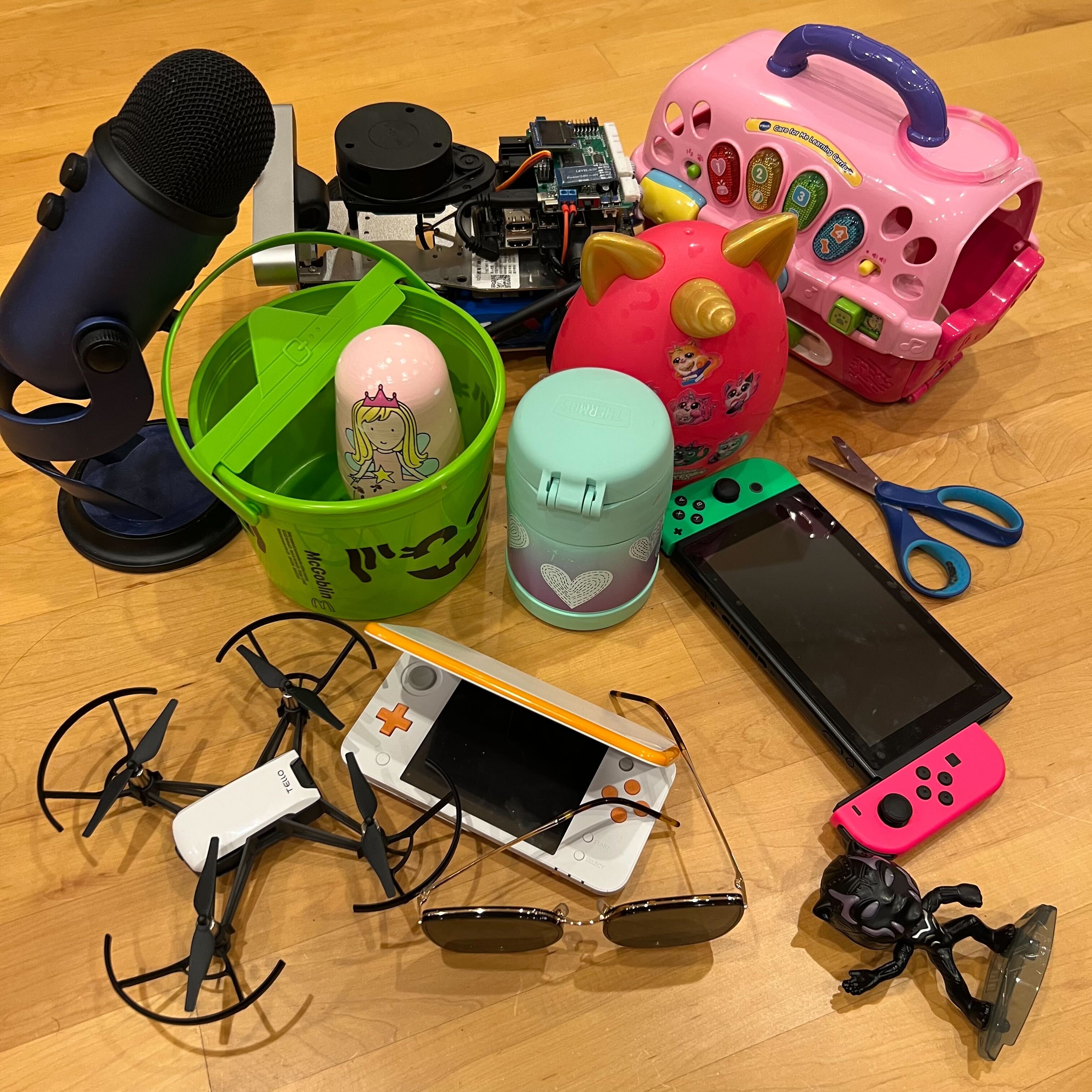}
\caption{Many man-made everyday objects can be explained with rigid parts connected in a kinematic tree with 1DOF joints.}
\vspace{-5mm}
\label{fig:objects}
\end{figure}

\section{Introduction}
\label{sec:intro}
Consider the sequence of points clouds observations of articulating everyday objects
shown in Figure~\ref{teaser}. As humans, we can readily infer the {\it kinematic
structure} of the underlying object, \ie the different object parts and their
connectivity and articulation relative with one
another~\cite{johansson1973visual}. This paper develops computational
techniques with similar abilities. Given point cloud videos of {\it arbitrary}
everyday objects (with an {\it arbitrary} number of parts) undergoing articulation, we
develop techniques to build animatable 3D reconstructions of the underlying
object by 
a) identifying the distinct object parts, 
b) inferring what parts are connected to what other parts, and 
c) the properties of the joint between each connected part pair.
Success at this task enables rearticulation of objects.
Given just a few user clicks specifying what point goes where, we can
fill in the remaining geometry as shown on the right side in Figure~\ref{teaser}.

Most past work on inferring how objects articulate tackles it in a {\it
category-specific} manner, be it for people~\cite{loper2015smpl, SMPL-X, romero2022embodied},
quadrupeds~\cite{Zuffi, wu2022CASA}, or even everyday objects~\cite{mu2021sdf}. 
Category-specific treatment
allows the use of specialized shape models (such as the SMPL model~\cite{loper2015smpl} for
people), or defines the output space (\eg location of 2 hinge joints for the
category eye-glasses). This limits applicability of such methods to
categories that have a canonical topology, leaving out categories with large
intra-class variation (\eg boxes that can have 1-4 hinge joints), 
or in-the-wild objects which may have an arbitrary number of parts
connected in arbitrary ways (\eg robots). 

\begin{table}
\setlength{\tabcolsep}{4pt}
\resizebox{\linewidth}{!}{
\begin{tabular}{lccc}
\toprule
                                            & \bf Arbitrary & \bf Realistic Joint & \bf Arbitrary  \\
                                            & \bf Parts     & \bf Constraints     & \bf Kinematics \\ \midrule
Category-specific \eg \\
\quad people~\cite{loper2015smpl}             & no        & yes             & no \\ 
\quad quadrupeds~\cite{Zuffi, wu2022CASA}     & no        & yes             & no \\ 
\quad cartoons~\cite{xu2020rignet}            & no        & yes             & no \\ \midrule

DeepPart~\cite{yi2018deep}                  & yes       & no              & no                   \\
NPP~\cite{hayden2020nonparametric}                                 & yes       & no              & no                   \\
ScrewNet~\cite{jain2021screwnet}            & yes       & yes             & no                   \\
UnsupMotion~\cite{sidi2011unsupervised}     & no        & yes             & no                   \\
Ditto~\cite{jiang2022ditto}                 & yes       & yes             & no                   \\
MultiBodySync~\cite{huang2021multibodysync} & yes       & no              & no                   \\
WatchItMove~\cite{noguchi2022watch}         & yes       & no              & yes                  \\
Ours                                        & yes       & yes             & yes    \\
\bottomrule
\end{tabular}}
\caption{Most past work on inferring rearticulable models is category specific. Building rearticulable models for arbitrary everyday man-made objects requires reasoning about arbitrary part geometries, arbitrary part connectivity, and realistic joint constraints (1DOF \wrt parent part). We situate past work along these 3 dimensions, and discuss major trends in \cref{sec:related}.}
\label{tab:related}
\vspace{-5mm}
\end{table}

Only a very few past works tackle the problem of inferring 
rearticulable models in a category-agnostic manner. Huang \etal~\cite{huang2021multibodysync} only infer part
segmentations, which by itself, is insufficient
for rearticulation. Jiang \etal~\cite{jiang2022ditto} only consider a single 1-DOF joint per object, dramatically restricting its application 
(think about a humanoid robot with four limbs, but the articulable model can only articulate one). Noguchi \etal~\cite{noguchi2022watch} present the
most complete solution but instead work with visual input and don't incorporate
the 1DOF constraint, \ie a part can only rotate or translate about a 
fixed axis on the parent part, common to a large number of man-made 
objects as can be seen in \cref{fig:objects}.
Inferring 3DOF / 6DOF joints leads to unrealistic rearticulation and is thus undesirable (consider the leg of eyeglasses can freely move or rotate). %
Our work fills this gap in the literature. Our method extracts 3D 
rearticulable models for arbitrary everyday objects (containing an 
arbitrary number of parts that are connected together in arbitrary 
ways via 1DOF joint) from point cloud videos. 
To the best of our knowledge, this is the first work to tackle this specific problem.

Our proposed method jointly reasons about part geometries and their
1-DOF inter-connectivity with one another. 
At the heart of our approach is
a novel continuous-discrete energy formulation that seeks to jointly learn 
parameters of the object model (\ie assignments of points in the canonical view
to parts, and the connectivity of parts to one another) by minimizing 
shape and motion reconstruction error 
(after appropriate articulation of the inferred model) 
on the given views. 
As it is difficult to directly optimize in the presence
of continuous and discrete variables with structured constraints,
we first estimate a {\it relaxed} model
that infers parts that are free to follow an arbitrary 6DOF trajectory over time
(\ie doesn't require parts to be connected in a kinematic tree with 1DOF joints). 
We {\it project} the estimated relaxed model to a kinematic model and continue
to optimize with the reconstruction error to further finetune the estimated 
joint parameters. 
Our joint approach leads to better models
and improved rearticulation as compared to adaptations of past 
methods~\cite{huang2021multibodysync, noguchi2022watch} to this task.

\section{Related Work}
\label{sec:related}
Building rearticulable models for arbitrary objects requires a) identifying the distinct parts, b) the kinematic topology that connects the different parts, and c) any constraints that there may be on the joints. This makes inferring articulable models from raw observation inputs challenging. Consequently, there are very few past works that tackle all these challenges to produce an end-to-end system for this task. A large body of work has focused on tackling subsets of these problems, or they operate in settings where some of these aspects are simplified. We overview past works with respect to these aspects in \cref{tab:related} and discuss major lines of work in more detail below.

A large body of work tackles this problem in a \textbf{category-specific} manner. This leads to a well specified definition of parts and their kinematic connectivity, and allows the use of sophisticated modeling. A good example is modeling for humans~\cite{loper2015smpl}, human hands~\cite{romero2022embodied, SMPL-X}, cartoon characters~\cite{xu2020rignet}, and quadrupeds~\cite{Zuffi, wu2022CASA, yang2022banmo}. However, doing this for each new category requires manual effort and this approach doesn't scale to arbitrary objects that we consider in our work. Many recent works have pursued extending such approaches to other categories making assumptions about the number of parts~\cite{mu2021sdf, abbatematteo2019learning, wei2022self}, kinematic topology~\cite{xue2021omad, desingh2019factored, abdul2022learning}, articulation type or common geometric features~\cite{weng2021captra, li2020category, heiden2022inferring}.
Researchers have also tackled intermediate problems that are useful for eventual rearticulation in a category-specific manner, \eg articulated pose estimation~\cite{li2020category, kumar2016spatiotemporal, daniele2020multiview}, motion prediction~\cite{sharf2014mobility, hu2017learning, jiang2022opd, li2016mobility}, and 3D reconstruction~\cite{mu2021sdf, kawana2021unsupervised, huang2012occlusion}. 

{\bf Category-agnostic modeling}, that is necessary to enable modelling of arbitrary objects, is considerably more challenging, and most past work only tackles subsets of the problem. A body of work focuses on {unsupervised part segmentation}, \ie parsing articulated objects into multiple rigid moving parts. Early work addressed the problem by clustering and co-segmentation~\cite{hu2012co, van2013co, sidi2011unsupervised, tzionas2016reconstructing, yuan2016space} by relying on motion and geometry, while more recent approaches~\cite{yi2018deep, chen2019bae, zeng2021visual, wang2019shape2motion, yan2020rpm, wang2018sgpn, huang2021multibodysync, liu2022autogpart} use more expressive features extracted using a neural network. Recent work has also tackled pose estimation for arbitrary 1DOF joints~\cite{jain2021screwnet, qian2022understanding, jiang2022ditto} or for given kinematic trees~\cite{liu2020nothing}. 

Closest to us are works from 
Huang \etal~\cite{huang2021multibodysync}, 
Jiang \etal~\cite{jiang2022ditto}, 
and Noguchi \etal~\cite{noguchi2022watch}.
Huang \etal~\cite{huang2021multibodysync} build upon part segmentation work from Yi \etal~\cite{yi2018deep} to output temporally-consistent part segmentation by solving a joint synchronization problem. This works very well for part segmentation. However, Huang \etal do not infer the kinematic tree connecting these parts together and parts can undergo 6DOF transformation relative to one another. Thus, their output, as is, falls short of the rearticulation goal.
Jiang \etal~\cite{jiang2022ditto} tackle rearticulation of generic objects and study the 1DOF nature of the joint. However, their formulation is limited to only infering a {\it single} 1DOF joint (\eg laptops), and is thus incapable of analysing more complex objects that our approach is able to handle. Lastly, 
Noguchi \etal~\cite{noguchi2022watch} fit articulated objects models to multi-view RGB video sequences and demonstrate impressive reanimation results. In contrast, we a) work with point-cloud input, b) model fine-grained shape for parts (as opposed to approximating them as ellipsoids), and c) additionally incorporate 1DOF constraint for joints as is common to everyday objects (as opposed to a general 3DOF spherical joints for~\cite{noguchi2022watch}).
To the best of our knowledge, our work is the first to simultaneously achieves all three desiderata for reanimation: inferring parts, kinematic connectivity, and joint constraints.

\begin{figure}
\includegraphics[trim={0cm 0cm 0cm 3cm}, clip, width=1.0\linewidth]{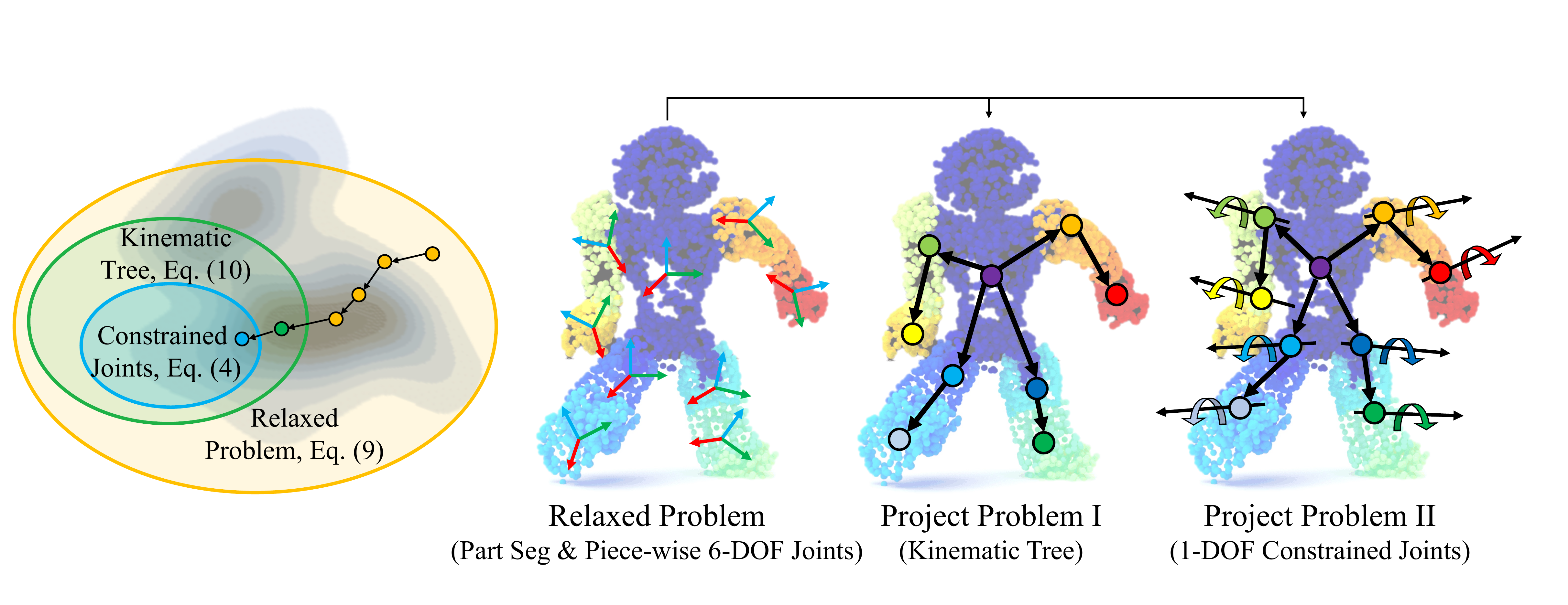}
\caption{\textbf{Overview:} We formulate the problem of rearticulable object modeling from 4D point cloud as an energy minimization problem. The optimization is divided into a {\it relaxation} stage that reasons about a 6-DOF piece-wise rigid model without kinematic constraints and a {\it project} stage that casts the solution to a valid kinematic tree (all joints satisfy 1-DoF constraints).}
\label{fig:overview}
\vspace{-3mm}
\end{figure}

\section{Approach}

Our goal is to infer the articulated structures, parts, and joint parameters jointly, given a point cloud video. Given as input a point cloud sequence $\mathcal{P} = \{ \bP^t\}_{t \in 1,...T}$, our goal is to produce an animatable kinematic model with part definitions, part connectivity, joint parameters, and joint states at each time step.

We first introduce the parameterized articulated model $\tree$ in \cref{sec:articulation}. We then cast estimation of this articulated model as an energy minimization problem (\cref{sec:energy}). It turns out that directly optimizing this energy is difficult, thus we design a two-stage {\it relax and project} approach to optimizing this energy (\cref{sec:inference}), as shown in \cref{fig:overview}. We first solve a more tractable relaxed problem (estimating a 6DOf piece-wise rigid model without any kinematic constraints, \cref{sec:relax}), project the solution to a valid kinematic tree (\cref{sec:project}) and further optimize the 1DOF joint parameters (\cref{sec:finetune}). 

\subsection{Articulated Model}
\label{sec:articulation}
\noindent \includegraphics[width=\linewidth]{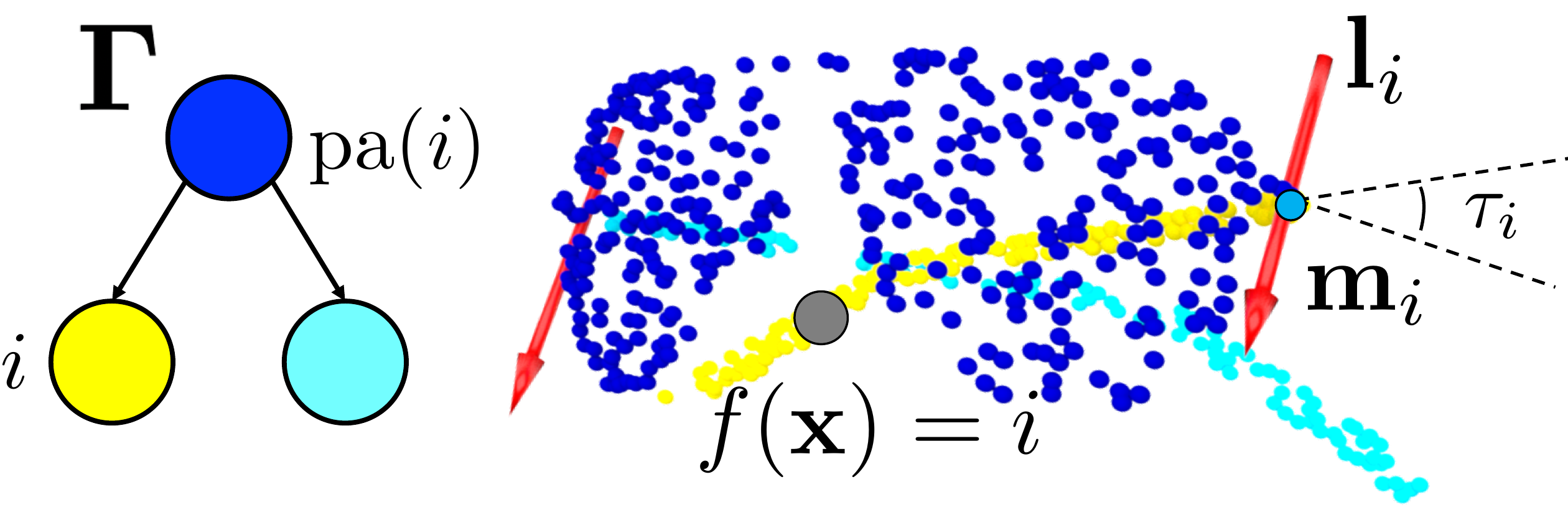}
Our kinematic model $\tree$ comprises of $n$ parts that are connected to one another. Each part $i \in [1\ldots n]$ (except the root joint) is connected to its parent part $\mathrm{pa}(i) \in \tree $ via a 1-DOF joint. 
Each joint, is either be a revolute joint or a prismatic joint parameterized by screw parameters~\cite{lynch2017modern}, $\bs_i = \{\bl_i, \mathbf{m}_i\}$, specifying where the joint is located ($\mathbf{m}_i$) and its axis ($\bl_i$, rotation axis for revolute joints or 
translation axis for prismatic joints) with respect to the parent part. 

The articulation is controlled by a set of joint state parameters $\btheta^t = \{\btheta^t_i\}_{i \in 1,\ldots n}$ 
where each joint parameter $\btheta^t_i = (\tau^t_i, d^t_i)$ represents the rotation $\tau$ or translation $d$ along its axis for joint $i$ at time step $t$. 

Parts are represented as a partition of Euclidean space, $\bbR^3$, at a {\it canonical} time step $c$, \ie a coordinate-based semantic field $f(\bx)$ that maps points $\bx \in \bbR^3$ to a part label in $[1\ldots n]$. We use a coordinate-based neural network to parameterize the part segmentation field.  
This completes the definition of the articulable model. 

Given a set of state parameters $\btheta^t$, we can deform a point cloud in canononical frame to the target pose through piece-wise rigid transform: 
\begin{equation}
\label{eq:articulation}
M(\btheta^t; \tree, f) = \{ \bT_{f(\bx)}^t \cdot \bx \}_{\bx \in \bP^c}
\end{equation}
where the rigid transformation of each part is computed through rigid pose composition along the kinematic chain from the part to root: 
\begin{equation}
\label{eq:kinematics}
\bT_i^t = \prod_{i^\prime \in \textrm{ans}(i)} \bT(\bs_{i^\prime}, \btheta_{i^\prime}^t)
\end{equation}
and $\mathrm{ans} = \{ i, \mathrm{pa}(i),\mathrm{pa}(\mathrm{pa}(i)),... \}$ denotes the ordered set of ancestor nodes of $i$, which forms a kinematic chain.  

Each joint's rigid transformation to its parent $\bT(\bs_i, \btheta_i^T)$ can be computed in closed-form using its screw parameters $(\bs_i, \btheta_i^T)$ and Rodriguez formula~\cite{dai2015euler}:
\begin{equation}
\label{eq:screw}
\bT(\bs, \btheta) = \Exp([\bomega, \bv]) = \Exp([ \tau \bl; \tau \mathbf{m} \times \bl + d\bl]^T)
\end{equation}
where $\bomega = \tau \bl \in \bbR^3$ is the Euler vector representing rotation; $\bv = \tau \bl \times \mathbf{m} + d \bl$ the translational vector; $\Exp: \mathbb{R}^6 \mapsto \SE{3}$ maps the minimal 6-DoF representation to an $\SE{3}$ rigid transform; $\times$ is the cross product between vectors. There exists two special cases. A joint is called {\bf prismatic joint} if its rotation angle is always zero, \ie $\tau = 0$. In this case, the rigid body can only slide along the axis $\bl$ without rotation (think about a drawer). 
A joint is called {\bf revolute joint} if translational component always equals to 0, \ie $d = 0$. In this case, the rigid body can only rotate along the rotation axis $\bl$ (think about a door);

\subsection{Energy Formulation}
\label{sec:energy}

Our approach takes as input a point cloud sequence $\mathcal{P} = \{ \bP^t\}_{t \in 1,...T}$. The output of our approach is the number of parts $n$, the part labeling
function $f$, a kinematic tree connecting the different parts $\tree$, and screw
parameters $\{\bs_i\}$ for each non-root joint. 

We estimate these parameters
using an analysis-by-synthesis approach and find model parameters $M$, such
that after appropriate per-time articulation $\btheta^t$, $M(\btheta^t; \tree, f)$ matches the
given point cloud $\bP^t$ well.
\begin{equation}
\label{eq:original}
\min_{n, f, \tree, \{\bs_i\}} \sum_t \min_{\btheta_t} E_\text{recons}
\left(M(\btheta^t; \tree, f), \bP^t\right),
\end{equation}
where $M(\btheta^t; \tree, f)$ denotes the canonical point cloud $\bP^c$ transformed using
the model parameters and the articulation $\bx^t$ at time $t$. 

$E_\text{recons}$ measures 
the compatibility between the point cloud articulated according to the inferred model and the observed point cloud at each step. Specifically, the energy consists of three terms:
\begin{equation}
\label{eq:recons}
E_\text{recons}
= \lambda_\text{CD} E_\text{CD} + \lambda_\text{EMD} E_\text{EMD}+ \lambda_\text{flow} E_\text{flow}
\end{equation}
\noindent \textbf{Chamfer distance term:}
$E_\text{CD}$ measures agreement between two point clouds using the Chamfer distance: 
\begin{equation}
E_\text{CD}(\mathbf{X},\mathbf{Y}) = %
\sum_{\mathbf{x} \in \mathbf{X}}\min_{\mathbf{y} }\|\mathbf{x}-\mathbf{y}\|_2^2 + %
\sum_{\mathbf{y} \in \mathbf{Y}}\min_{\mathbf{x}}\|\mathbf{x}-\mathbf{y}\|_2^2. 
\end{equation}

\noindent \textbf{Earth-mover distance term:} The earth-mover distance $E_\text{EMD}$ also measures geometric compatibility. It finds the optimal assignment between the two points sets by solving a bipartite matching problem~\cite{jonker1987shortest} and measure the residual loss:
\begin{equation}
    E_\text{EMD}(\mathbf{X},\mathbf{Y}) = \min_{\mathbf{S}}\|\mathbf{S}\mathbf{X} - \mathbf{Y}\|_F^2, 
\end{equation}
where $\mathbf{S}$ is a permutation matrix that represents the bipartite assignment. Compare against $E_\text{CD}$, $E_\text{EMD}$ could better capture details while $E_\text{CD}$ only focus on point coverage.
We compute the assignment via linear assignment solver~\cite{crouse2016implementing}.

\noindent \textbf{Flow energy term:} This term encourages the estimated 3D motion to be similar to observed point-wise 3D motion obtained from a scene flow network,
\begin{eqnarray}
   E_\text{flow} = \sum_\bx \| \bx^t - \bx^{t-1} - g( \bP^t, \bP^{t-1}; \bx^t) \|_2^2.
\end{eqnarray}
where $\bx^t - \bx^{t-1} = (\bT_{f(\bx)}^t  - \bT_{f(\bx)}^{t-1}) \cdot \bx$ is the predicted motion flow using the state for point $\bx$ at time $t$ and $t-1$.
$g(\bP^t, \bP^{t-1};\bx^t)$ is the observed 3D motion flow between two point clouds $\bP^t$ and $\bP^{t-1}$, at the query point location $\bx^t$. Since there is no guaranteed one-to-one mapping between $M(\btheta^t)$ and $\bP^t$, we use trilinear interpolation to estimate the inferred flow at an arbitrary point $\bx^t$:
\begin{align*}
    g(\bP^t, \bP^{t-1};\bx^t) = \frac{\sum_{\mathbf{x}_k^t \in \text{knn}(\mathbf{x}^t; \mathbf{P}^{t})} \|\mathbf{x}^t -\mathbf{x}_k^t \|^{-1}F(\mathbf{x}_k^t)}{\sum_{\mathbf{x}_k^t \in \text{knn}(\mathbf{x}^t; \mathbf{P}^{t})}\|\mathbf{x}^t -\mathbf{x}_k^t\|^{-1}},
\end{align*}
where $\mathbf{F}^{t}$ is the predicted flow between observation $\mathbf{P}^{t}$ and $\mathbf{P}^{t-1}$. Given the established flow $\mathbf{F}^{t}$ at locations in $\mathbf{P}^{t}$, $g(\cdot)$ takes query prediction $\mathbf{x}^t$ as input and returns an interpolated motion estimation at point $\bx^t$. $\text{knn}(\bx^t; \bP^t)$ retrieves the K-nearest neighbors of $\bx^t$ from $\bP^t$. 

\subsection{Inference via Relax and Project}
\label{sec:inference}
Directly optimizing over the set of model parameters as defined in \cref{eq:original} is difficult. It involves both discrete and continuous optimization variables as well as structured constraints such as the tree-structure of $\tree$, making it hard for both numerical approaches and combinatorial methods. 
Hence we pursue an alternate ``relax-and-project'' approach. 

\subsubsection{Fitting a Relaxed Model}
\label{sec:relax}
Our method first fits a {\it relaxed} model $\hat{M}$. 
This relaxed model doesn't constrain the parts to
form a kinematic tree and thus lets individual parts to follow their own
independent 6DOF trajectory over time. This relaxed model is parameterized via
the number of parts $n$ (same as for $M$), the part labeling function $f$ (also
same as for $M$). This model is articulated via a 6DOF pose for each part at
each time step $t$: $\hat{{\bT}}^t = \{\hat{{\bT}}^t_i\}_{i \in [1\ldots n]}$. 

We first optimize this relaxed model using the same cost
function as in \cref{eq:original} but evaluated for reposing under the
relaxed model at each time step $t$ via $\hat{\bT}^t$: 
\begin{equation}
\label{eq:relaxed}
\min_{n, f} \sum_t \min_{\bT^t} E_\text{recons}\left(\hat{M}(\bT^t),\bP^t\right),
\end{equation}
where $\hat{M(\bT^t)}$ denotes the canonical point cloud $\bP^c$ transformed using
the model parameters and the per-time step part transformations $\bT^t$ at time $t$. 

The energy function defined in \cref{eq:recons} involves a joint optimization over 6-DOF transformations $\hat{\bT}_i^t$ and a neural segmentation field $f$. For each point, the part segmentation field outputs a discrete-valued index $f(\bx)$, which is later used to query the corresponding rigid transform to warp the point cloud as defined in \cref{eq:articulation}. Optimizing through this discrete index $f(\bx)$ is hard. We leverage Gumbel-softmax trick~\cite{jang2017categorical} along with the straight-through gardient estimator~\cite{bengio2013estimating} to overcome this challenge.  This makes the energy function end-to-end differentiable \wrt $f$ and $\hat{\bT}^t$, allowing us to use gradient descent to minimize the energy. 

\subsubsection{Projecting to the Kinematic Model}
\label{sec:project}
The estimated relaxed model parameters are projected onto a valid kinematic model
by inducing a tree structured connectivity between the parts and projecting
absolute 6DOF transformations into child-parent screw parameters. This is done
using a cost function that measures the discrepancy between kinematic parameters ${\tree, \{\bs_i\}, \{\btheta^t\}}$ and relaxed 6DOF transformations $\{\hat{\bT^t}\}$:
\begin{equation}
\label{eq:project}
\min_{\tree, \{\bs_i\}, \{\btheta^t\}} E_\text{project}\left(
\left(\tree, \{\bs_i\}, \{\btheta^t\}\right), \{\hat{\bT^t}\}
\right).
\end{equation}
Given individual 6DOF part trajectories for all parts, we want to infer a
kinematic model that is as similar to the relaxed model while obeying the
kinematic constraints. The kinematic constraints consists of two aspects: the
tree structure (\ie the connectivity of parts with one another) and the 1DOF
constraint for the joint between each pair of connected parts. We tackle the
projection problem defined in \cref{eq:project} by minimizing the cost
over different trees topologies $\tree$ and associated screw parameters $\bs_i = \{\bl_i, \mathbf{m}_i\}$:
\begin{equation}
\label{eq:tree}
E_\text{project} = \lambda_\text{spatial}
E_\text{spatial} + 
\lambda_\text{1-DOF}
E_\text{1-DOF}.
\end{equation}
This loss function evaluates the total fitness of parent-child pairs in $\tree$.
$E_\text{spatial}$ measures the spatial proximity of parent-child pairs, $E_\text{spatial}(\tree) = \sum_i \min_{\bx \in \bp_i} \min_{\by \in \bp_{\text{pa}(i)}} \| \bx - \by \|_2^2$, where $\bp_i = \{ \bx \in \bP^c | f(\bx) = i\}$ is the set of points corresponding to part $i$. 
The $E_\text{1-DOF}$ term measures how close to a
1DOF motion is the motion of the child part relative to the parent part. 
$E_\text{1-DOF}$ is
computed as the error in approximating the temporal sequence of relative
transformation of the child part with respect to the parent part as a 1DOF transformation:
\begin{equation}
\label{eq:1-dof}
E_\text{1-DOF} = \sum_i \sum_t 
\text{trace}((\hat{\bT}_{pa(i)}^t \ominus \hat{\bT}_i^t ) \ominus \bT(\bs_i, \btheta_i^t)),
\end{equation}
where $\oplus$ is the rigid pose composition operator: $\bT_a \oplus \bT_b = \bT_b \cdot \bT_a$ and $\ominus$ is the inverse rigid pose composition operator $\bT_a \ominus \bT_b = \text{inv}(\bT_b) \cdot \bT_a$. 

Solving the projection problem defined in \cref{eq:tree} is also challenging: it's a joint optimization between discrete tree structure $\tree$ and continuous screw parameters $\bs_i, \btheta_i^t$; the desire for a valid tree topology $\tree$ introduces a complicated structured constraint. We observe that screw parameters $\bs_i, \btheta_i^t$ are only involved in exactly 1 1-DOF energy term under any valid tree. Thus, they can be independently optimized.   
For all part pairs $(i,j)$, we compute screw parameters from $\hat{\bT}_i$ and $\hat{\bT}_j$ under the assumption that $j = \text{pa}(i)$ using screw theory~\cite{stramigioli2001geometry}. This let us compute the 1DOF energy for all part pairs $(i,j)$. The spatial term can also be similarly computed for all part pairs. The minimization in \cref{eq:tree} then reduces to finding the minimum spanning tree which can be done efficiently~\cite{cormen2022introduction}. 

\noindent \textbf{Merging.} Before the projection, we iteratively merge parts that are spatially close and don't have relative motion. The latter falls out directly from \cref{eq:1-dof} if the approximation against the identity transform is below $\epsilon_{m}$.

\subsubsection{Final Fitting}
\label{sec:finetune}
We further finetune $\{\bs_i\}$, $\{\btheta_i^t\}$ over the original problem in \cref{eq:original} with gradient descent while keeping $f$ and $\tree$ fixed. This gives the final estimation of tree structure $\tree$, part segmentation $f$, screw parameters $\{\bs_i\}$, and joint states $\{\btheta_i^t\}$. 

\noindent \textbf{Canonical frame selection.} Given the non-convexity of the optimization, we run the optimization multiple times by choosing different time steps $c$ as {\it canonical} frame to build neural part field. We pick the best $c$ that has the lowest $E_\text{project}$ in \cref{eq:project}.
\section{Experiments}
\label{sec:exp}
We performing experiments on two 3D asset datasets and real-world setting, demonstrates our method could be applied on arbitrary articulated objects both qualitatively and quantitatively. 

\begin{table*}[]
    \centering
    \footnotesize
    \resizebox{\linewidth}{!}{
    \setlength{\tabcolsep}{0.2em} %
    \begin{tabular}{cccc|ccc}
        \cellcolor{lgray}{Input} & 
        \multicolumn{3}{c}{\cellcolor{lred}{Predictions}} &
        \multicolumn{3}{c}{\cellcolor{lgreen}{Ground Truth}} \\
        \robotvis{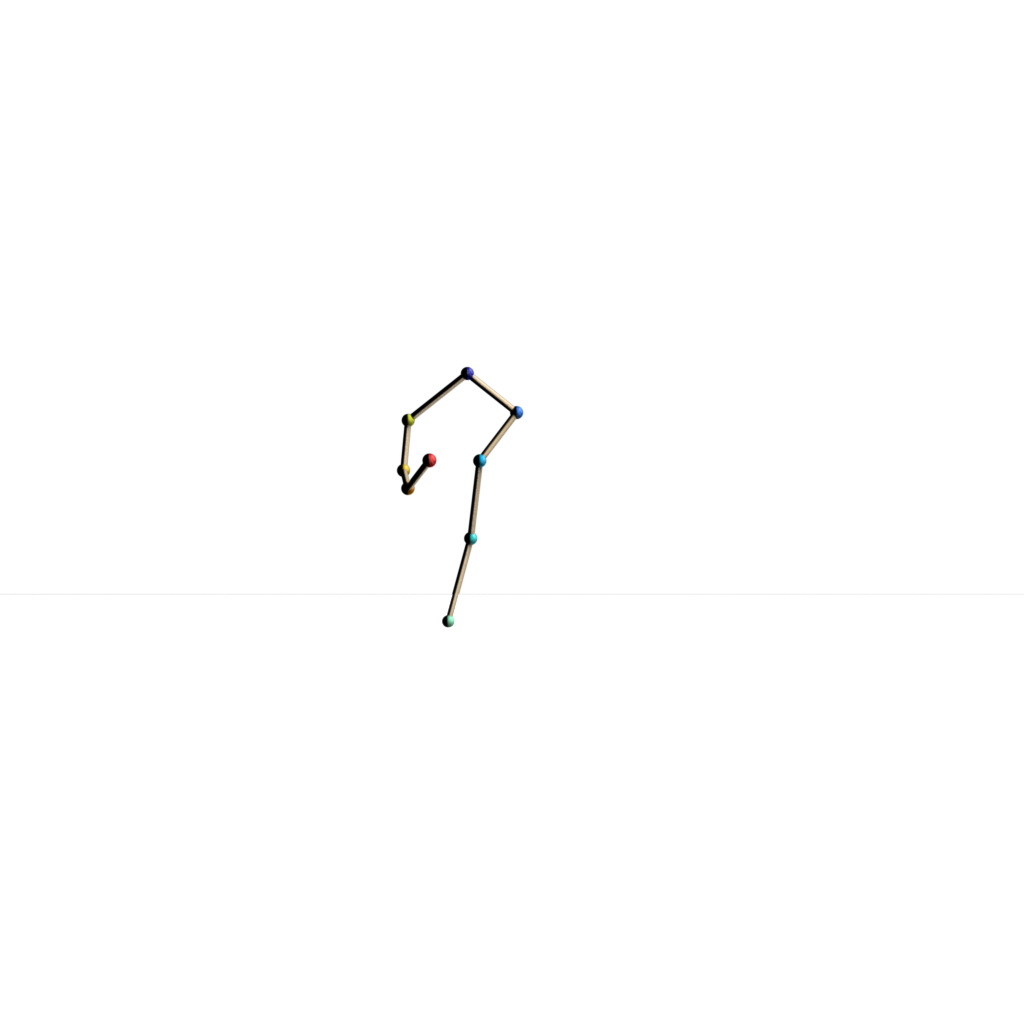}{0.15}{7cm 8cm 11cm 10cm}
        \robotvis{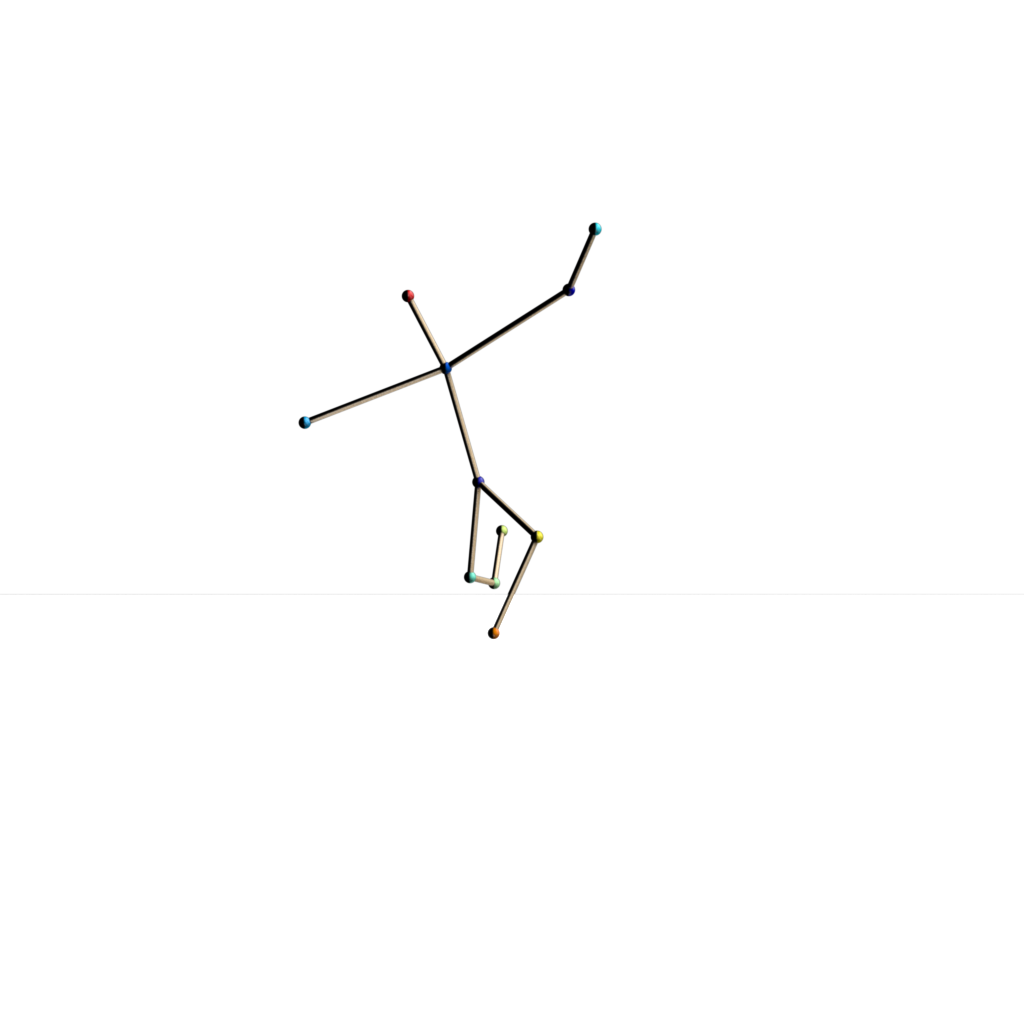}{0.15}{6.5cm 10cm 10.5cm 7cm}
        \robotvis{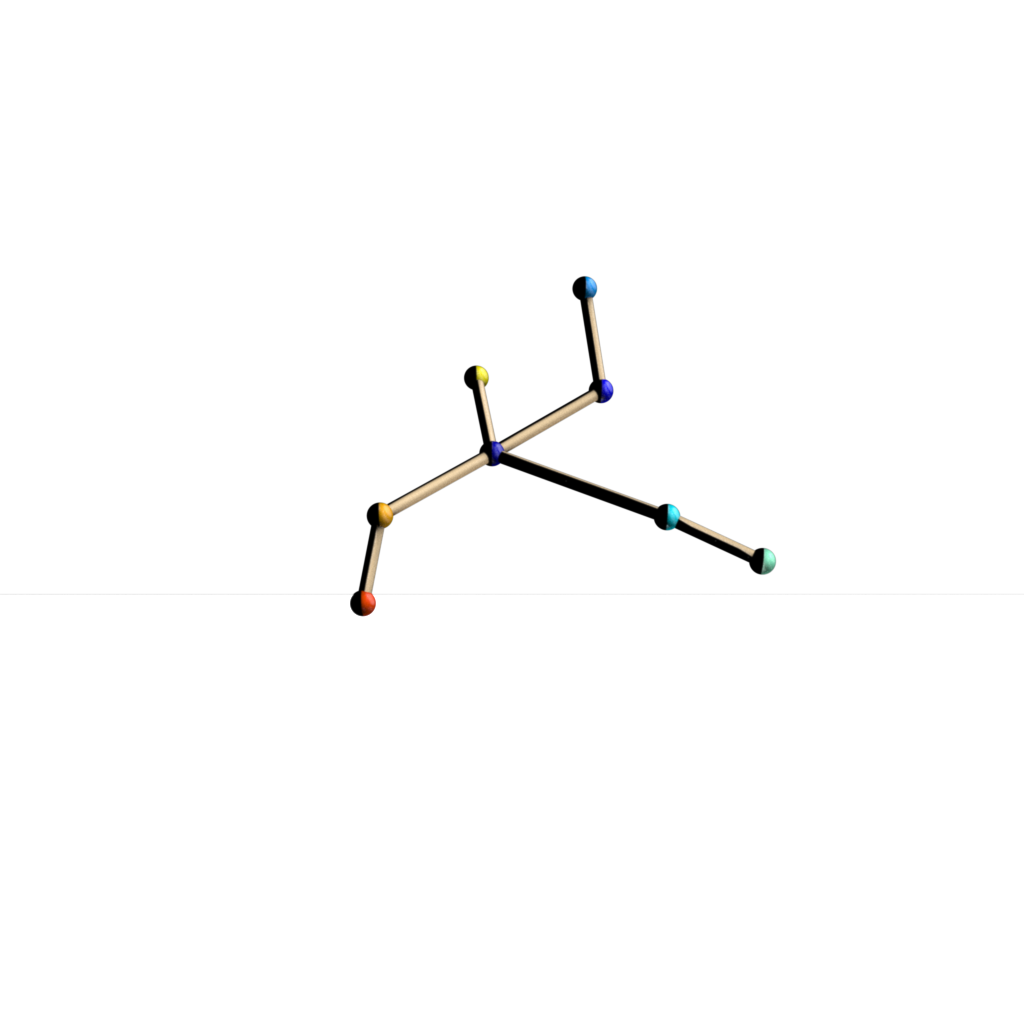}{0.15}{9.75cm 10cm 7.25cm 7cm}
        \robotvis{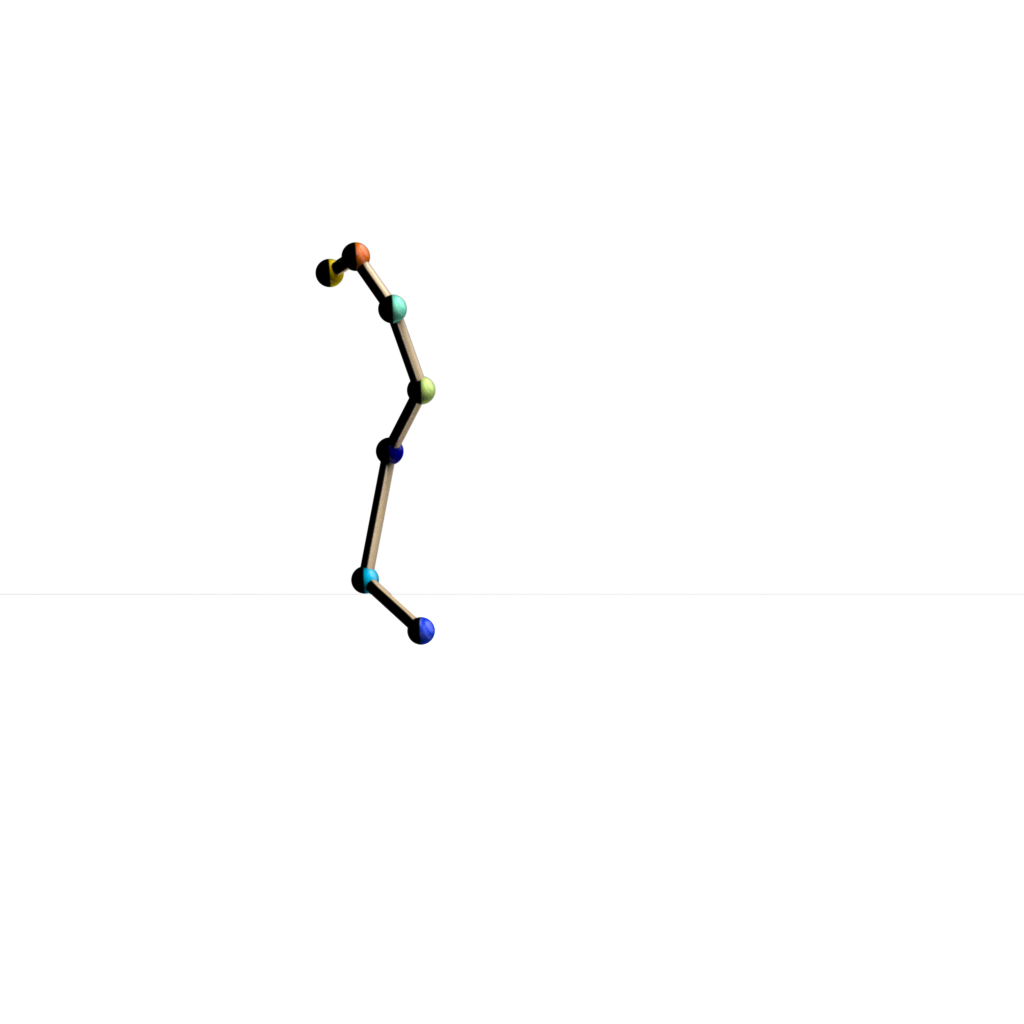}{0.15}{4.50cm 9.75cm 12.50cm 7.25cm}
        (a) Input Point Clouds &
        (b) Part Segmentation &
        (c) Topology &
        (d) Flow &
        (e) Part Segmentation &
        (f) Topology &
        (g) Flow \\
    \end{tabular}
    }
\captionof{figure}{\textbf{Qualitative Results on \robotD Dataset.} Given the
input point cloud sequence (shown in (a)), we show the part segmentation,  part connectivity, and implied flow using our inferred articulated model (in (b, c, d)) and the ground truth articulated model (in (e, f, g)) across 4 different type of robots. Our method can successfully deal with various parts connected in complex ways.} 
\label{fig:robots}
\vspace{-3mm}
\end{table*}

\begin{table}[]
    \centering
    \small
    \resizebox{\linewidth}{!}{
    \setlength{\tabcolsep}{0.1em} %
    \begin{tabular}{cccc}
        \comparevis{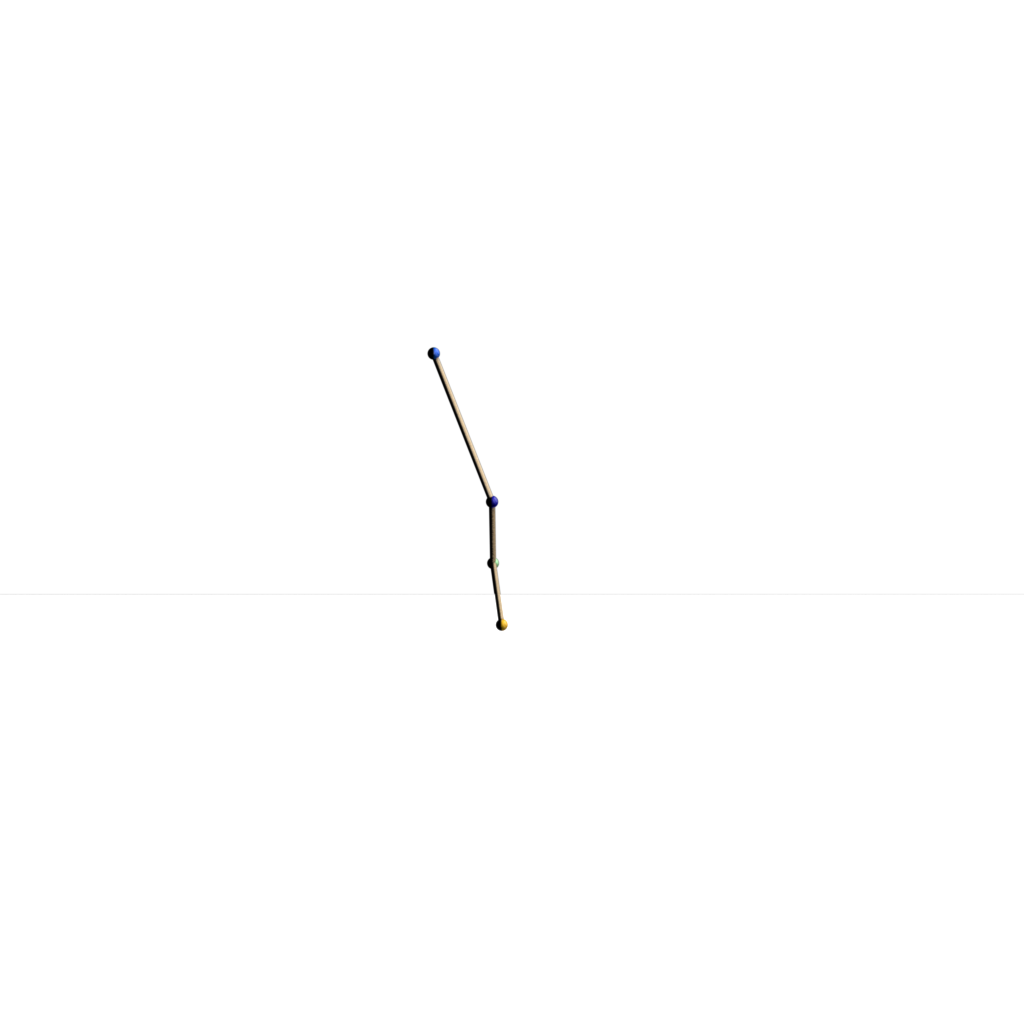}{0.3}{6.5cm 10cm 10.5cm 7cm}
        \comparevis{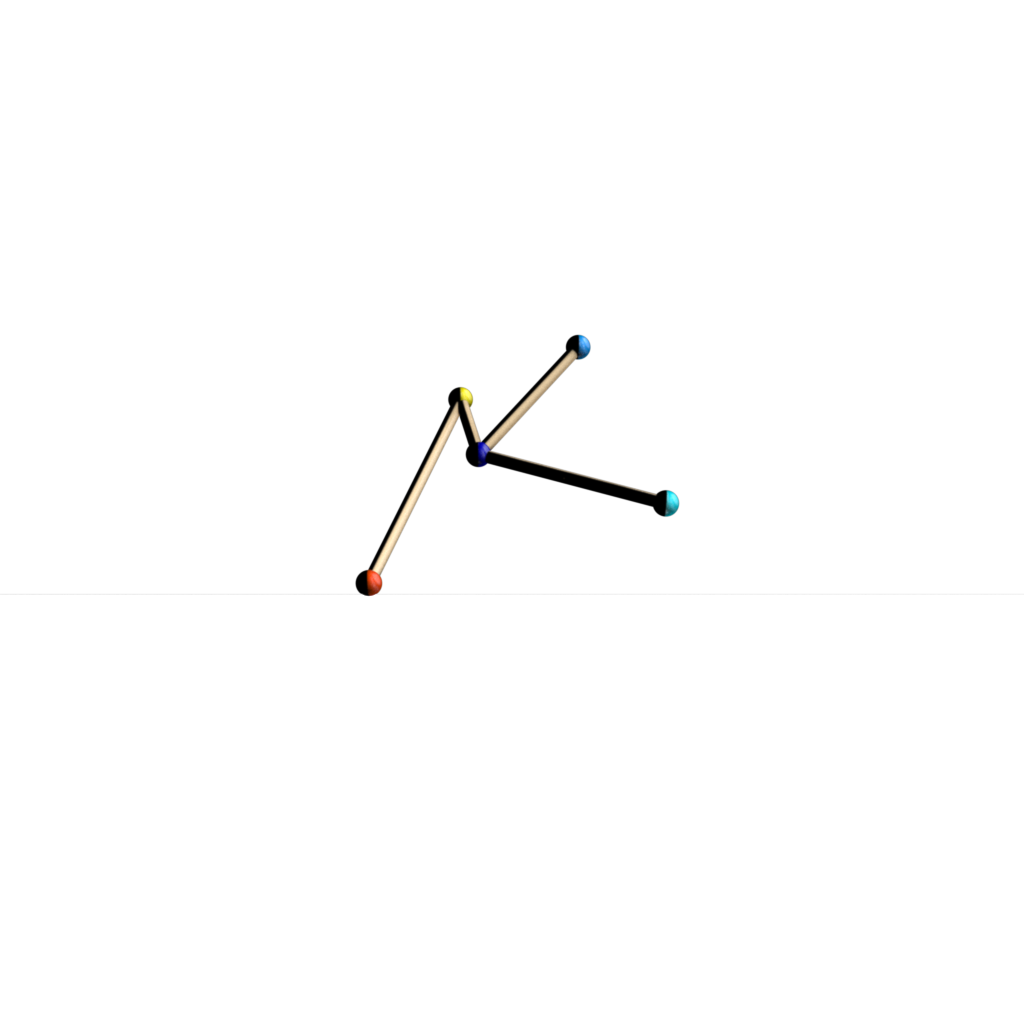}{0.3}{9.75cm 10cm 7.25cm 7cm}
        MultiBodySync*~\cite{huang2021multibodysync} & 
        \wim~\cite{noguchi2022watch} &
        Ours & 
        Ground Truth
    \end{tabular}
    }
\captionof{figure}{\textbf{Qualitative comparison against \mbs~\cite{huang2021multibodysync} and \wim~\cite{noguchi2022watch} on the \robotD dataset.} Note, a) \mbs by itself doesn't produce a kinematic tree, we use our method on top of their output to generate one, and b) we provide \wim~\cite{noguchi2022watch} with the ground truth SDFs and number of parts (which are not used by our method). Even after these modifications, the past methods cannot solve the task as well as ours.}
\label{fig:comparison}
\vspace{-3mm}
\end{table}
\begin{table}[]
    \centering
    \footnotesize
    \resizebox{\linewidth}{!}{
    \setlength{\tabcolsep}{0.1em} %
    \begin{tabular}{ccc}
        \sapienvis{56}{2cm 6cm 7cm 3cm} \\
         \multicolumn{3}{c}{Microwave}\\
        \sapienvis{212}{7cm 7cm 7cm 7cm} \\
        \multicolumn{3}{c}{Eyeglasses}\\
        \sapienvis{272}{7cm 7cm 7cm 7cm} \\
        \multicolumn{3}{c}{Trash Can}\\
        \sapienvis{343}{4cm 4cm 0cm 0cm} \\
        \multicolumn{3}{c}{Box}\\
        (a) Input & (b) Predicted Model & (c) Ground Truth
    \end{tabular}
    }
\captionof{figure}{\textbf{Qualitative results on \sapiensD dataset from~\cite{huang2021multibodysync}.} We visualize the predicted and ground truth articulated models. We show results on 4 daily objects with different number of parts and articulation types, a microwave (single revolute), eyeglasses (two revolute), trash can (a prismatic and a revolute) and a box (4 revolute). Different parts are in different colors, and we also show the screw parameters (in red) for the inferred joints. }
\label{fig:sapien}
\vspace{-3mm}
\end{table}

\subsection{Experimental Setup} 
\noindent\textbf{Datasets.} We conduct experiments on two datasets: the
{\it \robotD} dataset that we introduce in this work, and the {\it \sapiensD} dataset from~\cite{huang2021multibodysync}. The \robotD dataset consists of 18 popular robots which span manipulator like panda robot to humanoid robots like nao. These robots have large variation in number of parts (7--15), part geometries (barrett robot's fingers to atlas robot's trunk), and part connectivity (linear chains to complex trees). We split the dataset in training (4 robots), validation (4 robots), and test (10 robots). For each robot we articulate them using~\cite{urdfpy} to record a 10 time-step points cloud sequence, each containing 4096 points. 
Crucially, we make sure that these 4096 points are randomly resampled at each time step to prevent any leakage of correspondence information between time steps. Supplementary material shows visualizations for different robots and summarizes more statistics.
The \sapiensD dataset from~\cite{huang2021multibodysync} contains 720 test sequences across 20 different object categories from Sapien~\cite{xiang2020sapien}. The dataset provides 4 point cloud frames as input for each object.
The 4 frames all have a different global coordinate frame.

\noindent\textbf{Metrics.} Our experiments measure the final reanimation error, intermediate metrics that evaluate part segmentation and  
inferred kinematics, and reconstruction metrics that evaluate how well our inferred models explain the input point clouds. We detail them next.
\textbf{Reconstruction metrics} include a) Reconstruction Error that measures the mean end-point-error between the object articulated using the inferred articulation parameters \vs with the ground truth parameters, 
b) Flow Error that measures the error between the flow between consecutive frames implied by the predicted model \vs the ground truth model, and c) Flow Accuracy that measures the \%age of points for which flow implied by the predictions is within a $\delta$ threshold from ground truth flow.
\textbf{Intermediate metrics} include a) Random Index (RI)~\cite{chen2009benchmark} that measures overlap between the partition induced by predicted part labels \vs the partition induced by ground truth part labels and b) Tree Edit Distance~\cite{zhang1989simple, pawlik2015efficient} that measures the similarity between predicted kinematic tree and the ground truth kinematic tree. These metrics have been used in past works, ~\cite{huang2021multibodysync} and~\cite{xu2020rignet} respectively. 
{\bf Reanimataion Metric:} 
Our inferred articulated model can be reanimated given new locations for a sparse set of points (1 per part). We measure the quality of such reanimation via the mean end-point-error between the ground truth reanimated model and the predicted reanimated model. Precise metric definition is provided in the supplementary. 

\noindent\textbf{Implementation Details.}
Our training requires us to compute the flow between pairs of frames. We train a Siamese correspondence network using PointNet++~\cite{qi2017pointnet++}. At test time, we compute the similarity between each pair of points in the two point clouds and use correspondences that are mutually the best. The correspondence network is trained on the \sapiensD and the \robotD datasets and does not include any objects or robots that we evaluate upon (neither for validation nor for testing).
{\bf Part Segmentation Field} function $f$ is realized using an MLP with one hidden layer of 128 dimensions. All optimizations are done using PyTorch. We use the standard Adam optimizer~\cite{kingma2014adam} for all optimization with a learning rate of 1e-3 for the MLP and 1e-2 for transformations. During {\bf relaxed model estimation stage}, we set maximum number of parts to 20 and all 6DOF transformations are initialized to be identity. We found it helpful to only optimize with $E_\text{CD}$ and $E_\text{flow}$ for the first 5000 iterations, and replace $E_\text{CD}$ by $E_\text{EMD}$ and optimizing for another 10000 iterations. For efficiency reasons, we apply $E_\text{EMD}$ on 4$\times$ downsampled point clouds and only update the optimal assignment every 5 iterations.
During {\bf projection stage}, we first merge parts that are spatially close and don't move relative to one another with $\epsilon_{m}=3e-2$. 
During the final optimization stage, we only optimize the screw parameters and their states. We keep the number of parts and part segmentation fixed. Merging details and visualization can be found in supplementary.

\begin{table}[t]
    \centering
    \footnotesize
    \caption{{\bf Comparison to past methods on \robotD test set.}
    We outperform past methods on all metrics that measure input reconstruction quality (shape reconstruction error, flow error, flow accuracy), model accuracy (part segmentation rand index, and kinematic tree edit distance), and reanimiation error.}
    \label{tab:robots}
    \setlength{\tabcolsep}{2pt}
    \resizebox{\linewidth}{!}{
    \begin{tabular}{lcccccc}
        \toprule
         Method & Recons.            & Flow               & Flow            & Rand           & Tree Edit & Reanimate \\
                & error $\downarrow$ & error $\downarrow$ & acc. $\uparrow$ & Index $\uparrow$ & Distance $\downarrow$ & Error  $\downarrow$ \\
         \midrule 
         \mbs~\cite{huang2021multibodysync}  & 4.76 & 3.42 & 0.26 & 0.70 &  6.6 & 9.72\\
         \wim~\cite{noguchi2022watch}  & 12.77 & 8.42 & 0.06 & 0.78 & 6.6 &  7.43\\
         Ours & \textbf{1.26} & \textbf{0.57} & \textbf{0.59} & \textbf{0.86} & \textbf{2.9} & \textbf{3.66}\\
         \bottomrule
    \end{tabular}}
\end{table}

\begin{table}[t]
    \centering
    \footnotesize
    \caption{{\bf Comparison to past methods on \sapiensD dataset~\cite{huang2021multibodysync}.} 
    We do better in both flow estimation and segmentation under two different evaluation settings.}
    \label{tab:sapiens}
    \setlength{\tabcolsep}{4pt}
    \resizebox{\linewidth}{!}{
    \begin{tabular}{lccc}
        \toprule
        Method & Flow Error $\downarrow$ & Multi-scan RI$\uparrow$ & Per-scan 
        RI$\uparrow$ \\
        \midrule
        PWC-Net~\cite{wu2020pointpwc} & 6.20 & - & - \\
        PointNet++~\cite{qi2017pointnet++} & - & 0.62 & 0.65 \\
        MeteorNet~\cite{liu2019meteornet} & - & 0.59 & 0.60 \\
        Deep Part~\cite{yi2018deep} & 5.95 & 0.64 & 0.67 \\
        NPP~\cite{hayden2020nonparametric} & 21.22 & 0.63 & 0.66 \\
        MultiBodySync~\cite{huang2021multibodysync} & 5.03 & 0.76 & 0.77 \\
        Ours & \bf 4.79 & \bf 0.79 & \bf 0.79\\
         \bottomrule
    \end{tabular}}
\end{table}

\begin{table}[t]
    \centering
    \footnotesize
    \caption{\textbf{Role of different energy terms in \cref{eq:recons} evaluated on the \robotD validation set.} To isolate the direct impact of the energy terms, we conduct this experiment w/o the canonical frame selection (we always use the middle frame) and w/o screw parameter finetuing after projection to a kinematic model. All three terms are important with the flow term being the most important.}
    \label{tab:energy}
    \setlength{\tabcolsep}{3pt}
    \resizebox{\linewidth}{!}{
    \begin{tabular}{ccc|ccccc}
        \toprule
         ${E}_\text{CD}$ & ${E}_\text{flow}$ & ${E}_\text{EMD}$ & Recons.& Flow  & Flow & Rand & Tree Edit \\
         & &                                                    & error $\downarrow$ & error $\downarrow$ & acc. $\uparrow$ & Index $\uparrow$ & Distance $\downarrow$ \\
         \midrule
         \cmark &  &  & 6.97 & 8.05 & 0.31 & 0.76 & 6.50 \\
          & \cmark &  & 2.97 & 1.64 & 0.31 & 0.83 & 5.00 \\
          &  & \cmark & 3.49 & 3.03 & 0.13 & 0.28 & 6.75  \\
         \cmark & \cmark &  & 1.47 & 0.88 & \bf 0.48 & 0.83 & 4.00 \\
         \cmark &  & \cmark & 1.93 & 1.99 & 0.17 & \bf 0.86 & 4.25 \\
          & \cmark & \cmark & 1.54 & 0.97 & 0.33 & 0.83 & 5.25 \\
         \cmark & \cmark & \cmark & \bf 1.31 & \bf 0.86 & 0.40 & \bf 0.86 & \bf 3.25 \\
         \bottomrule
    \end{tabular}}
\end{table}

\begin{table}[t]
    \centering
    \footnotesize
    \caption{{\bf Evaluation of other design choices on the \robotD validation set.} Neural segmentation field ($\textbf{f}$), Gumbel-softmax (\textbf{Gumbel}), projection to the kinematic model (\textbf{Project}), canonical frame selection (\textbf{Cano}) all contribute to the final performance.}
    \label{tab:ablations}
    \setlength{\tabcolsep}{3pt}
    \resizebox{\linewidth}{!}{
    \begin{tabular}{lccccccc}
        \toprule
         Designs & Recons.            & Flow               & Flow            & Rand           & Tree Edit & Reanimate \\
                & error $\downarrow$ & error $\downarrow$ & acc. $\uparrow$ & Index $\uparrow$ & Distance $\downarrow$ & Error $\downarrow$ \\
         \midrule
         w/o $\textbf{f}$ & 2.63 & 1.26 & 0.40 & 0.76 & 7.25 & 11.68\\
         w/o \textbf{Gumbel} & 3.54 & 1.65 & 0.34 & 0.74 & 7.25 & 11.69  \\
         w/o \textbf{Project} & 1.25 & 0.75 & 0.45 & \bf 0.88 & \bf 3.00 & \phantom{1}8.86\\
         w/o \textbf{Cano} &  1.54 & 0.74 & 0.47 & 0.85 & 3.33 & \phantom{1}6.83\\
         ours & \bf 1.19 & \bf 0.64 & \bf 0.49 & \bf 0.88 & \bf \bf 3.00 & \phantom{1}6.50 \\
         \bottomrule
    \end{tabular}}
\end{table}

\begin{table}[t]
    \centering
    \footnotesize
     \caption{{\bf Testing performance between prismatic \vs revolute joints on \sapiensD dataset.} Prismatic joints could be harder to predict than revolute joints.}
    \label{tab:supp-joint-type}
    \setlength{\tabcolsep}{4pt}
    \resizebox{\linewidth}{!}
    {
    \begin{tabular}{lccc}
        \toprule
        Method & Flow Error $\downarrow$ & Multi-scan RI$\uparrow$ & Per-scan 
        RI$\uparrow$ \\
        \midrule
        Prismatic & 4.80 & 0.64 & 0.64 \\
        Revolute & 4.78 & 0.80 & 0.80 \\
        \bottomrule
    \end{tabular}}
\end{table}

\subsection{Articulation Object Modeling Results}
\noindent\textbf{Comparison on the \robotD Dataset.} As discussed in \cref{sec:related}, to the best of our knowledge, there isn't a past work that exactly solves the entire task we tackle. We adapt and extend  \mbs~\cite{huang2021multibodysync} and \wim~\cite{noguchi2022watch} for our task and compare them on the \robotD dataset. \mbs~\cite{huang2021multibodysync} uses pairwise flow prediction and sets up a joint synchronization problem to obtain a temporally-consistent part segmentation. We use the part segmentation and pairwise flows to produce a kinematic tree and associated screw parameters using the projection part of our algorithm.
\wim~\cite{noguchi2022watch} was designed for building articulable models from RGB videos. We modify it to work with point clouds. For point clouds, there is no rendering loss but only a loss on the predicted SDF. We provide ground truth SDFs and ground truth numbers of parts to their method.

Qualitative results on different types of robots are shown in \cref{fig:robots}. Our method can successfully deal with various parts connected in complex ways. \cref{tab:robots} presents the quantitative results on the \robotD. We outperform both these past methods by a large margin across all metrics. We looked into the poor performance of these past methods. \mbs relies on accurate {\it pairwise} flow predictions. For objects with a long chain of articulation, \eg a robot arm, over time, the object deforms quite a bit limiting the performance of the correspondence network. For \wim, we believe poor performance comes from some their use of {\it soft} assignment of points to segments during training. This leads to incorrect assignment of points to parts at test time.
In our own ablations (presented in \cref{sec:ablations})  we note that our hard assignment of points to segments during training is crucial to the final performance. \cref{fig:comparison} show qualitative comparisons.

\begin{table}[]
    \centering
    \small
    \resizebox{\linewidth}{!}{
    \setlength{\tabcolsep}{0.1em} %
    \begin{tabular}{cccc}
        \reanimatecompare{jvrc}{0.3}{8.0cm 11cm 9.0cm 6.0cm}{0}
        \reanimatecompare{solo}{0.3}{7.25cm 9cm 5.75cm 6cm}{2}
        (a) Input & 
        (b) w/o kinematics &
        (c) w/ kinematics & 
        (d) GT 
    \end{tabular}
    }
\captionof{figure}{\textbf{Reanimation Results on the \robotD Dataset.} Given new locations for a sparse set of points on the object(shown in (a)), our full method (shown in (c)) is able to generate a reasonable reanimation to match the specified points. (b) shows reanimation results from an ablated version of our model where we don't restrict the parts to form a kinematic tree. This results in poor reanimation. Thus, correctly inferring the connectivity of different parts with one another is crucial for high-quality reanimation.}
\label{fig:animation}
\vspace{-5mm}
\end{table}

\begin{table}[]
    \centering
    \footnotesize
    \resizebox{\linewidth}{!}{
    \setlength{\tabcolsep}{0.1em} %
    \begin{tabular}{cccc}
        \imgtile{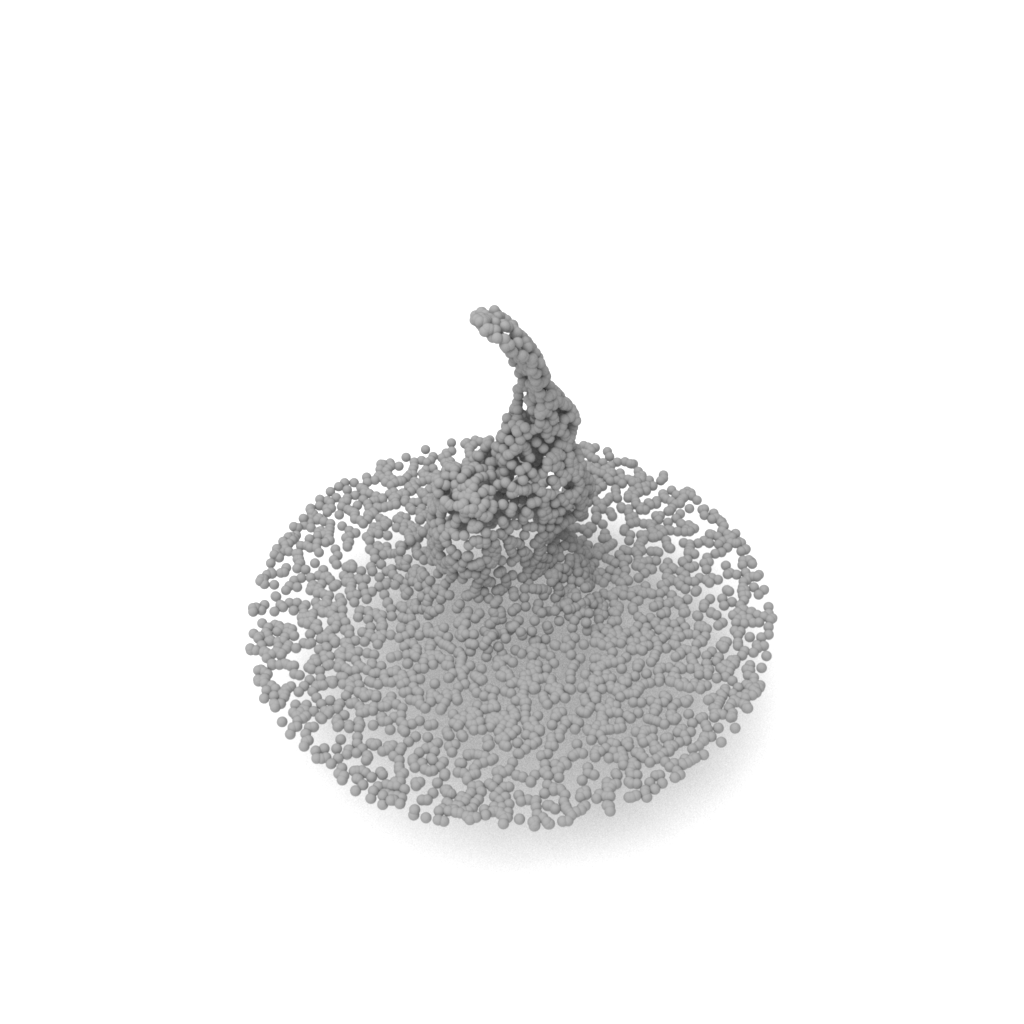}{0.22} & 
        \includegraphics[trim={5cm 5cm 5cm 5cm},clip,width=0.22\linewidth]{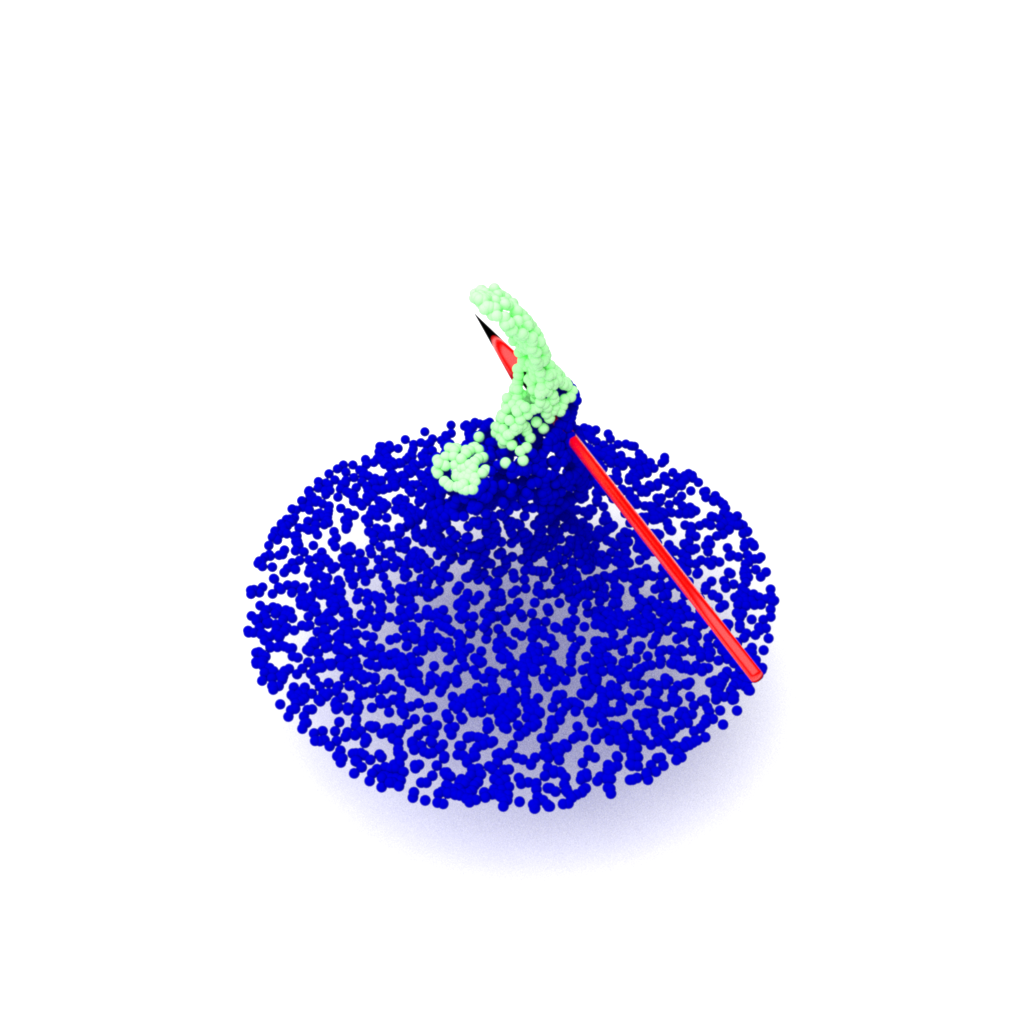} & 
        \includegraphics[trim={5cm 5cm 5cm 5cm},clip,width=0.22\linewidth]{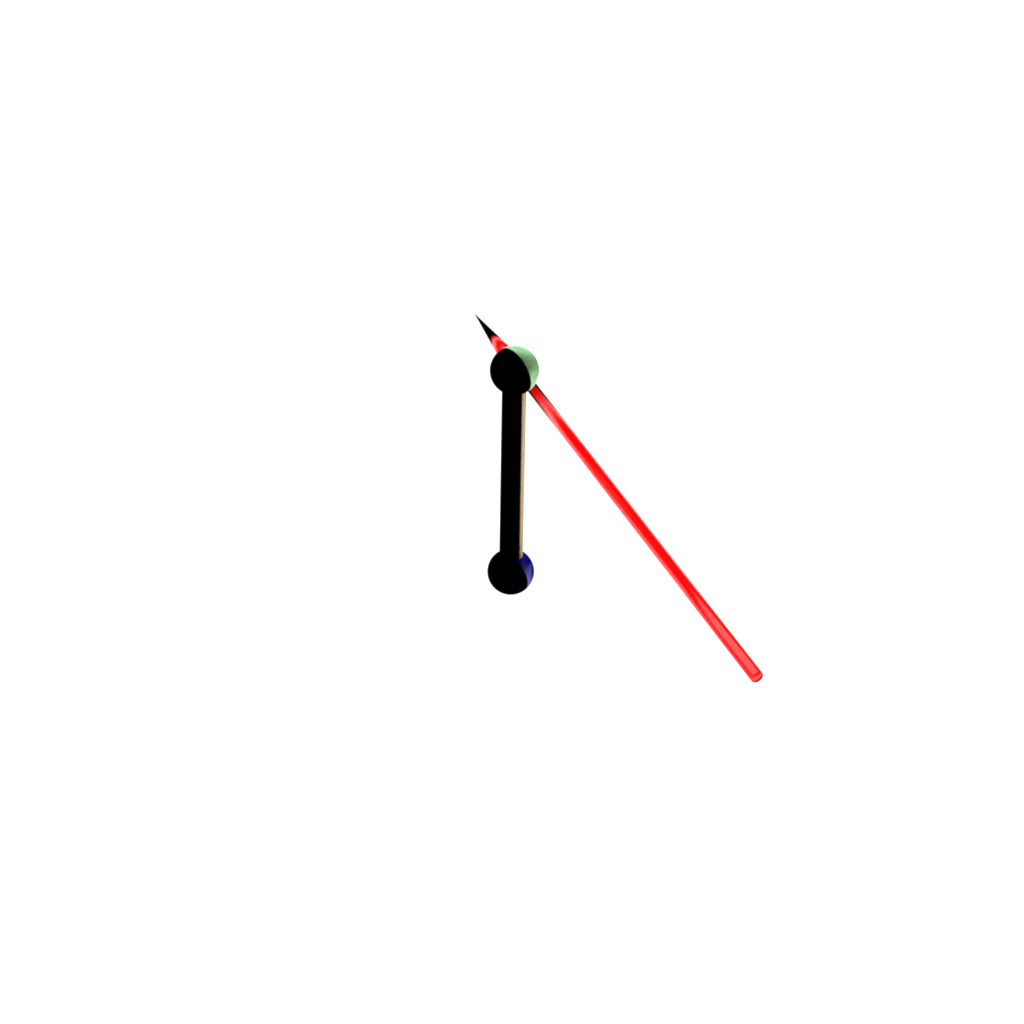} & 
        \includegraphics[width=0.20\linewidth]{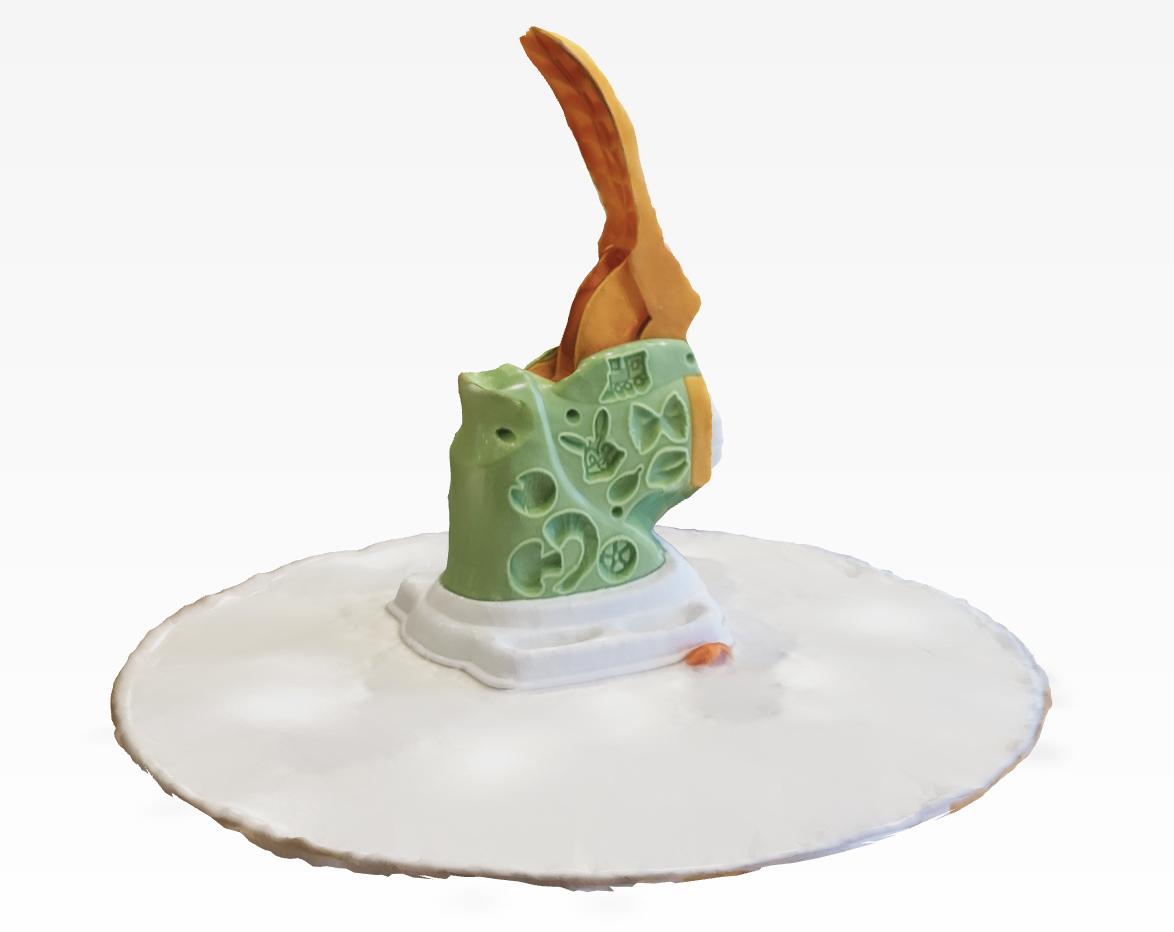} \\
        \multicolumn{4}{c}{Toy}\\
        \imgtile{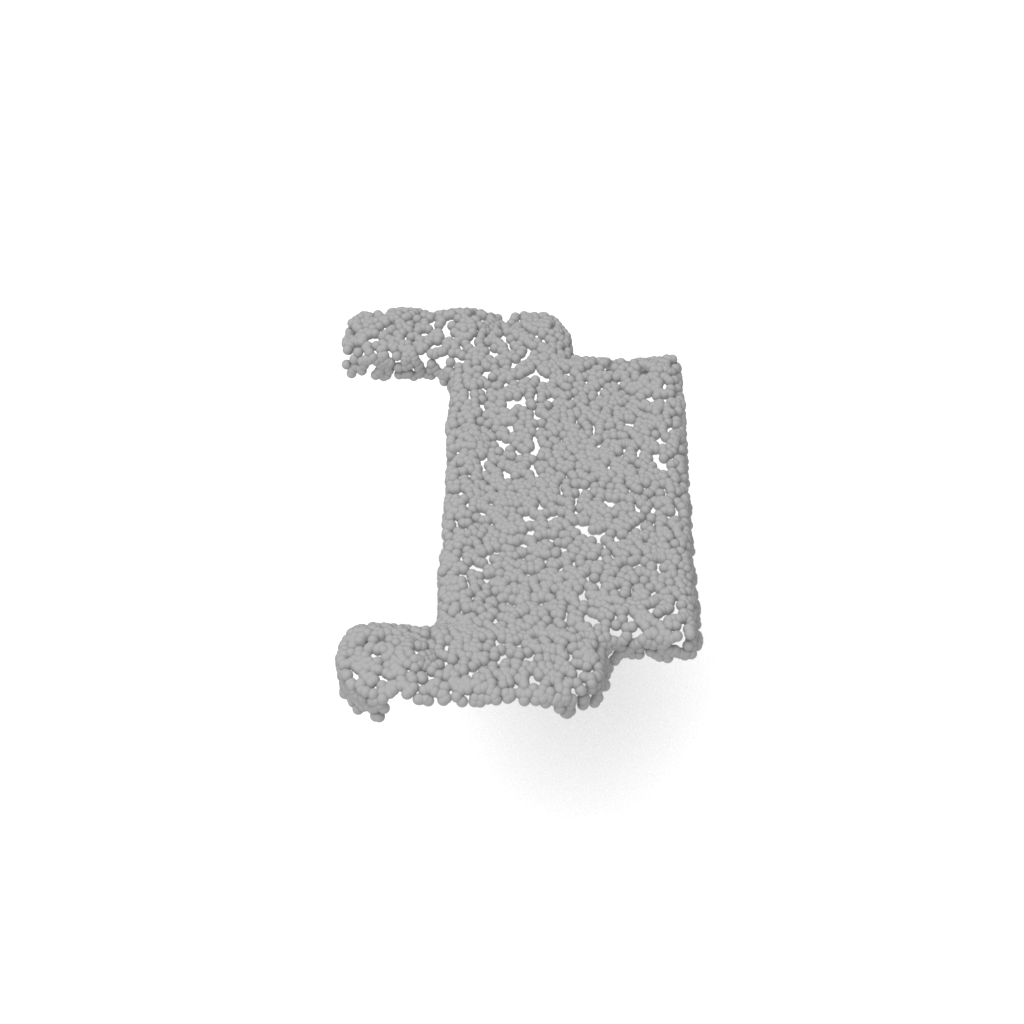}{0.22} & 
        \includegraphics[trim={7cm 7cm 7cm 7cm},clip,width=0.22\linewidth]{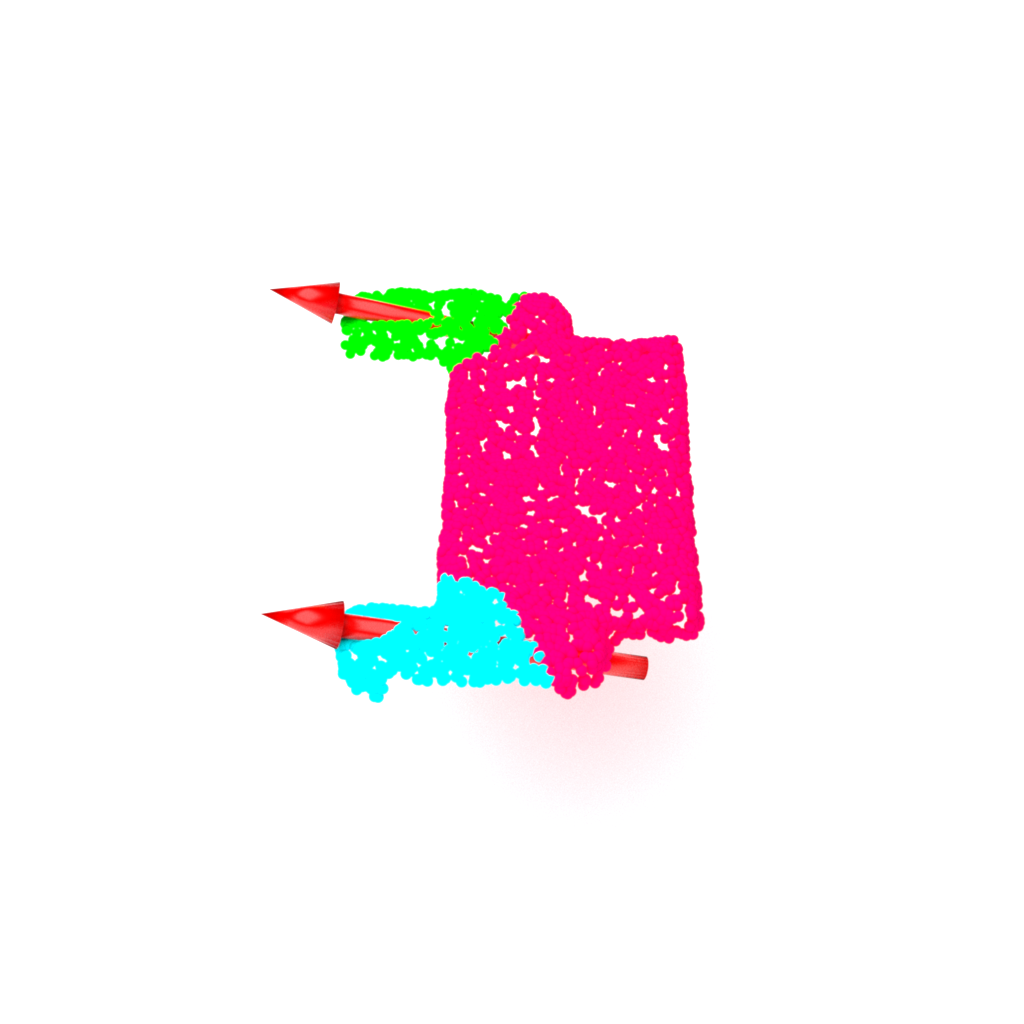} & 
        \includegraphics[trim={7cm 7cm 7cm 7cm},clip,width=0.22\linewidth]{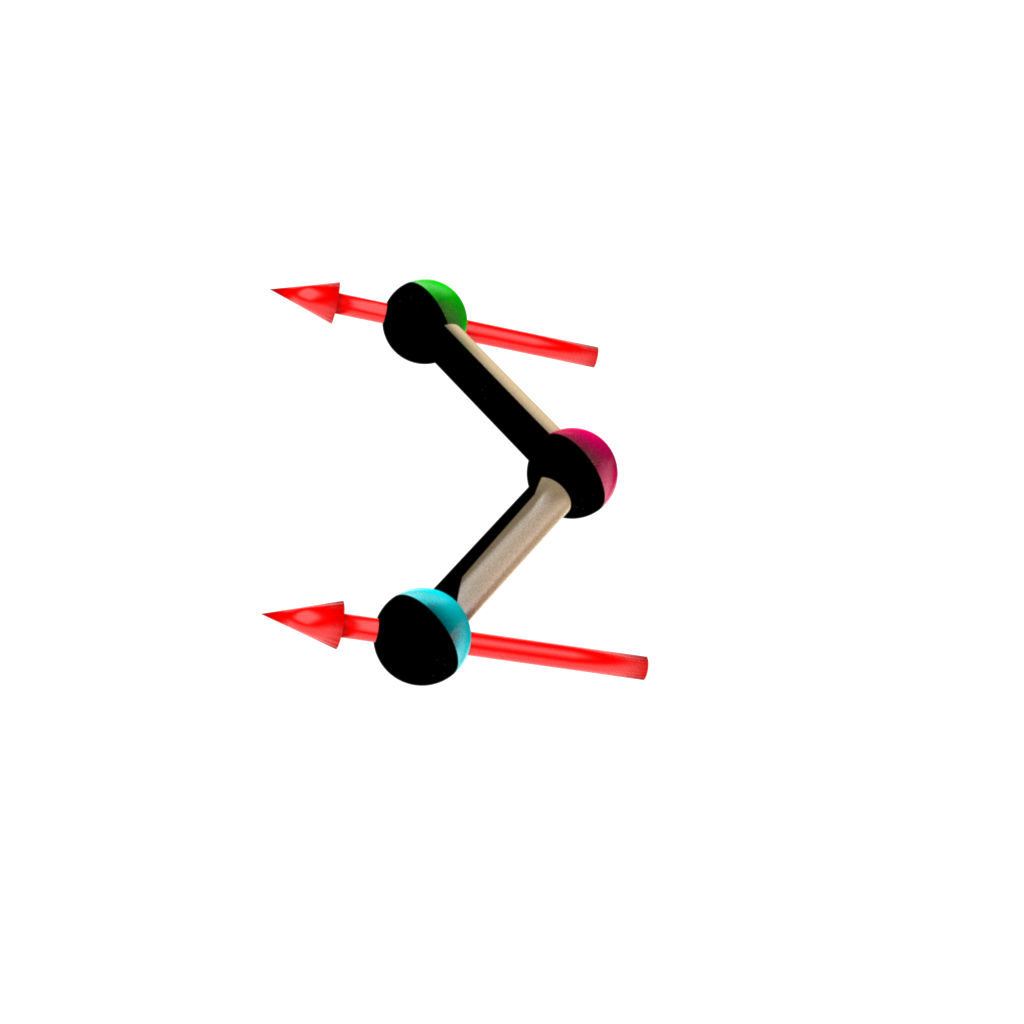} & 
        \includegraphics[width=0.20\linewidth]{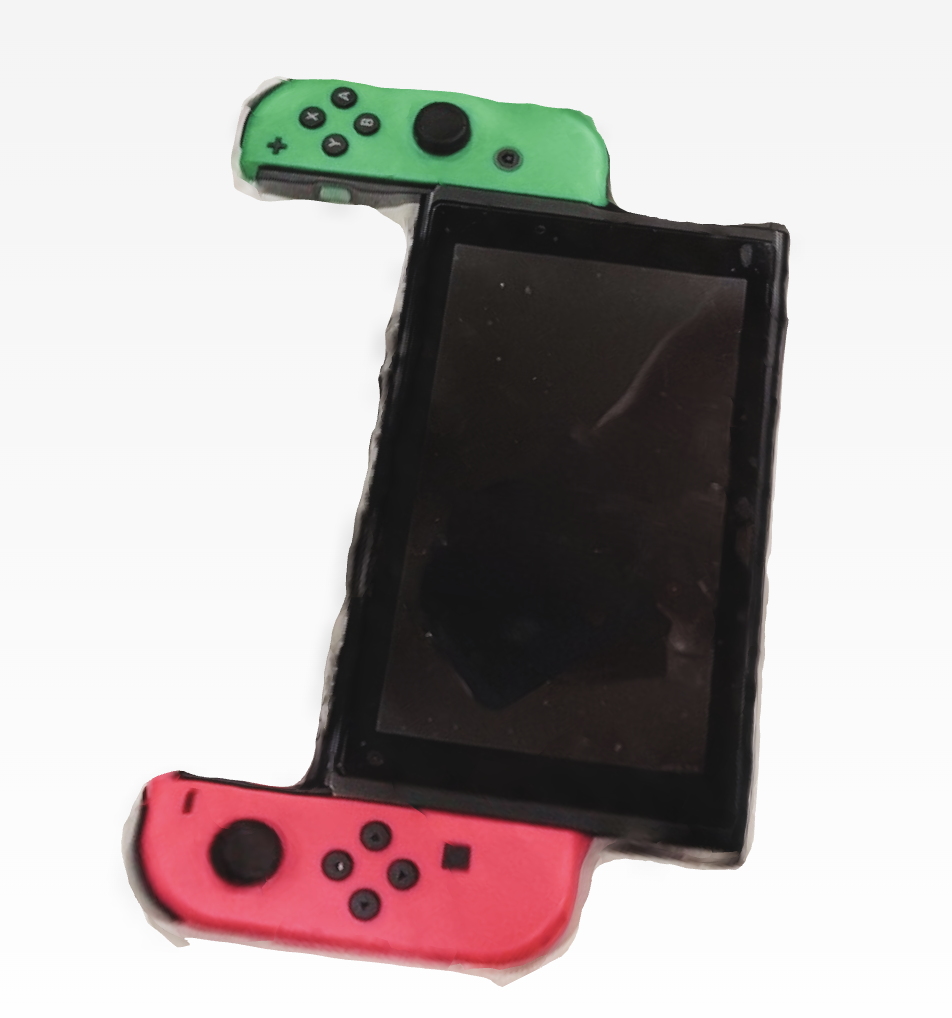}\\
        \multicolumn{4}{c}{Switch}\\
        \imgtile{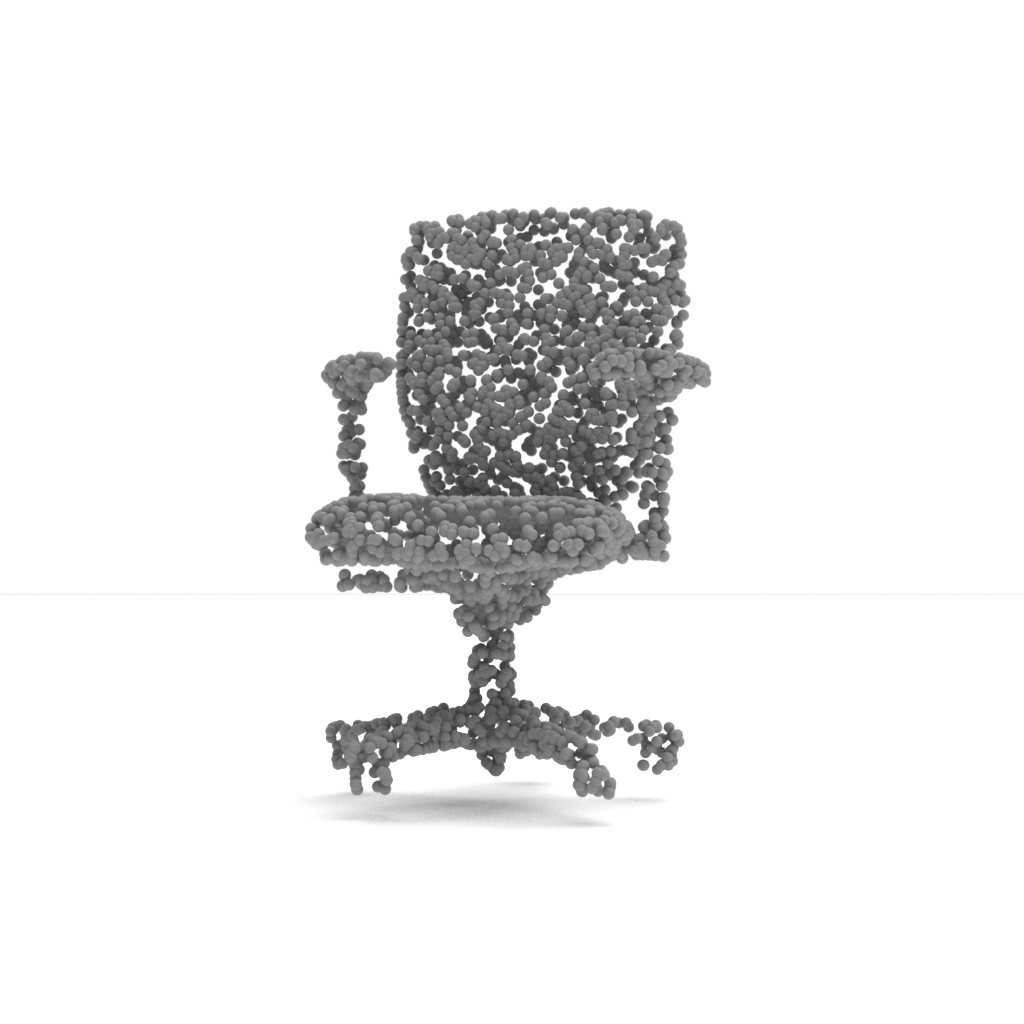}{0.22} & 
        \includegraphics[trim={5cm 5cm 5cm 5cm},clip,width=0.22\linewidth]{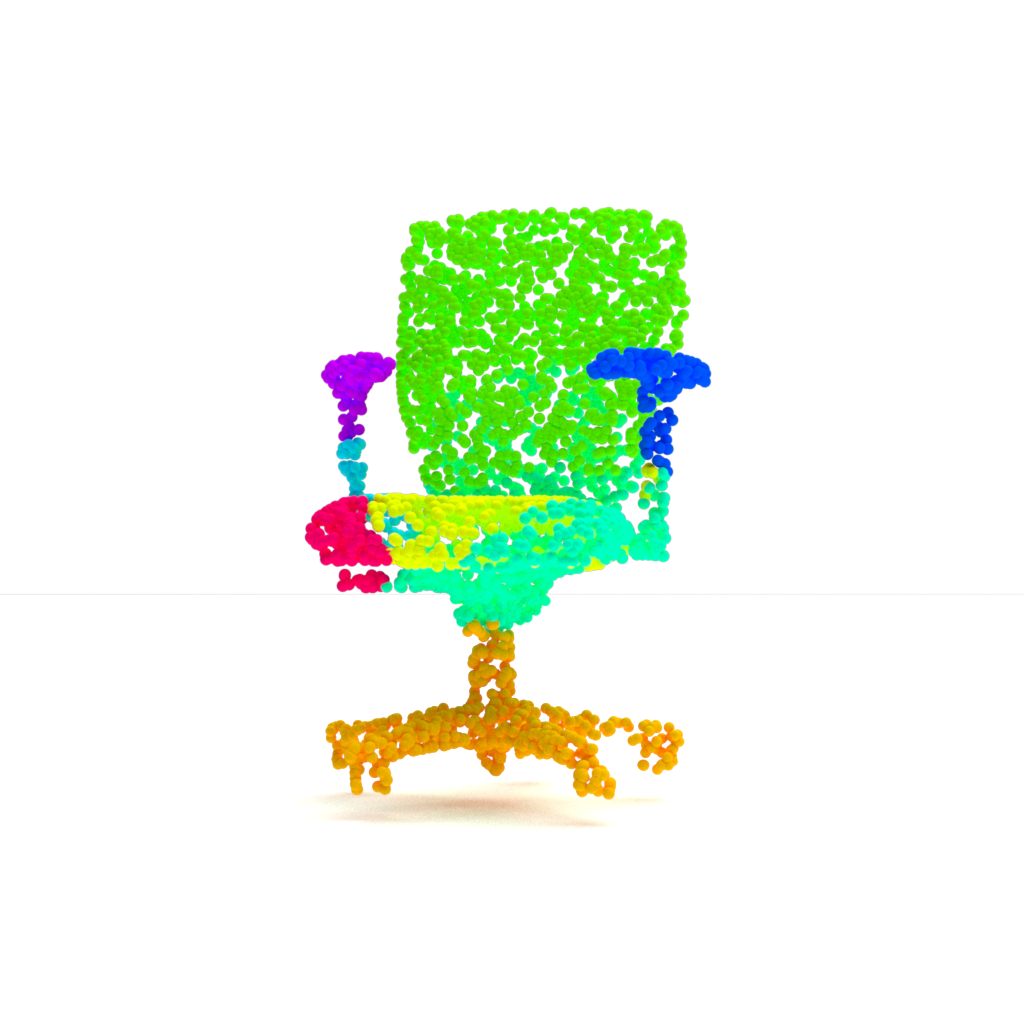} & 
        \includegraphics[trim={5cm 5cm 5cm 5cm},clip,width=0.22\linewidth]{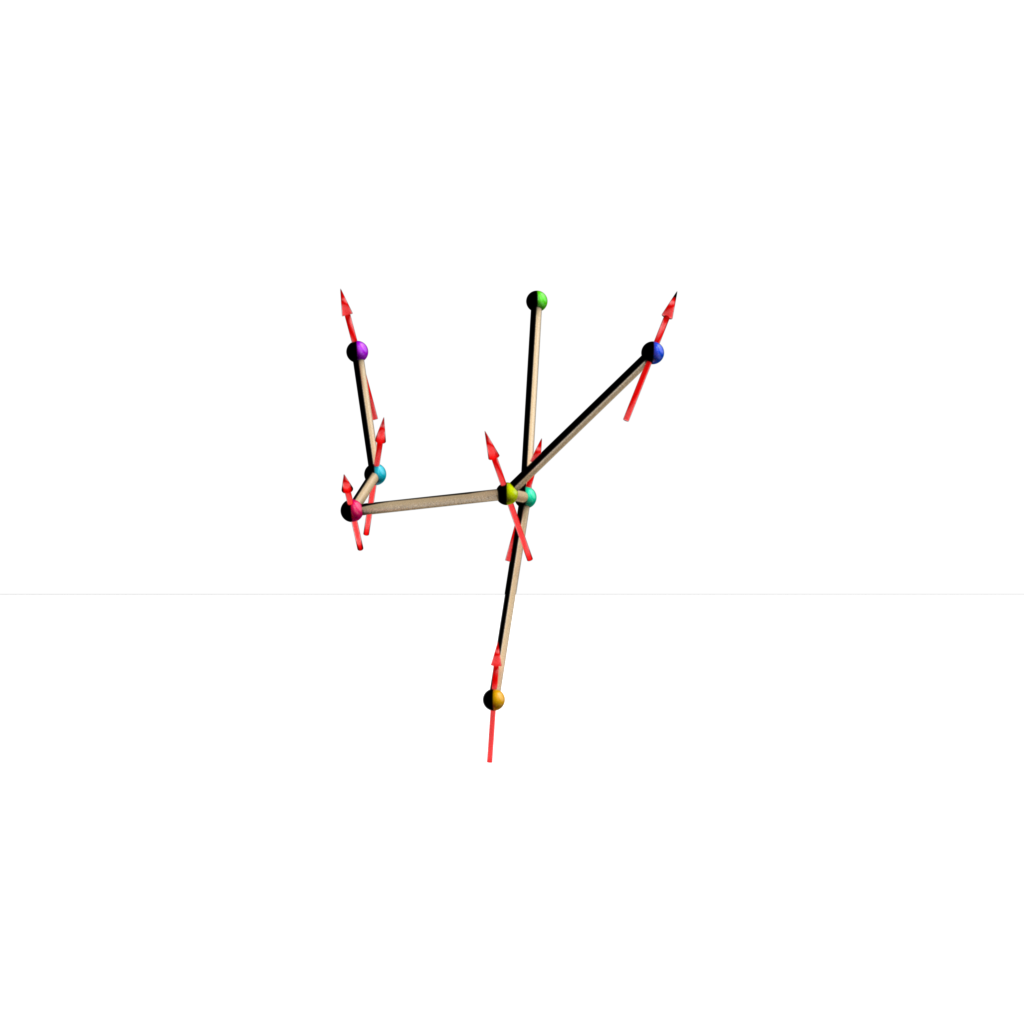} & 
        \includegraphics[width=0.15\linewidth]{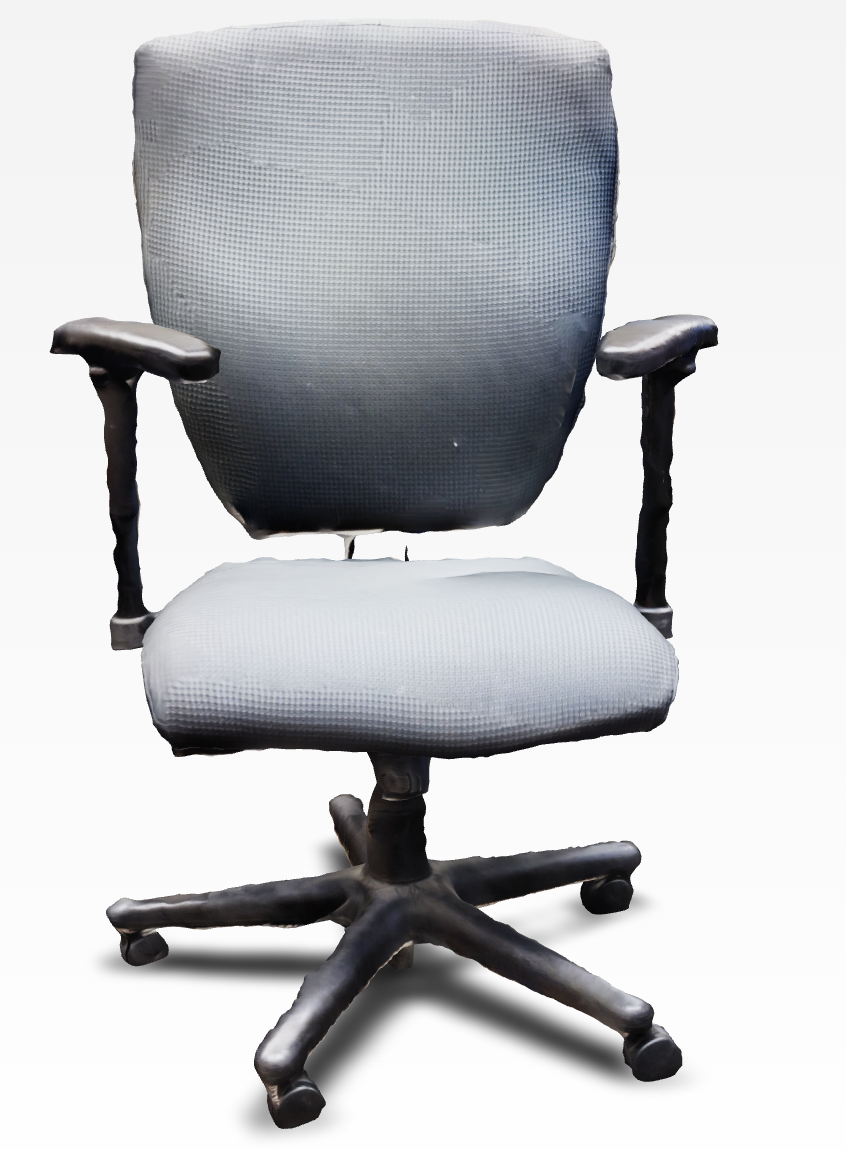}\\
        \multicolumn{4}{c}{Chair}\\
        (a) Input &
        (b) Segmentation &
        (c) Kinematics &
        (d) Real scan \\
    \end{tabular}
    }
    \captionof{figure}{\textbf{Qualitative results in real-world setting.} We verify our method on three daily objects in each row, a toy (single revolute), a switch (two prismatic) and a chair (multiple revolute and prismatic). Each row shows in the input, part segmentation, part connectivity and screw parameters (in red) for the inferred joints, and the real scan captured by scanning apps of iPhone. Our method is robust to noise and partial observations. Chair cushion gets slightly over-seg due to noisy surface.}
\label{fig:real}
\end{table}

\noindent \textbf{Comparison on \sapiensD Dataset.}
We also compare performance on specific sub-tasks (part segmentation and flow prediction) useful for re-animation as studied in past work~\cite{huang2021multibodysync} on their \sapiensD dataset in \cref{tab:sapiens}. Here, we also compare a number of past methods, including PWC-Net~\cite{wu2020pointpwc}, PointNet++\cite{qi2017pointnet++}, \etc. Please refer to~\cite{huang2021multibodysync} for details about these methods. We outperform all these past methods. We believe our strong performance on this dataset is because of the additional joint and kinematic constraints enforced by our method, which are not used by these past methods. See qualitative results in \cref{fig:sapien} on objects from \sapiensD dataset with different number of parts and
articulation types (revolute and prismatic). For a fair comparison, we use the flow network from~\cite{huang2021multibodysync} for our method.

\subsection{Ablations}
\label{sec:ablations}
\cref{tab:energy} evaluates the importance of the different energy terms in our formulation on \robotD validation set. We note that all terms are important, but particularly $E_\text{flow}$ has the largest effect. We also find that just the chamfer distance term, $E_\text{CD}$, is insufficient and works poorly. We believe this is because $E_\text{CD}$ doesn't provide good gradients (it tries to snap a point to the closest point in the shape, which may not necessarily be the correct location for it). 

\cref{tab:ablations} reports the impact of our other design choices. Representing the part segmentation as a neural field induces a smoothness prior on the parts. Removing this prior by optimizing each point' segmentation logits independently 
works worse (Row 1: w/o $\textbf{f}$). Taking Gumbel-softmax as
hard assignment during training is important. Soft assignment (softmax) of points to transformations provide a lot more freedom for optimization. This prevents learning of the correct transformations, which don't work when used with hard assignments at test time (Row 2, w/o \textbf{Gumbel}). Projecting relaxed model to valid kinematic one leads to better reconstruction quality and lower re-animation error by further constraint the transformations and regularize the motions (Row 3, w/o \textbf{Project}). We also found that the choice of frame for building the neural part field is important. Sometimes two otherwise far away parts can be close to one another in some frame (\eg left and right arms of a humanoid) making it difficult for the part segmentation field to separate them. Searching for which frame to use as the canonical frame helps in over-coming the non-convexity in the optimization (Row 4, w/o $\textbf{Cano}$).

\noindent \textbf{Re-animation Results.} We now showcase how the reconstructed rearticulatable model can be retargeted to new poses. Given an inferred model consisting of part segmentation, kinematic tree, and 1DoF joint parameters, we use inverse kinematics (IK) to compute the transformation between our canonical point frame and a target pose, using new locations for a sparse set of points on the object. We then re-animate the point cloud through the inferred joint parameters. \cref{fig:animation} shows the results. 
We see that the re-animation quality significantly improves by inferring the kinematic linking of parts.

\noindent \textbf{Prismatic joints.} We compare testing performance between prismatic vs. revolute joints on \sapiensD dataset. The results are shown in \cref{tab:supp-joint-type}. Compare against revolute joints, prismatic joints could be harder to predict. The reasons include 1) less training samples; 2) segmentation is hard between base part and cluttered prismatic part. Qualitative comparisons of prismatic joints against revolute joints are shown in \cref{fig:sapien} and \cref{fig:real} under real-world setting.

\noindent\textbf{Run-time.} 
On \robotD sequences, ours takes 45 minutes on a RTX 3090 GPU while \wim takes 2 hours. On \sapiensD, ours takes $5$min (same as \mbs).

\subsection{Real-World Experiments}
\label{sec:real-exp}
We verified our method on real world scans with diverse articulations and kinematic structures. We choose three daily objects, a toy (single revolute), a switch (two prismatic) and a chair (multiple revolute and prismatic) and reconstruct their geometry. Each object has five articulation states. The scans are gathered using scanning apps on iPhone. The results are shown in \cref{fig:real}.  Our method is robust to noise and partial observations. Chair cushion gets slightly over-segmentation due to noisy surface.

\subsection{Limitation.} 
\label{sec:limitation}
Our method heavily relies on motion cues. It might fail to distinguish two rigid parts if they undergo the same motion in the entire sequence (\eg a humanoid robot moves both arms synchronously). We leave this to future work and plan to tackle it by incorporating appearance information.

\vspace{-2mm}
\section{Conclusion}
We presented a novel method for building arbitrary rearticulable models from a point cloud sequence. Our approach jointly infers part segmentation, screw-parametric joints, and kinematic tree structures in a category-agnostic manner. We validated our method's efficacy on two challenging datasets, showing superior results over previous leading works. We further showed that the inferred model could be retargeted at any novel pose, demonstrating the potential for reanimation and manipulation. 

{\footnotesize \noindent {\textbf{Acknowledgements:} This material is based upon work supported by an NSF CAREER Award (IIS-2143873) and the USDA/NSF AIFARMS National AI Institute (USDA \#2020-67021-32799). SW is supported, in part, by Amazon research award and Insper innovation grant.}}

{\small
\bibliographystyle{ieee_fullname}
\bibliography{egbib}
}

\appendix
\clearpage
\begin{center}
\textbf{\Large Appendix}
\end{center}

The supplementary material provides implementation details, additional results and details of \robotD dataset to support the main paper. In summary, we include
\begin{itemize}
    \item \cref{sec:supp-implementation}. \textbf{Implementation details.}  
    Implementation details include flow prediction network, kinematic model projection, Merging, final fitting, canonical frame selection, optimization details, rearticulation, baseline implementations and evaluation metrics.
     \item \cref{sec:supp-dataset} \textbf{Details of \robotD dataset.}
    \item \cref{sec:supp-results}. \textbf{Additional results.}  
    Additional quantitative and qualitative results on \sapiensD and \robotD datasets, and results beyond 1 DOF joints.
    
\end{itemize}

\section{Implementation details}
\label{sec:supp-implementation}

\subsection{Flow Prediction Network}
The flow $\mathbf{F}^{t}$ is computed between pairs of frames $\bP^t$ and $\bP^{t-1}$ and used in the flow energy term in Sec. 3.2, where $\mathbf{F}^{t} = F(\mathbf{P}^{t})$. To ensure good generalization, we first establish correspondence between input frames $\bP^t$ and $\bP^{t-1}$ via a correspondence network and induce the flow by correspondence. 

Built upon PointNet++~\cite{qi2017pointnet++} MSG segmentation network, the correspondence network takes point cloud $\bP \in \bbR^{N \times 3}$ as input and outputs a point-wise feature map $\mathbf{f}(\bP)$. Given two features map $\mathbf{f}(\bP^{t-1})$ and $\mathbf{f}(\bP^t) \in \bbR^{N \times d}$, where $d=64$ is the feature dimension. 
We compute matching score matrix $\mathbf{S}(\bP) \in \bbR^{N \times N}$: 
\begin{align*}
   \mathbf{S}(\bP) = \text{softmax}(\frac{1}{\sqrt{d}}\mathbf{f}(\bP^{t-1}) \cdot {\mathbf{f}(\bP^t)}^T)
\end{align*}
Each column of $\mathbf{S(\bP)}$ represents the probability of matching point $\mathbf{x}^t \in \bP^t$ into a point in $ \bP^{t-1}$. At inference, 
Given the query $\mathbf{x}^t$, we find the matching point  $\mathbf{x}_{match}^{t-1} \in \mathbf{P}^{t-1}$ by taking the argmax position in corresponding column of  $\mathbf{S}(\bP)$. Then the induced flow at location $\mathbf{x}^t$ is given by $F(\mathbf{x}^t) = \mathbf{x}^t - \mathbf{x}_{match}^{t-1}$. To ensure the induced flow achieves high quality, we filter out spurious correspondences by applying mutual
nearest neighbors (MNN) criteria, which guarantees the match falls into each other's nearest neighbor. 

We train the correspondence network by minimizing the the contrastive cross-entropy loss~\cite{zbontar2015computing, luo2016efficient}. Each point in $\mathbf{P}^{t-1}$ is treated as one class, and the ground truth label is computed as the nearest neighbor of a point $\mathbf{x}^t$ in $\mathbf{P}^{t-1}$ when $\mathbf{P}^{t}$ and $\mathbf{P}^{t-1}$ are aligned. We train the correspondence network under cross-entropy loss between predicted scores $\mathbf{S}(\bP)$ and ground-truth labels $\mathbf{S}^{*}(\bP)$:
\begin{align*}
   \mathcal{L}_{\text{corr}} = -\sum_{j=1}^{N}\mathbf{S}^{*}(\mathbf{P}_j)log\mathbf{S}(\mathbf{P}_j)
\end{align*}

\subsection{Projecting to the Kinematic Model}
The projection from estimated relaxed model to the valid kinematic model is achieved by minimizing the cost over $E_\text{project}$, which consists of a spatial term $E_\text{spatial}$ and the 1-DOF motion term $E_\text{1-DOF}$, we explain each term in details.

\paragraph{$E_\text{spatial}$.} 
If two parts are linked, they should be close in 3D space.  The $E_\text{spatial}$ measures the spatial proximity of the parent-child pair $\text{pa}(i)$ and $i$ in canonical frame $\bP^c$. We query the part segmentation field $f$ and extract the corresponding part segmentation points of parent and child $\bp_{\text{pa}(i)} = \{ \bx \in \bP^c | f(\bx) = {\text{pa}(i)}\}$ and $\bp_i = \{ \bx \in \bP^c | f(\bx) = i\}$. The $E_\text{spatial}(i, \text{pa}(i)) = \min_{\bx \in \bp_i} \min_{\by \in \bp_{\text{pa}(i)}} \| \bx - \by \|_2^2$. To improve efficiency, we do farthest point sampling from $\bp_{\text{pa}(i)}$ and $\text{pa}(i)$ and sample $20$ points per part to represent the set. We compute the $E_\text{spatial}(i, \text{pa}(i))$ for all part pairs in parallel. 

\paragraph{$E_\text{1-DOF}$.} 
In articulated objects, if two parts are linked, their relative transformation should be explained by a 1-DOF screw joint. The $E_\text{1-DOF}$ in Eq. (12) computed the approximation error for the temporal sequence of relative transformation between parent $\text{pa}(i)$ and child $i$ treating as a 1DOF transformation. The relative transformation sequence is computed as $\{\hat{\bT}_{pa(i)}^t \ominus \hat{\bT}_i^t\}_{t \in 1,...T}$ between parent $\text{pa}(i)$. We compute the approximated screw parameters $\bs_i, \{\btheta^t\}$ by the following objective:
\begin{align*}
   \bs_i, \{\btheta^t\} = \argmin_{\bs_i, \{\btheta^t\}}\left(\sum_t 
\text{trace}((\hat{\bT}_{pa(i)}^t \ominus \hat{\bT}_i^t ) \ominus \bT(\bs_i, \btheta_i^t))
\right).
\end{align*}
We solve the above for all part pairs $i$ and $\text{pa}(i)$. The residual error is taken as $E_\text{1-DOF}(i, \text{pa}(i))$.

\subsection{Merging.} 
To make the kinematic topology compact, we merge parts that are close in space with small relative motion. The static joint is a special case of the 1-DOF screw joint, where the rotation and translation component both equal to 0. Similar to $E_\text{1-DOF}$, we define $E_\text{merge}(i, \text{pa}(i)) = \sum_t 
\text{trace}((\hat{\bT}_{pa(i)}^t \ominus \hat{\bT}_i^t ) \ominus \textbf{I}$, where $\textbf{I}$ is the identity matrix. We merge pair $\text{pa}(i)$ and $i$ if $E_\text{merge}(i, \text{pa}(i)) < \epsilon_{m}$, meaning their relative motion is small. The merging is done iteratively. We start from the part pair $\text{pa}(i)$ and $i$ with the lowest $E_\text{spatial}$ and stop merging until all remained part pairs have $E_\text{merge} \geq \epsilon_{m}$. In \cref{fig:supp-merge}, we show segmentation of real-world switch before and after merging step.

\subsection{Final Fitting.} After projection and merging, we obtain a  valid kinematic model $\tree, {\{\bs_i\}, \{\btheta^t\}}$, we infer the joint type (revolute or prismatic) for part $i$ and its parents by check $\{\btheta^t_i\}_{t \in 1,...T}$, where $\btheta^t_i = (\tau^t_i, d^t_i)$. As dicussed in Sec. 3.1, the rotation angle of a prismatic joint is always zero, \ie $\{\tau^t_i=0\}_{t \in 1,...T}$, while the  translational component always be 0 for a revolute joint, \ie $\{d^t_i=0\}_{t \in 1,...T}$. We compute the mean  $\bar{\tau_i} = \sum_{t=1}^{T}\tau^t_i$ and $\bar{d_i} = \sum_{t=1}^{T}d^t_i$ for part $i$. If $\bar{\tau_i} < \bar{d_i}$, we treat the joint between $i$ and its parent $\text{pa}_{i}$ as a prismatic joint, otherwise as a revolute joint. In final fitting stage, we ensure all the joints fall into these two classes and keep $\{\tau^t_i=0\}_{t \in 1,...T}$ for prismatic joint and $\{d^t_i=0\}_{t \in 1,...T}$ for revolute joint during optimization.

\subsection{Canonical Frame Selection.} Our algorithm is flexible in taking arbitrary frames in the input sequence as the canonical frame $c$. Certain frames could make part segmentation field more easily separating different parts, \eg if two rigid parts undergo some similar motion throughout the entire sequence, certain frames could better capture those subtle  differences and gives better segmentation result. Thus, we develop a criteria for selecting best canonical frame within the input sequence. We pick the canonical frame by selecting the one with lowest $E_\text{project}$ + $E_\text{group}$. $E_\text{project}$ is the same defined in Eq. (11). $E_\text{group}$ is used to measure the deviation of each cluster in the segmentation field. For each part $i \in [1\ldots n]$, point segmentation cluster $\bp_i = \{ \bx \in \bP^c | f(\bx) = i\}$, we compute the cluster center $\mathbf{c}_i = \frac{1}{|\bp_i|}\sum_{\bx \in \bp_i}\bx$, the $E_\text{group}$ is computed as:

\begin{align*}
   E_\text{group} = \frac{1}{n}\sum_{i=1}^{n} \frac{1}{|\bp_i|}\sum_{\bx \in \bp_i}(\bx -\mathbf{c}_i)^2
\end{align*}

\subsection{Optimization Details}
We set $\lambda_\text{CD}=1.0$, $\lambda_\text{EMD}=0.3$, and $\lambda_\text{flow}=1.0$ for $E_\text{recons}$ in Eq. (5). 
Those parameters are tuned on validation set and fixed for all testing samples. We use $\text{knn}=3$ for flow trilinear interpolation. We set $\lambda_\text{spatial}=100$ and $\lambda_\text{1-DOF}=1.0$ in Eq. (11), merging threshold $\epsilon_{m}=3e-2$. In relaxed model estimation stage, 
We optimize the model for 15,000 iterations, $E_\text{EMD}$ is applied on 4$\times$ downsampled point clouds and updated every 5 iterations. We use  a cosine annealing schedule anneal for Gumbel-softmax temperature. It start from 5.0 and decay 1.0 in the last iteration. In final optimization stage, we optimize the model for 200 iterations, $E_\text{EMD}$ is applied on 2$\times$ downsampled point clouds and updated every iteration.

\subsection{Rearticulation}
\label{sec:supp-rearticulation}
We can re-articulate our predicted model to a given target pose by only given a sparse set of point locations (Figure. 1). Given the source points, We use the part segmentation field to infer part labels and use forward kinematics in Eq. (2) of the model $M(\btheta^t; \tree, f)$ to deform those points to match target points. We fix $\tree, f$ and only optimize the joint state parameters $\btheta$ for 200 iterations with learning rate $0.1$. We optimize the MSE loss between the deformed points and provided target points $\bP'$. 
\begin{align*}
\btheta = \argmin_{\btheta} \mathcal{L}_{mse}\left(M(\btheta; \tree, f), \bP'\right)
\end{align*}

\subsection{Implementation of the Baselines}
We describe the implementation of baselines  MultiBodySync~\cite{huang2021multibodysync} and WatchItMove~\cite{noguchi2022watch}.

\paragraph{MultiBodySync.} 
MultiBodySync synchronizes between all $\binom{T}{2}$ states in the input sequence of length $T$ and iteratively refinem the motion prediction and segmentation. The method requires the eigen-decomposition of a Laplacian matrix with size $\bbR^{NT \times NT}$, N is the number of points at each state. When input sequence becomes long, the matrix computation becomes the bottleneck of the method and could very hard to fit into the memory. To this end, we 2$\times$ downsampled input point clouds to 2048 points as input. MultiBodySync estimates the number of parts by analysing the spectrum of predicted motion segmentation matrices and counting the number of eigenvalues larger than a cutting threshold. We found out this strategy works well on \sapiensD dataset, but performs poorly on \robotD dataset given the more complicated part motions controlled by the kinematic tree. 
We choosing the cutting threshold among $[0.05, 0.005, 0.001]$ and choose the best one which is $0.001$ on the validation set. We also increase the number of iterations from $4$ to $6$ for better iterative refinement. However, we found out the method still severely suffer from missing parts and wrong motion prediction as shown in \cref{fig:supp-comparison}. The method requires pairwise flow prediction, this could be extremely challenging in robot case with large deformations between the start and the end of a long sequence.

\paragraph{WatchItMove.} 

The original WatchItMove takes as input multi-view RGB videos with strong cues on both geometry and appearance. To apply WatchItMove to our setting with the 4D point cloud, we adjust their $\text{implementation}$ \footnote{https://github.com/NVlabs/watch-it-move} with two major changes: 1) Replace the photometric reconstruction loss with SDF $\mathcal{L}_1$ loss, where the label comes from ground-truth signed distance field; 2) We use the ground-truth \# of rigid components. Both changes give certain levels of advantage to WatchItMove. However, the results demonstrated that motion cue is indispensable. Without motion cues, there is no constraint to regularize the ellipsoids motion. Though the overall shape could match to the input and SDF loss could be minimized, those ellipsoids could move with random motion internally. The result also justify the importance of hard assignment of points to segments during training. Instead of using hard assignment, \wim uses soft assignment during training. The motion of a certain point is blended by all ellipsoid motions. At inference, we require each point follow one corresponding ellipsoid motion by taking the argmax of segmentation weights from all parts. The inconsistency between training and testing hinders the motion estimation performance. We also note that it is crucial to incorporate the 1-DOF constraint when building kinematic tree. Without considering the  constraint could result in unmeaningful linkage as shown in \cref{fig:supp-comparison}.

\subsection{Evaluation Metrics}
We discuss the reconstruction metrics, intermediate metrics and reanimataion metric in more details. 

\paragraph{Reconstruction Metrics.} We reconstruct the input sequence using our built animatable model $M(\btheta^t; \tree, f)$. We measure the per-point reconstruction error across all time steps $T$. The flow is computed between the canonical frame and all reconstructions in the sequence. For flow accuracy, we set the treshold $\delta=0.005$ on \robotD dataset and $\delta=0.05$ on \sapiensD dataset.

\paragraph{Intermediate Metrics.} The tree edit distance is the minimal-cost sequence of node edit operations to turn the predicted tree into ground-truth. The three allowed operations are delete, insert and rename. Follow~\cite{xu2020rignet}, we set rename cost to be 0 and all other two operations cost to be 1. Given the predicted kinematic tree is undirected, we traverse all possible orders of the tree and select the minimum one as the final metric.

\paragraph{Reanimataion Metric.} Given the ground-truth point cloud in a novel frame, we sample one pair of correspondence per-part between the canonical frame and novel frame, which guarantees the novel part poses is impossible to recovered from ICP~\cite{besl1992method} or Kabsch~\cite{kabsch1976solution} algorithm. We use the provided sparse correspondences and algorithm described in~\cref{sec:supp-rearticulation} to deform the canonical frame into novel frame, and measures the per-point error against the ground truth. 

\begin{table}[t]
    \centering
    \small
    {
    \setlength{\tabcolsep}{0.1em} %
    \begin{tabular}{ccc}
        \includegraphics[trim={0cm 0cm 0cm 0cm},clip,width=0.3\linewidth]{figures/real/real_switch.png} &
        \includegraphics[trim={10cm 10cm 10cm 10cm},clip,width=0.3\linewidth]{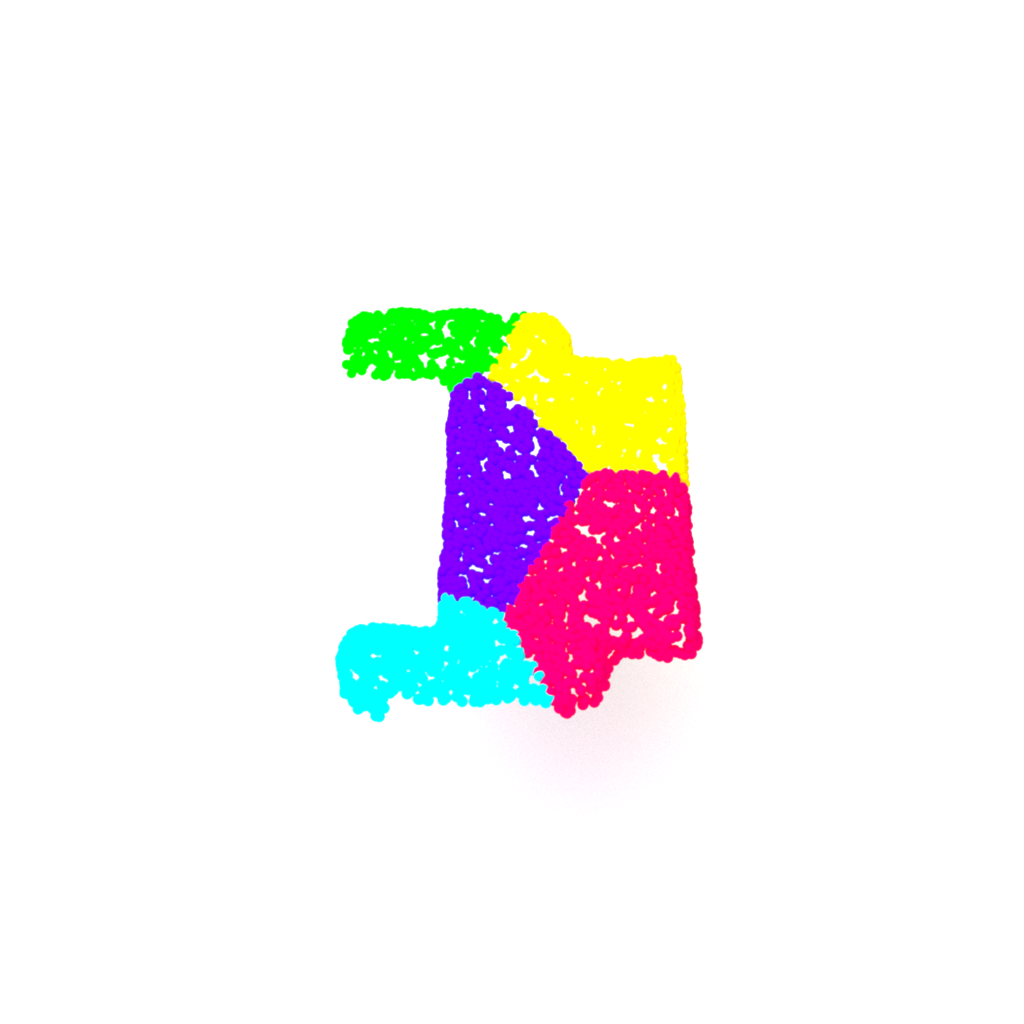} &
        \includegraphics[trim={10cm 10cm 10cm 10cm},clip,width=0.3\linewidth]{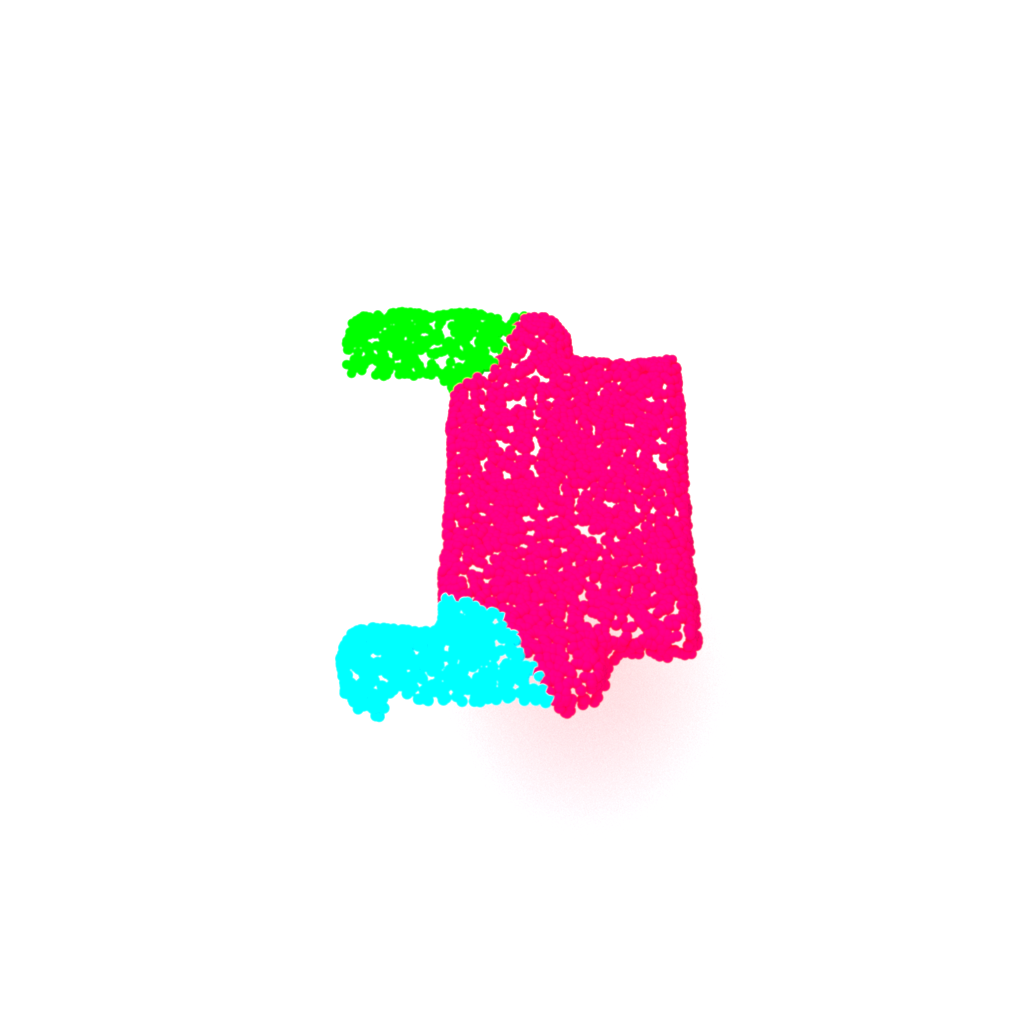}\\
        (a) Real scan &
        (b) Before merging &
        (c) After merging \\
    \end{tabular}
    }
    \captionof{figure}{\textbf{Visualization of merging.} We show the segmentation of real-world switch before and after merging step.}
    \label{fig:supp-merge}
\end{table}

\begin{table}[t]
    \centering
    \small
    \resizebox{\linewidth}{!}
    {
    \setlength{\tabcolsep}{0.1em} %
    \begin{tabular}{cccc}
        \imgtile{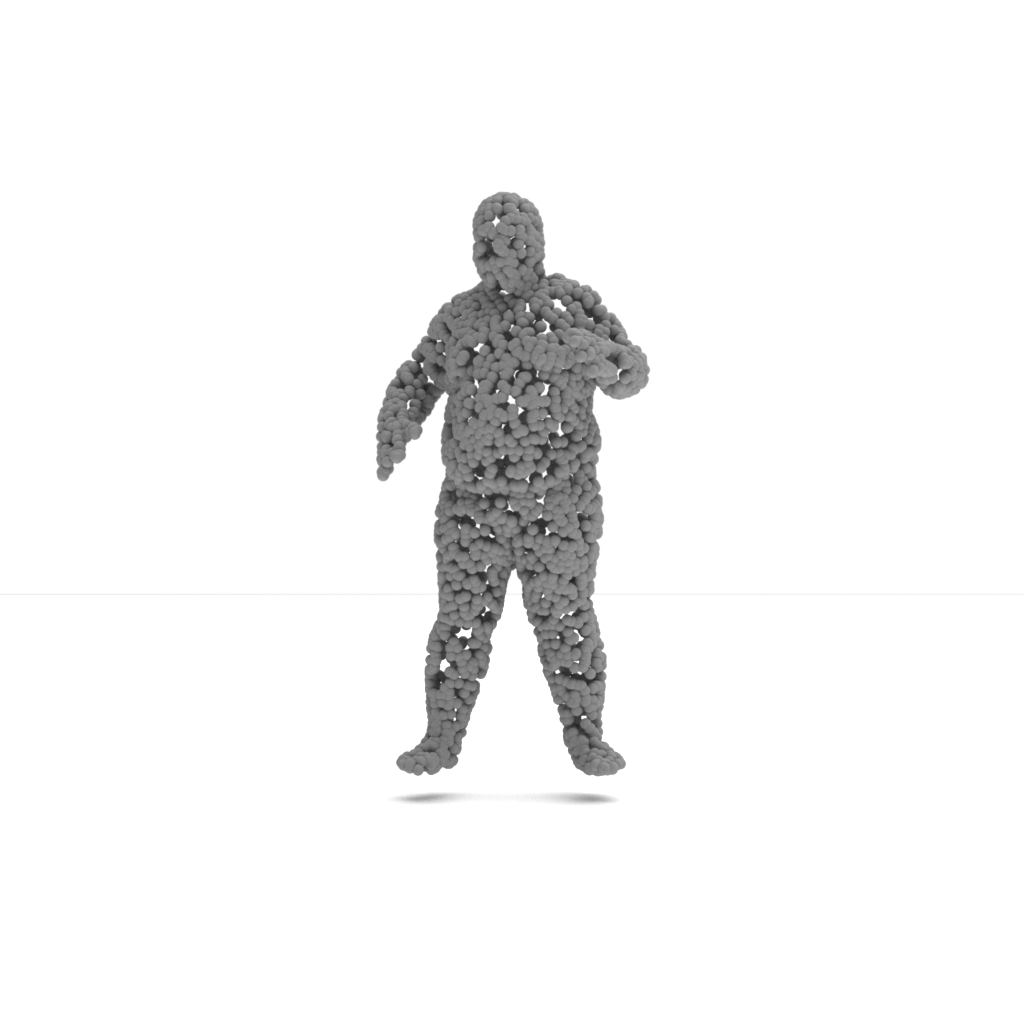}{0.25} & 
        \includegraphics[trim={7cm 7cm 7cm 7cm},clip, width=0.22\linewidth]{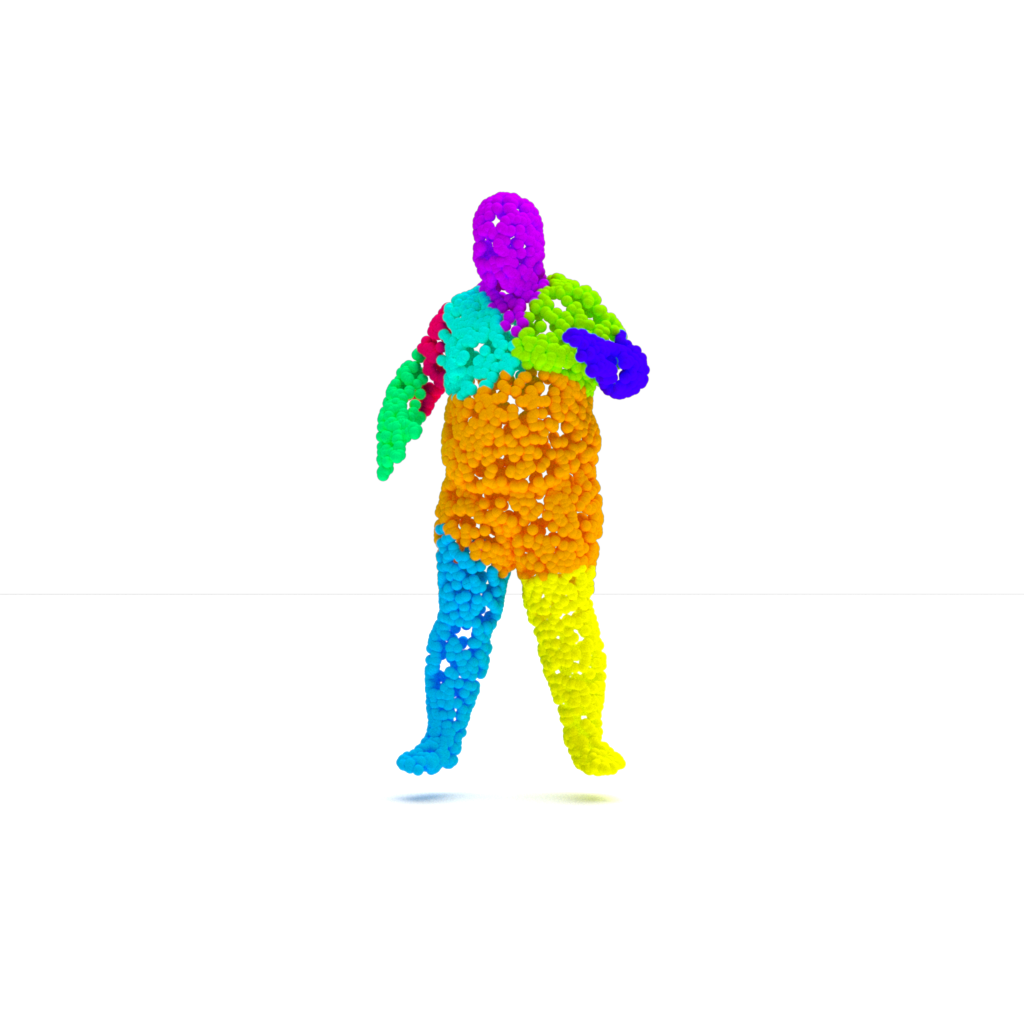} & 
        \includegraphics[trim={7cm 7cm 7cm 7cm},clip, width=0.22\linewidth]{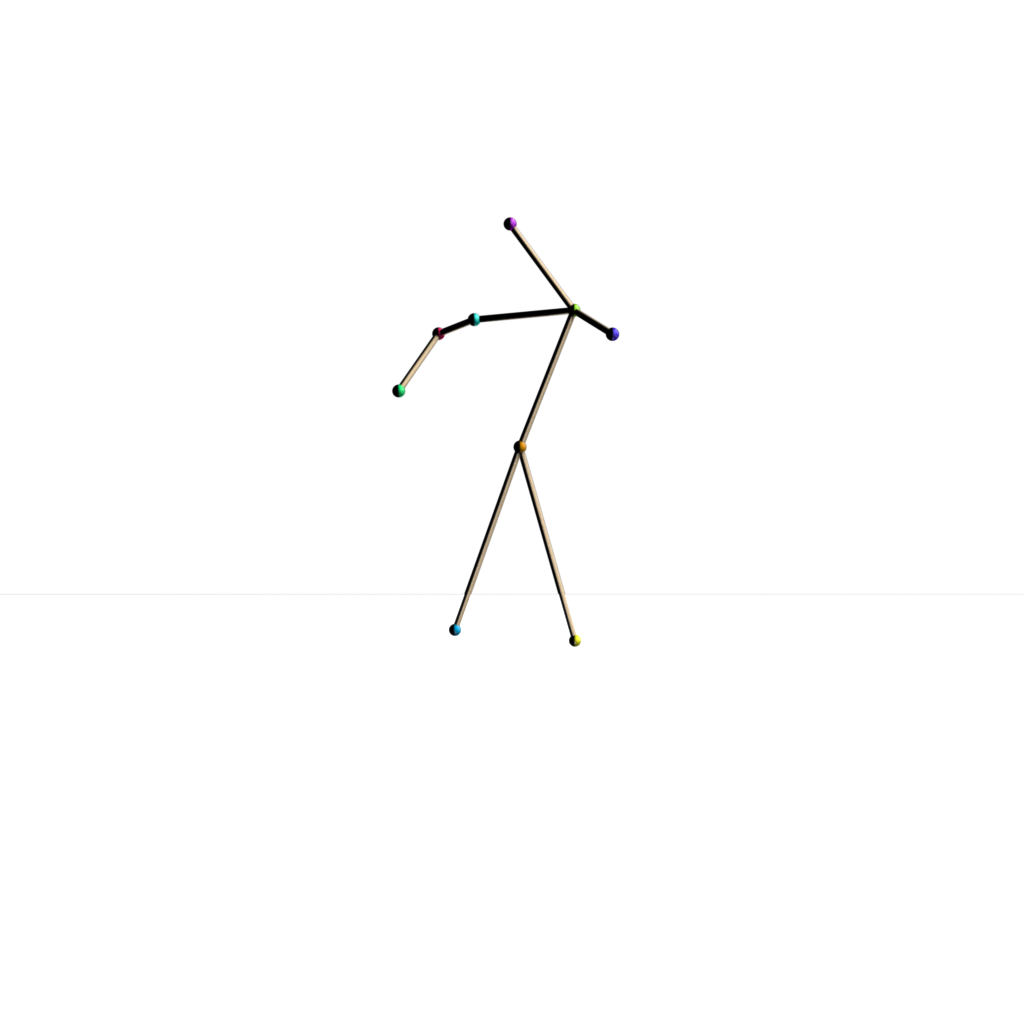} & \includegraphics[trim={7cm 7cm 7cm 7cm},clip, width=0.22\linewidth]{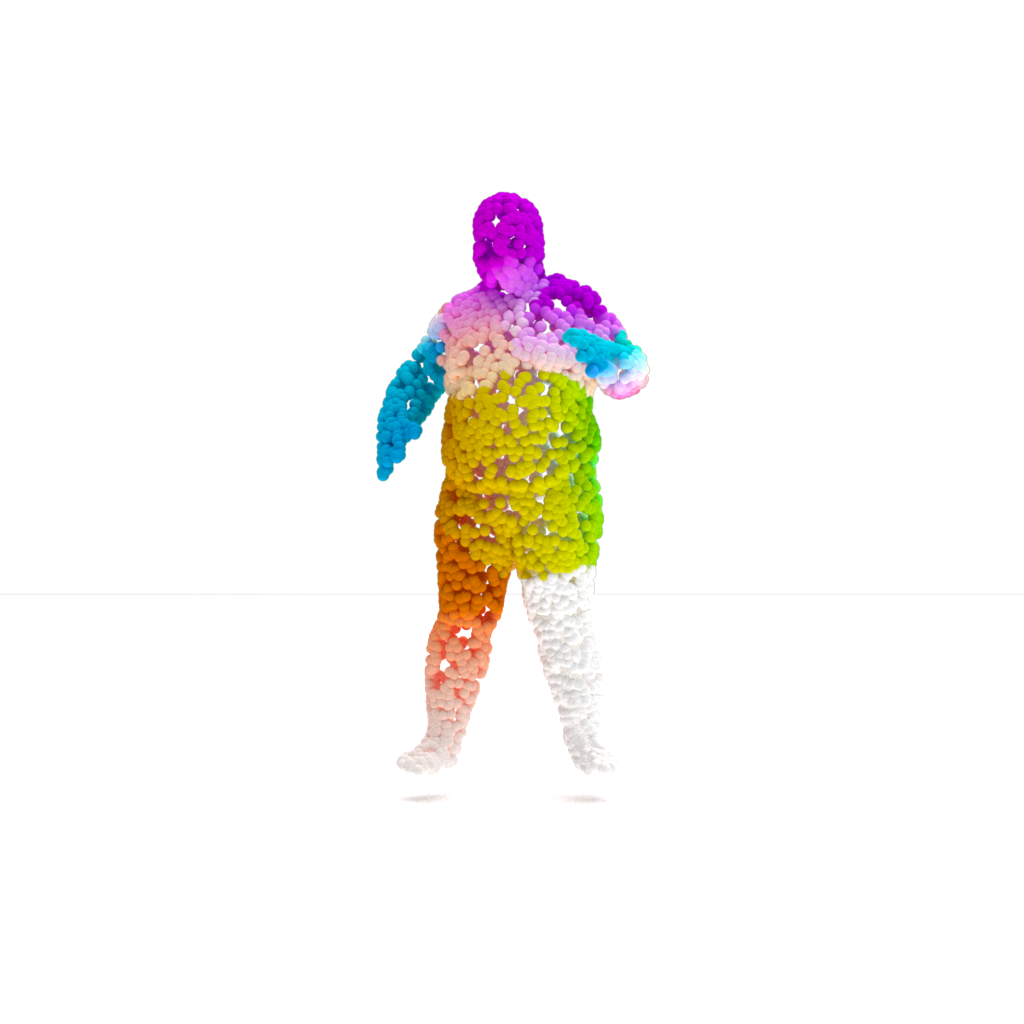} \\
        (a) Input &
        (b) Segmentation &
        (c) Topology &
        (d) Flow \\
    \end{tabular}
    }
\captionof{figure}{\textbf{Qualitative results beyond 1DOF joints.} We apply our method to model spherical joints of human on D-FAUST dataset~\cite{bogo2017dynamic}. From left to right, we show the input human point cloud sequence, part segmentation, part connectivity, and implied flow.  This demonstrates that our framework is general and can tackle other joint types beyond 1DOF joints.}
\label{fig:supp-human}
\end{table}

\begin{table*}[t]
    \centering
    \footnotesize
     \caption{{\bf Per-category performance on \sapiensD dataset.} We report \red{flow error} $\downarrow$ and \blue{Multi-scan RI} $\uparrow$.}
    \label{tab:supp-sapien-cat}
    \setlength{\tabcolsep}{2pt}
    \resizebox{\linewidth}{!}
    {
        \begin{tabular}{cccccccccc} 
            \toprule
            Box & Dishwasher & Display & Furniture & Eyeglasses & Faucet & Kettle & Knife & Laptop & Lighter \\
            \red{6.4}/\blue{0.84} & \red{6.7}/\blue{0.82} & \red{3.6}/\blue{0.68} & \red{4.2}/\blue{0.84} & \red{2.9}/\blue{0.85} & \red{2.9}/\blue{0.71} & \red{5.5}/\blue{0.76} & \red{4.2}/\blue{0.72} & \red{5.7}/\blue{0.79} & \red{3.0}/\blue{0.88} \\ 
            Oven & Phone & Washer & Pliers & Safe & Stapler & Door & Toilet & TrashCan & Microwave \\
            \red{7.0}/\blue{0.75} & \red{3.5}/\blue{0.66} & \red{3.6}/\blue{0.76} &\red{2.19}/\blue{0.77} & \red{3.7}/\blue{0.84} & \red{7.5}/\blue{0.78} & \red{2.9}/\blue{0.74} & \red{3.0}/\blue{0.77} & \red{6.5}/\blue{0.82}& \red{5.8}/\blue{0.83}\\
            \bottomrule
        \end{tabular}
    }
\end{table*}

\begin{table*}[t]
    \parbox[t]{.48\linewidth}{
    \centering
    \footnotesize
    \caption{{\bf Robot type and categories on \robotD dataset.}}
    \label{tab:robot-type}
    \setlength{\tabcolsep}{2em}
    \resizebox{\linewidth}{!}{
    \begin{tabular}{lc}
        \toprule
         Robot Type & Robot Categories \\
         \midrule
         Arms & Panda, UR5, Baxter, Kinova, iiwa \\
         Bipeds & Bolt, Cassie\\
         Hands & Allegro, Barrett \\
         Mobile Manipulators & Reachy \\
         Humanoids & Nao, Atlas, iCub, JVRC \\
         Quadrupeds & A1, Laikago, Solo, Spot\\
         \bottomrule
    \end{tabular}}
    }
    \hfill
    \parbox[t]{.48\linewidth}{
        \centering
    \footnotesize
    \caption{{\bf \robotD dataset train, validation, and test split.}}
    \label{tab:robot-split}
    \setlength{\tabcolsep}{2em}
    \resizebox{\linewidth}{!}{
    \begin{tabular}{lc}
        \toprule
         Split & Robot Categories \\
         \midrule
         Train & Atlas, Baxter, Laikago, iiwa \\
         Validation & Panda, Cassie, Spot, Panda \\
         \multirow{2}{*}{Test} & Kinova, UR5, Bolt, Allegro, Barrett\\
          & Reachy, iCub, JVRC, A1, Solo \\
         \bottomrule
    \end{tabular}}
    }
\end{table*}

\begin{figure}[htbp]
\centering
\includegraphics[width=0.8\linewidth]{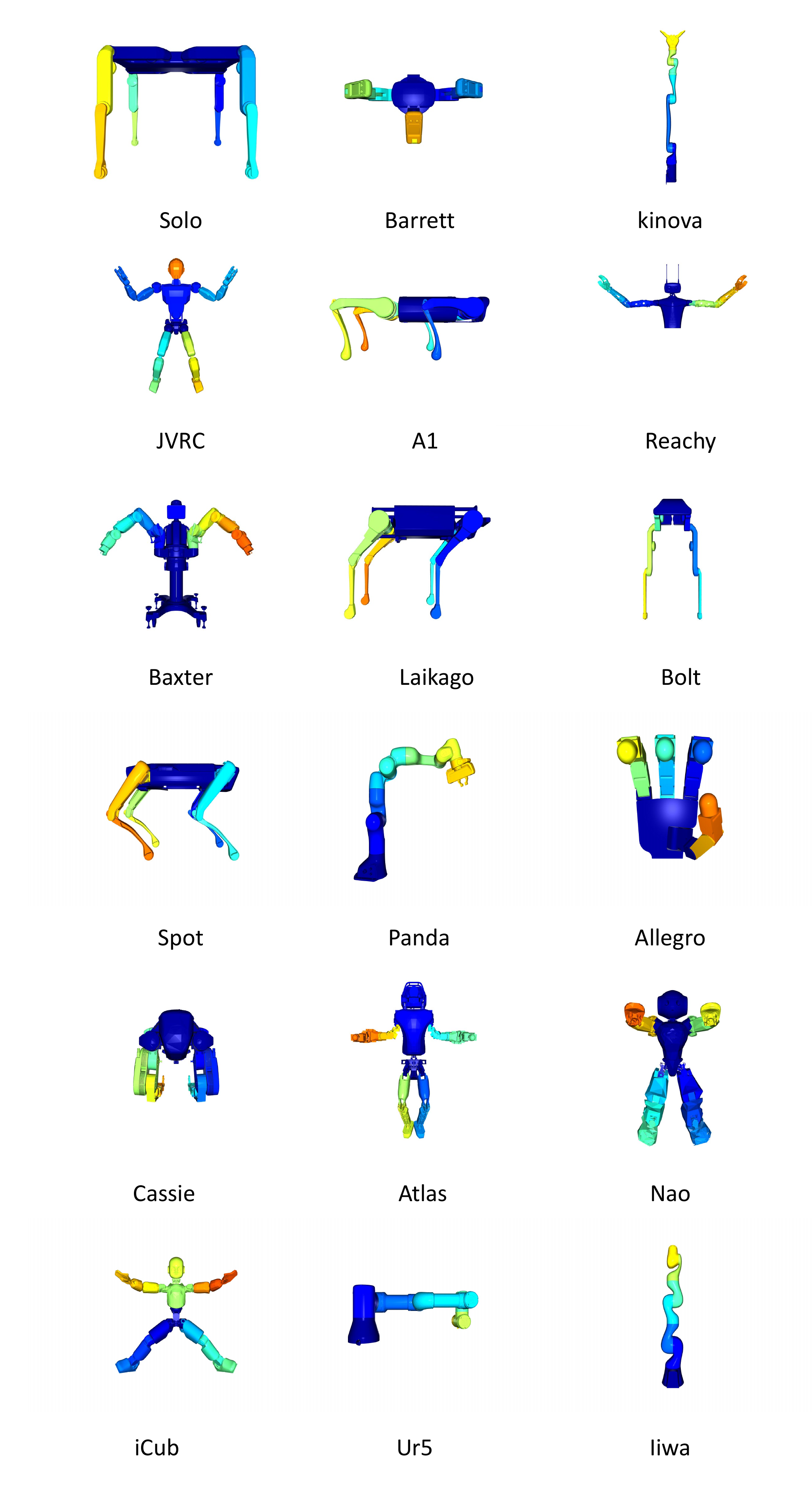}
\caption{\textbf{Robot categories visualization on \robotD dataset. Different parts are in different colors.}}
\label{fig:supp-robot-dataset}
\end{figure}
\begin{table*}[ht]
    \begin{subfigure}{.48\linewidth}
    \centering
    \small
    \resizebox{\linewidth}{!}{
    \setlength{\tabcolsep}{0.1em} %
    \begin{tabular}{cccc}
        \comparevis{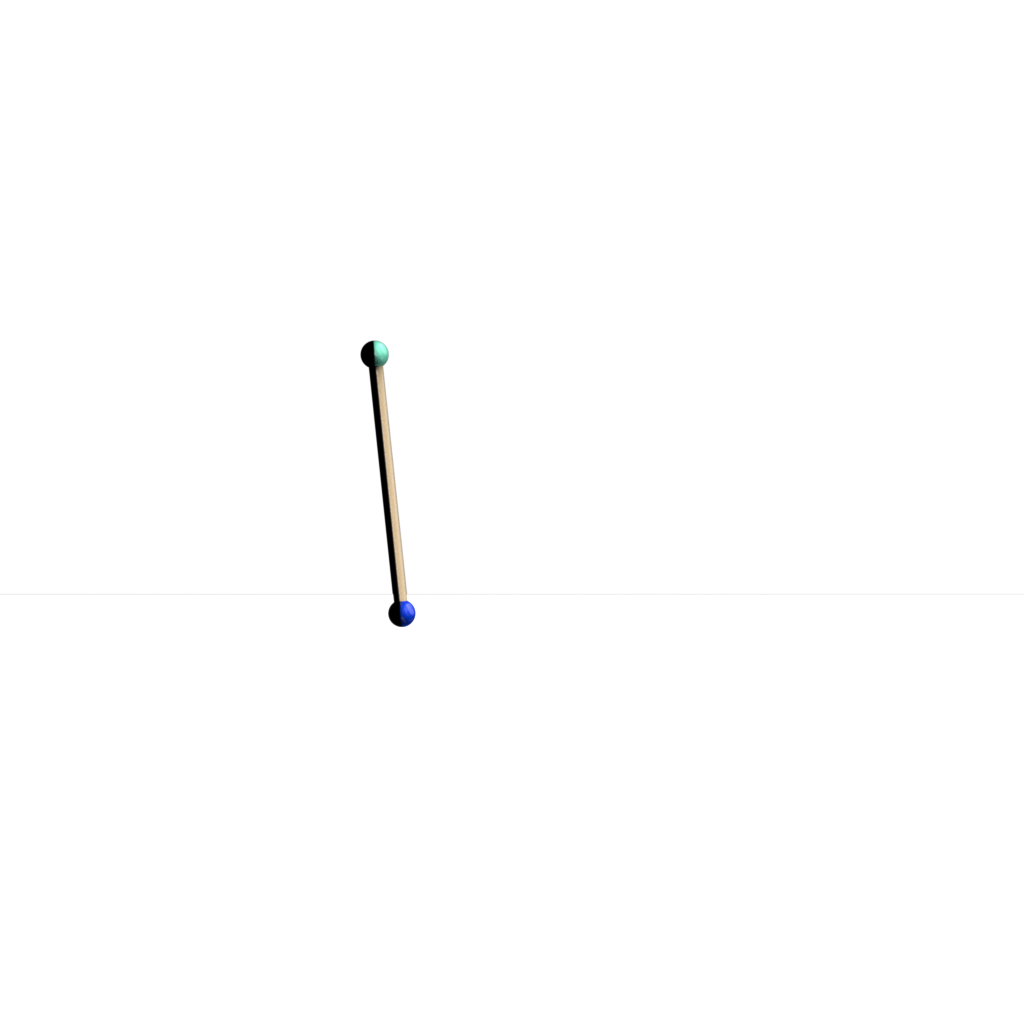}{0.3}{9.5cm 10cm 10.5cm 7cm} \\
        \comparevis{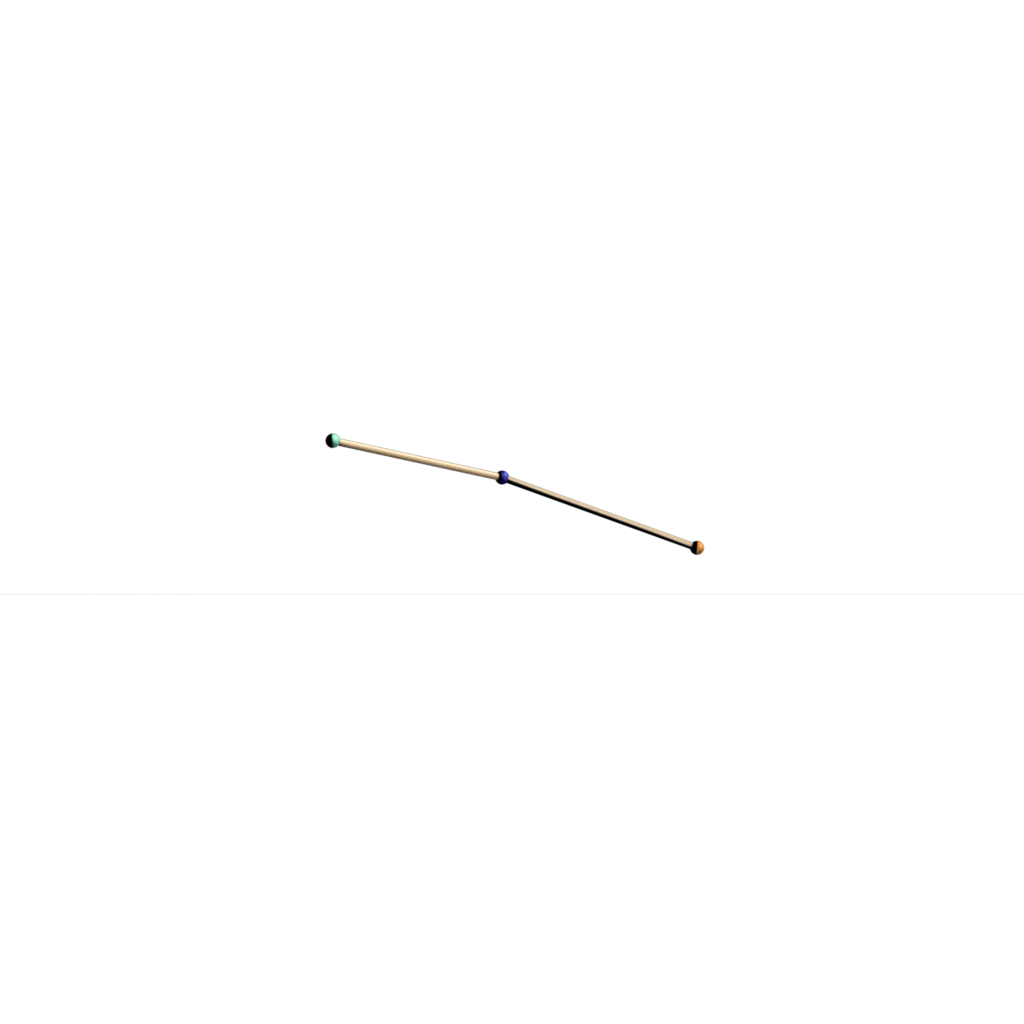}{0.3}{9cm 12cm 9cm 12cm} \\
        \comparevis{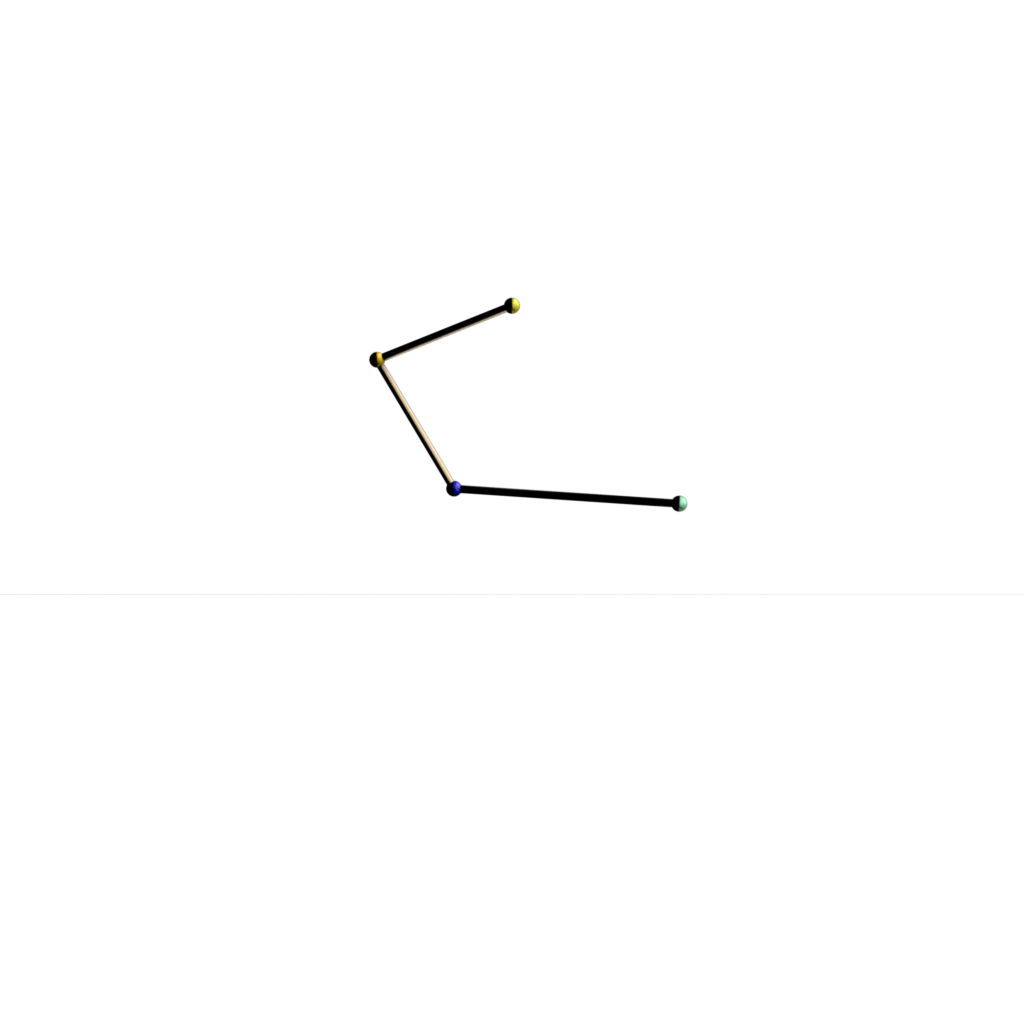}{0.3}{9.75cm 8cm 10cm 8cm} \\
        \comparevis{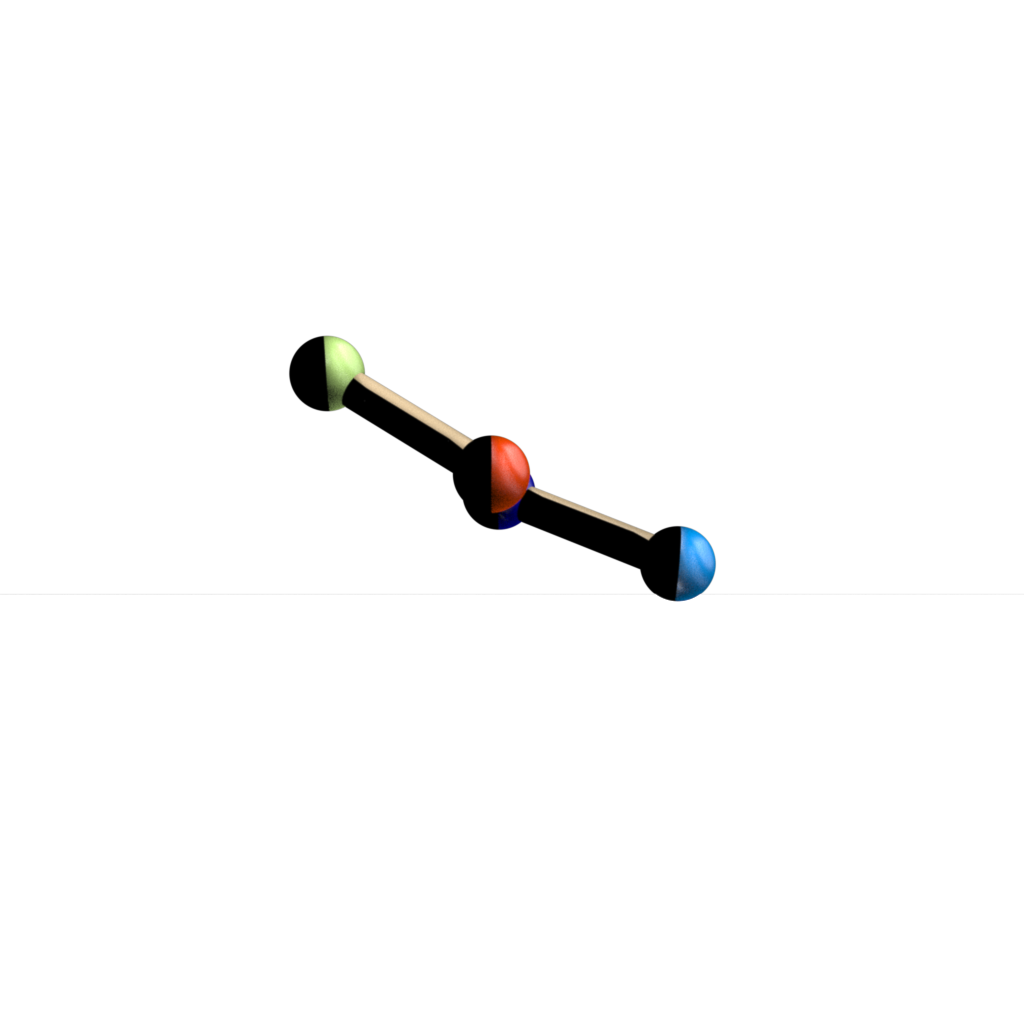}{0.3}{5cm 5cm 5cm 5cm} \\
         \mbs & \wim & Ours & GT\\
    \end{tabular}
    }
    \end{subfigure}
    \hfill\vline\hfill
    \begin{subfigure}{.48\linewidth}
    \centering
    \small
    \resizebox{\linewidth}{!}{
    \setlength{\tabcolsep}{0.1em} %
    \begin{tabular}{cccc}
        \comparevis{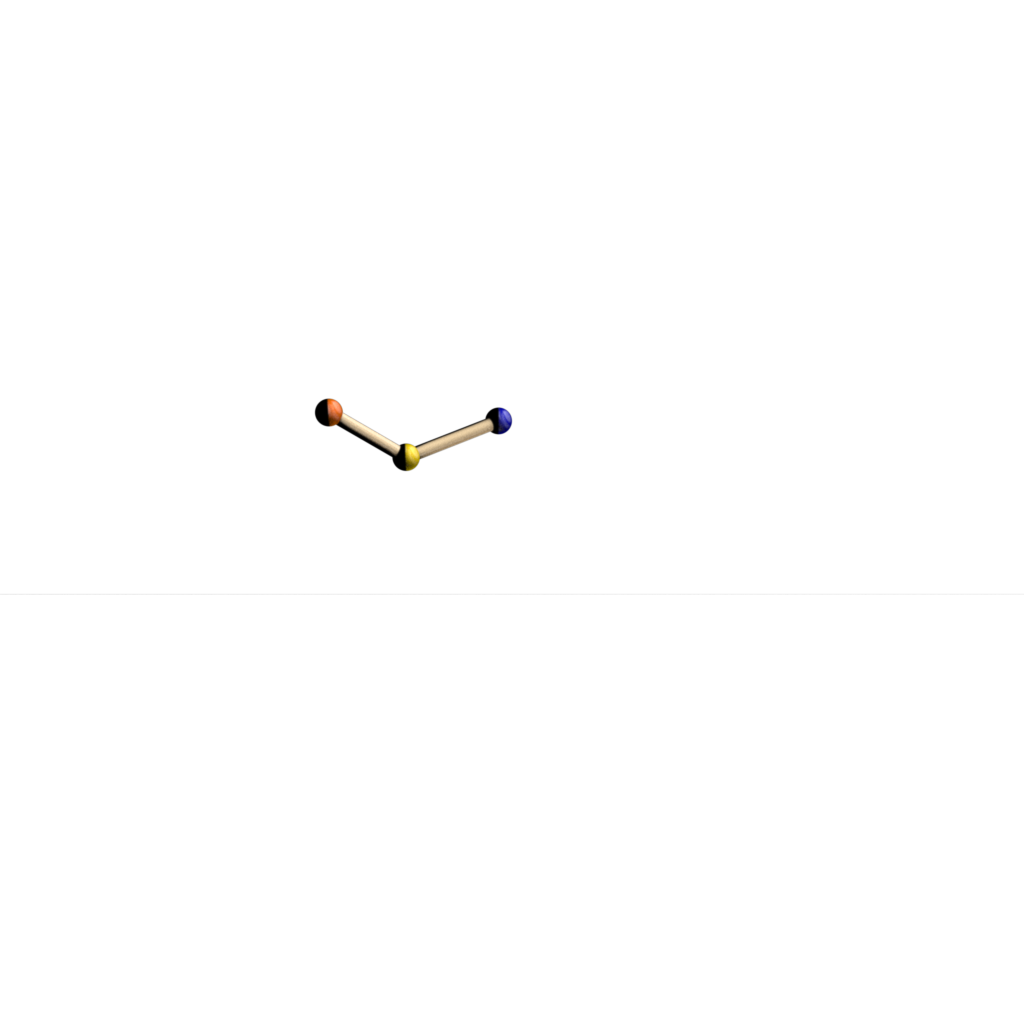}{0.3}{9.5cm 10cm 10cm 7cm} \\
        \comparevis{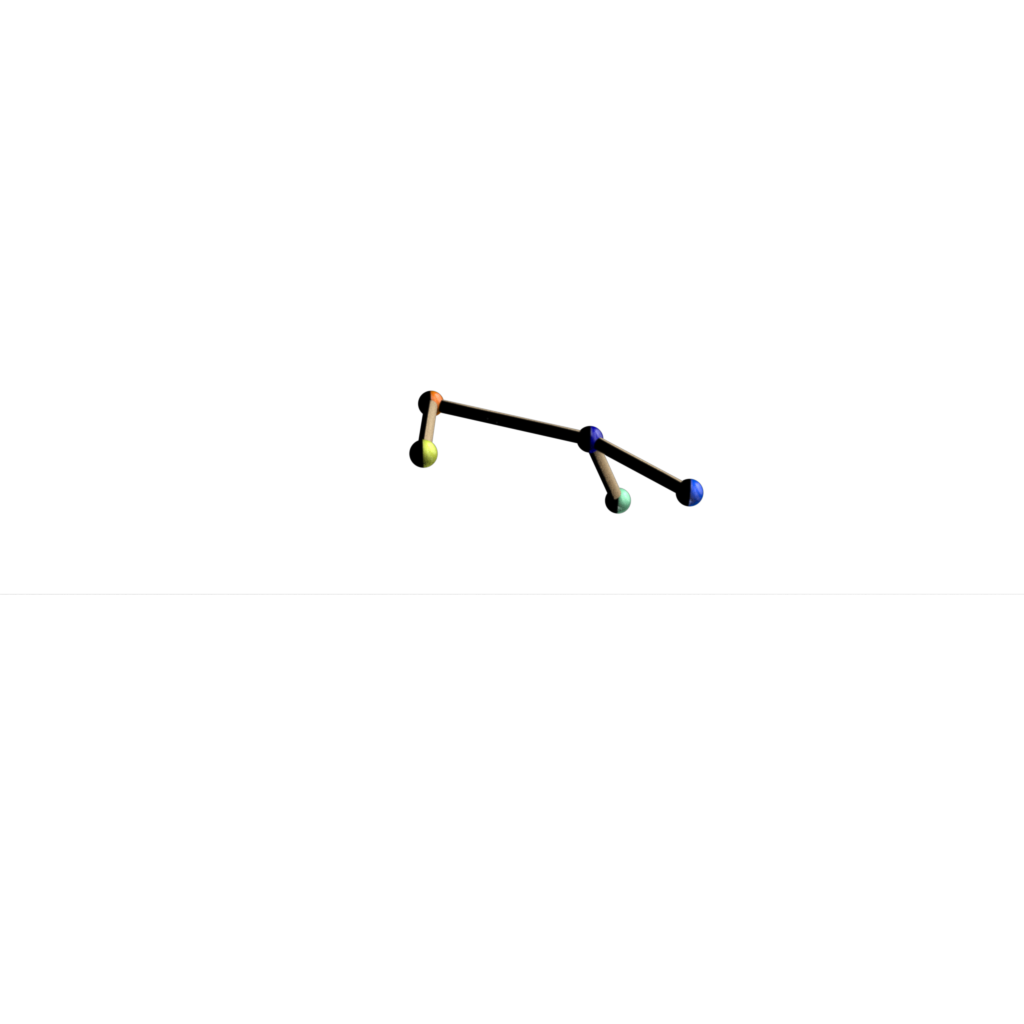}{0.3}{8cm 9cm 6cm 5cm} \\
        \comparevis{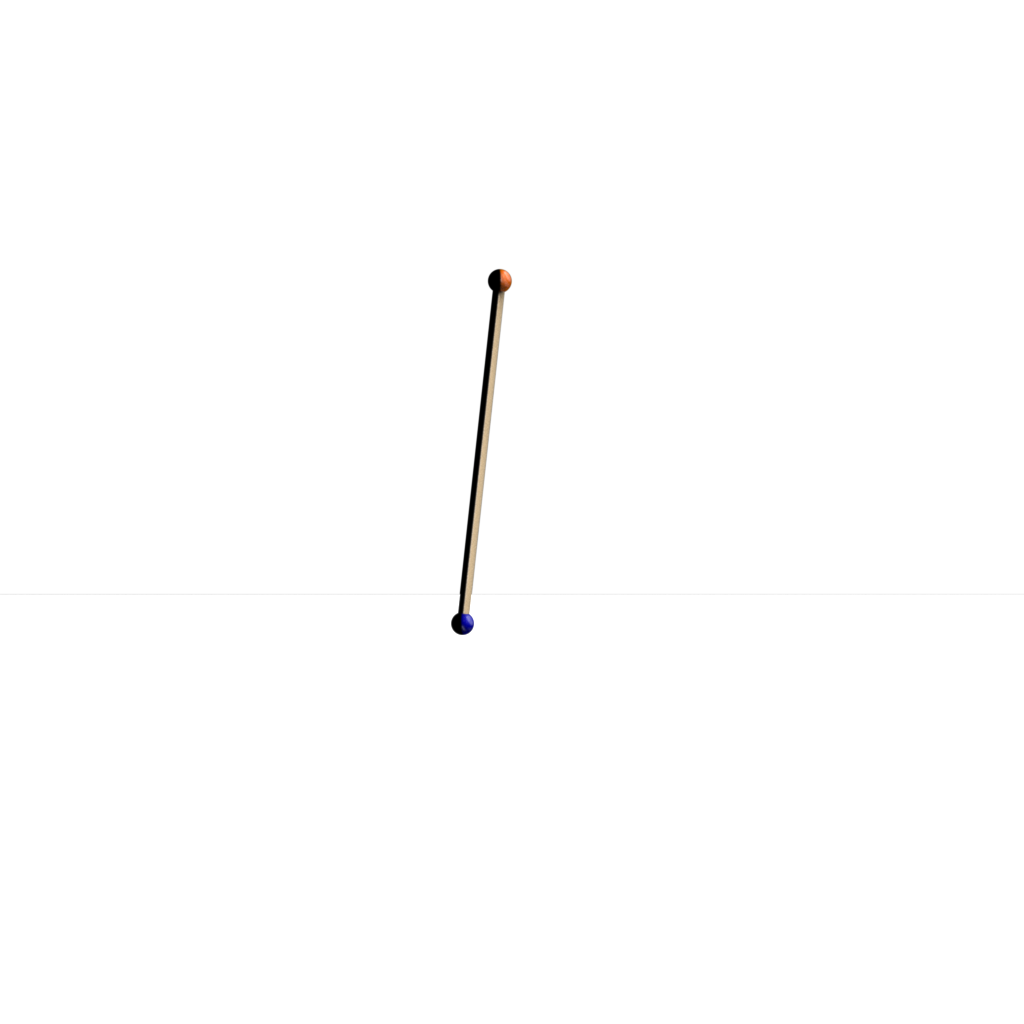}{0.3}{9.5cm 10.5cm 10.5cm 8cm} \\
        \comparevis{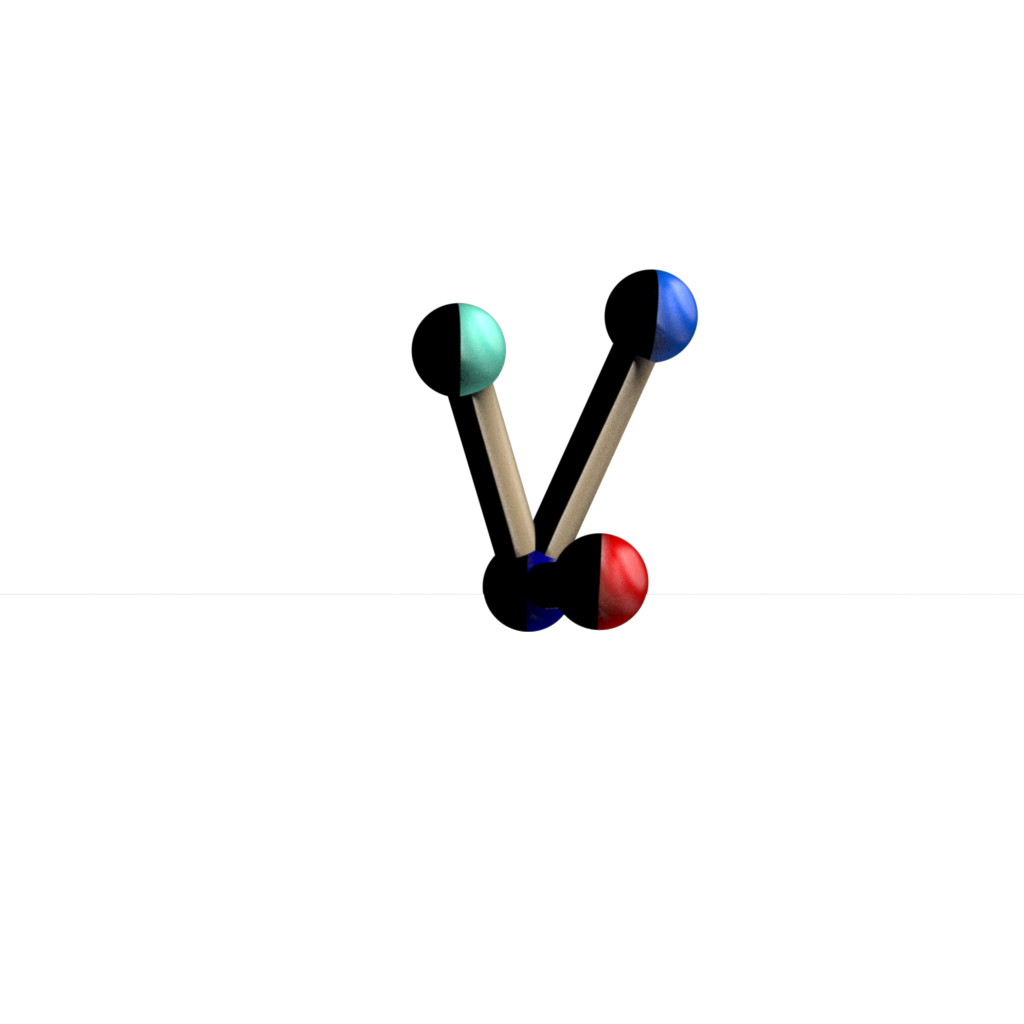}{0.3}{3cm 3cm 3cm 3cm} \\
        \mbs & \wim & Ours & GT\\
    \end{tabular}
    }
    \end{subfigure}
\captionof{figure}{\textbf{Qualitative comparison against \mbs~\cite{huang2021multibodysync} and \wim~\cite{noguchi2022watch} on the \robotD dataset test set.} Note, a) \mbs by itself doesn't produce a kinematic tree, we use our method on top of their output to generate one, and b) we provide \wim~\cite{noguchi2022watch} with the ground truth SDFs and number of parts (which are not used by our method). Even after these modifications, the past methods cannot solve the task as well as ours.}
\label{fig:supp-comparison}
\end{table*}

\begin{table*}[ht]
    \centering
    \footnotesize
    \resizebox{\linewidth}{!}{
    \setlength{\tabcolsep}{0.2em} %
    \begin{tabular}{cccc|ccc}
        \cellcolor{lgray}{Input} & 
        \multicolumn{3}{c}{\cellcolor{lred}{Predictions}} &
        \multicolumn{3}{c}{\cellcolor{lgreen}{Ground Truth}} \\
        \robotvis{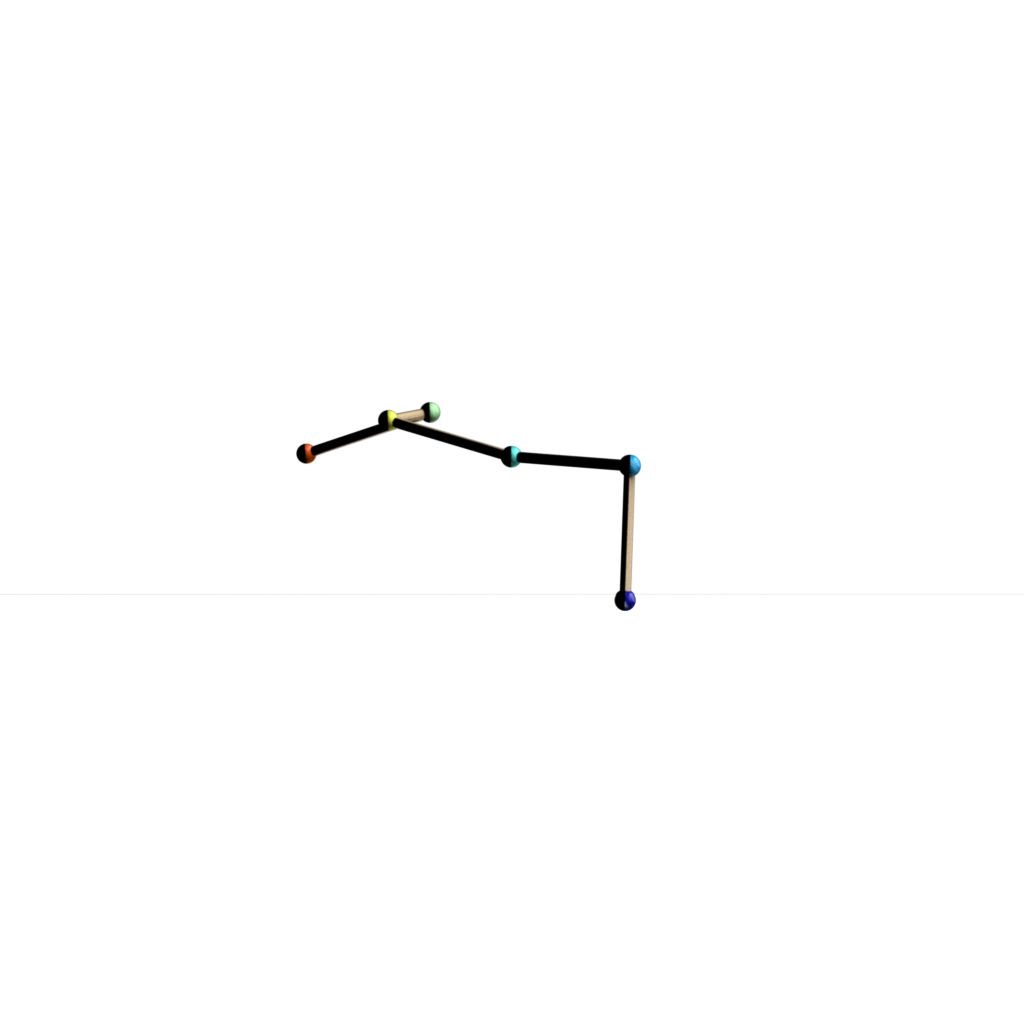}{0.15}{7cm 8cm 11cm 10cm}
        \robotvis{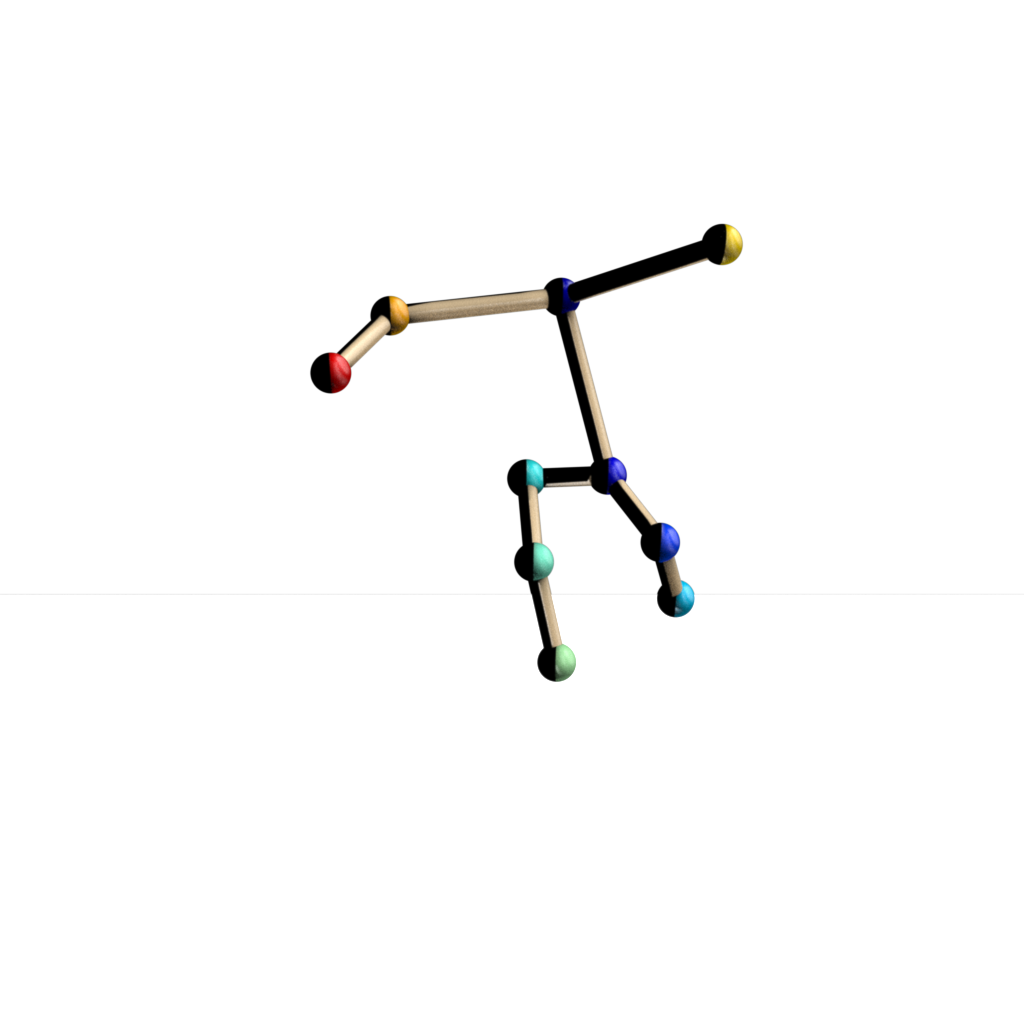}{0.15}{6cm 6cm 6cm 6cm}
        \robotvis{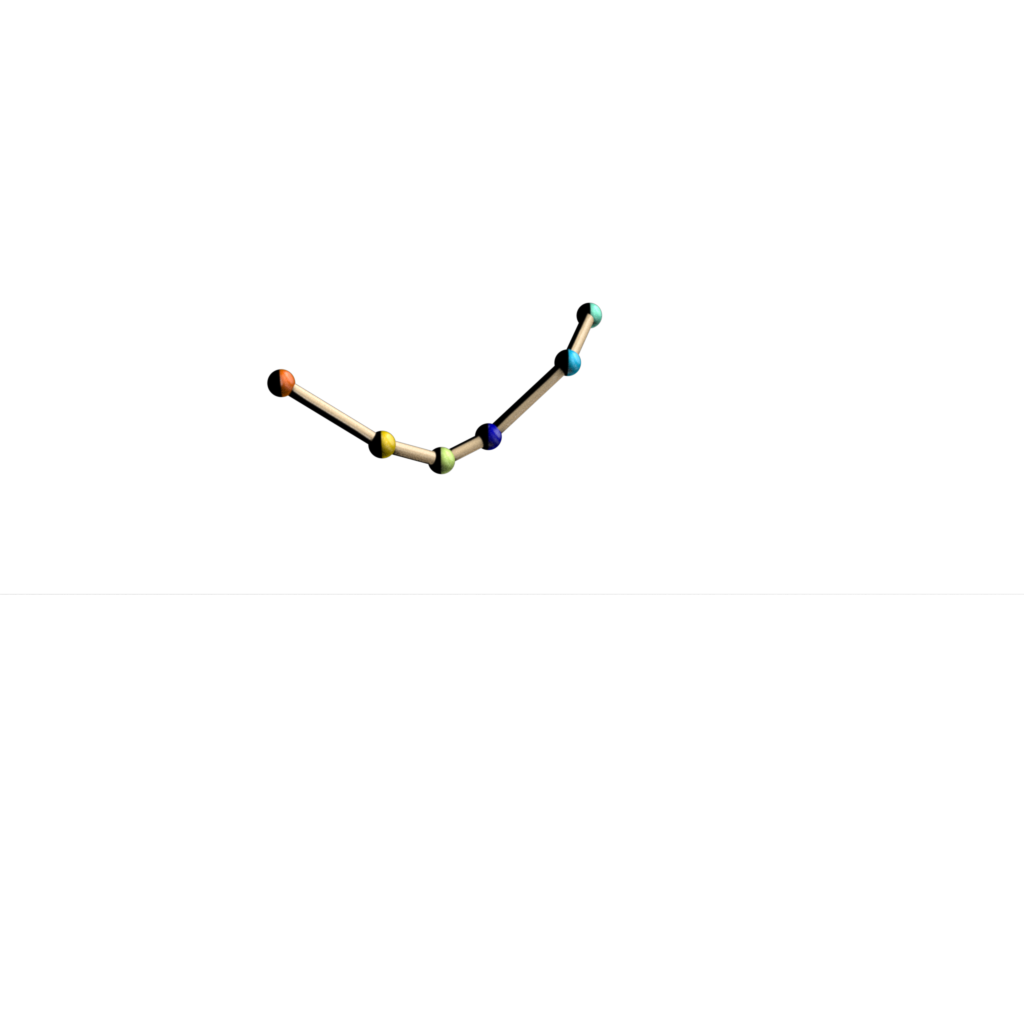}{0.15}{8cm 12cm 10cm 6cm}
        \robotvis{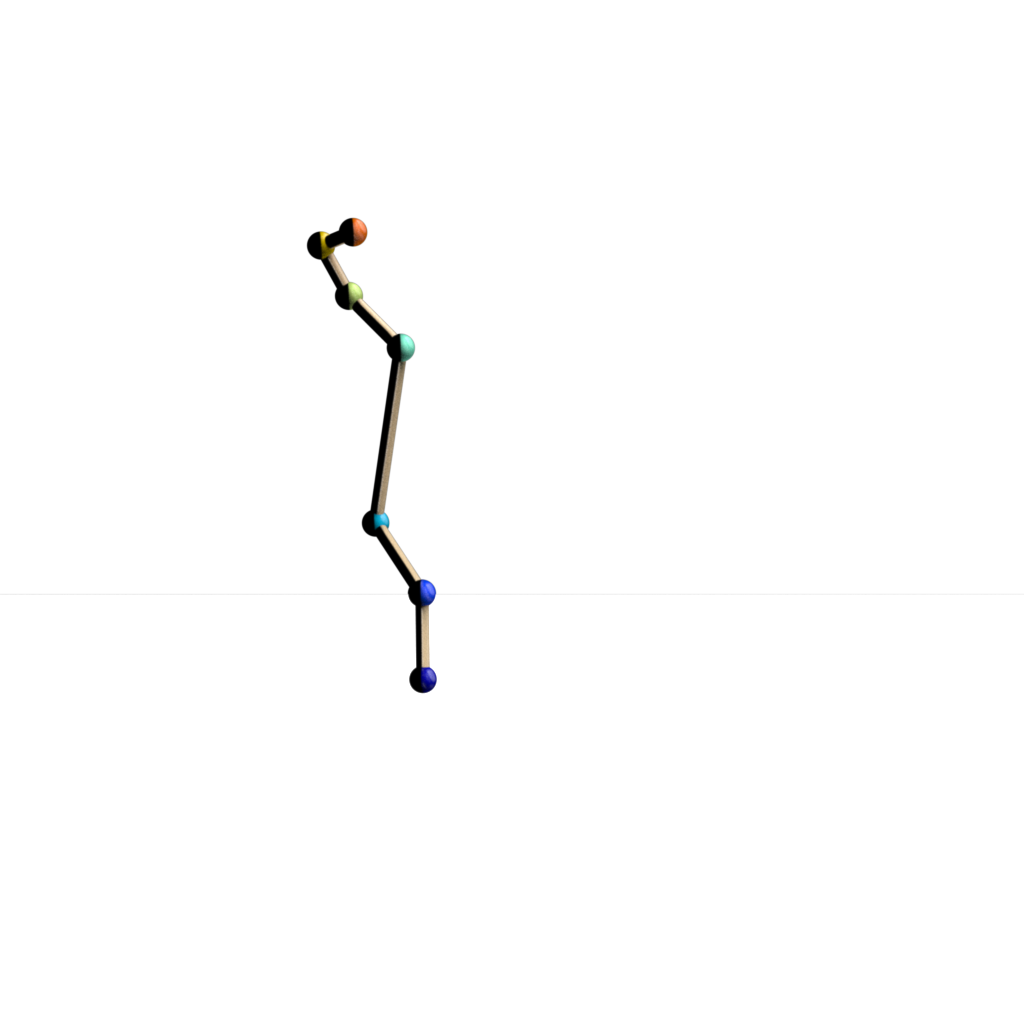}{0.15}{5cm 8cm 10cm 8cm}
        \robotvis{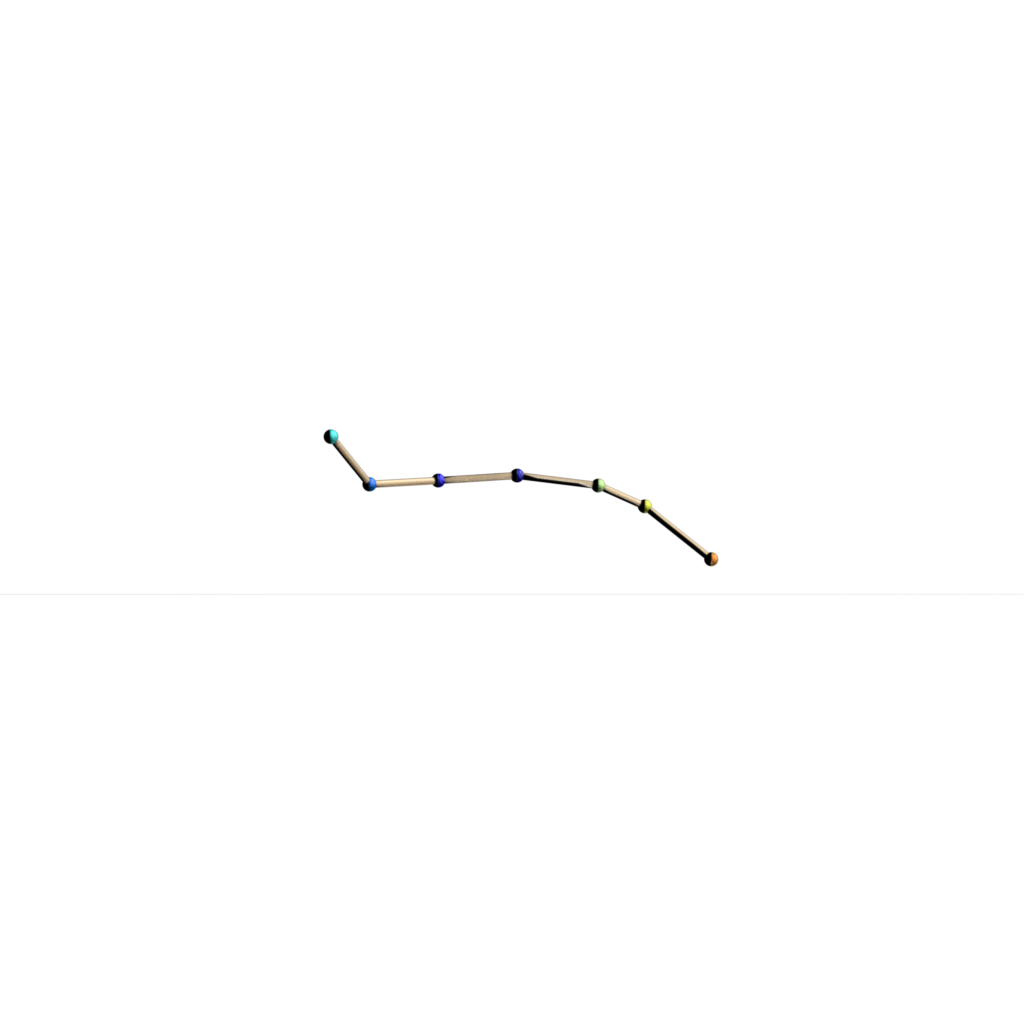}{0.15}{8cm 12cm 10cm 12cm}
        \robotvis{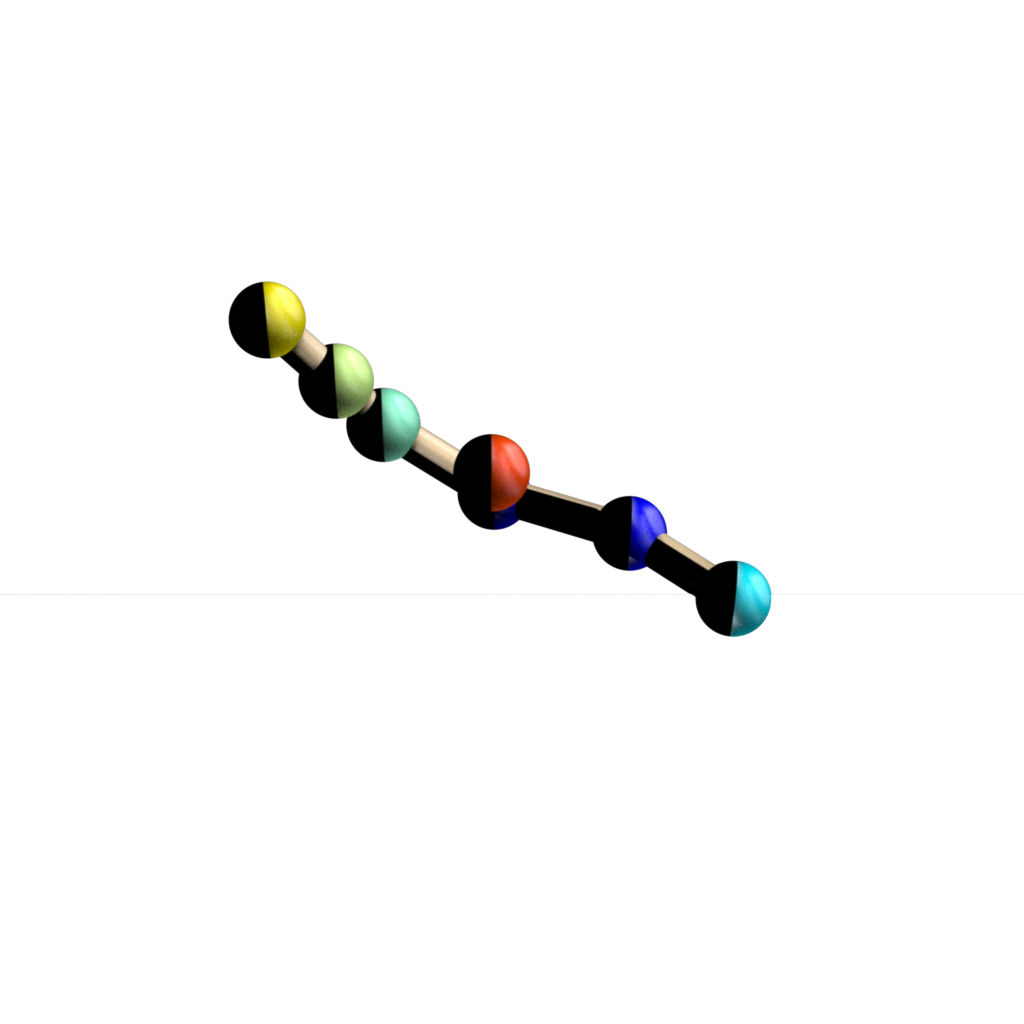}{0.15}{3cm 3cm 3cm 3cm}
         \robotvis{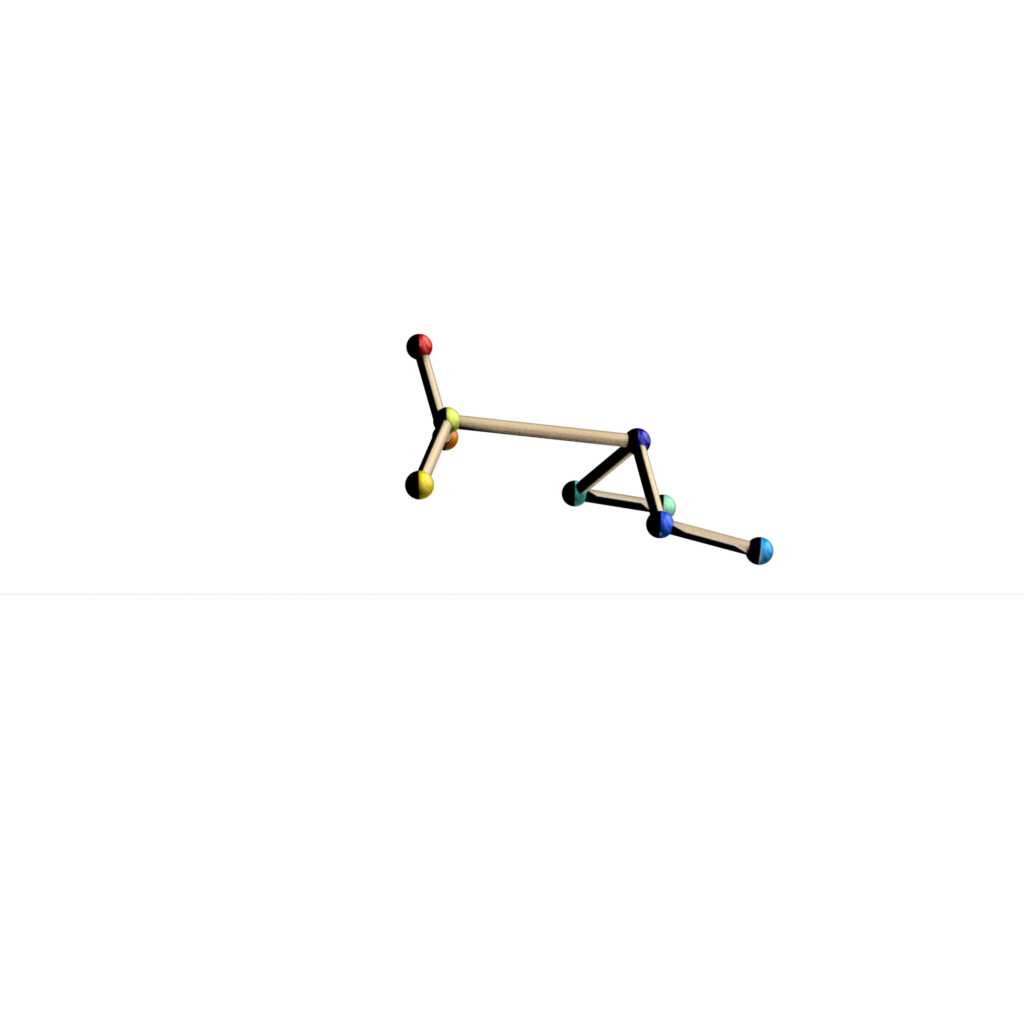}{0.15}{6cm 8cm 6cm 8cm}
        (a) Input Point Clouds &
        (b) Part Segmentation &
        (c) Topology &
        (d) Flow &
        (e) Part Segmentation &
        (f) Topology &
        (g) Flow \\
    \end{tabular}
    }
\captionof{figure}{\textbf{Qualitative Results on \robotD Dataset (1/2).} Given the input point cloud sequence (shown in (a)), we show the part segmentation,  part connectivity, and implied flow using our inferred articulated model (in (b, c, d)) and the ground truth articulated model (in (e, f, g))} 
\label{fig:supp-robots-1}
\end{table*}

\begin{table*}[ht]
    \centering
    \footnotesize
    \resizebox{\linewidth}{!}{
    \setlength{\tabcolsep}{0.2em} %
    \begin{tabular}{cccc|ccc}
        \cellcolor{lgray}{Input} & 
        \multicolumn{3}{c}{\cellcolor{lred}{Predictions}} &
        \multicolumn{3}{c}{\cellcolor{lgreen}{Ground Truth}} \\
         \robotvis{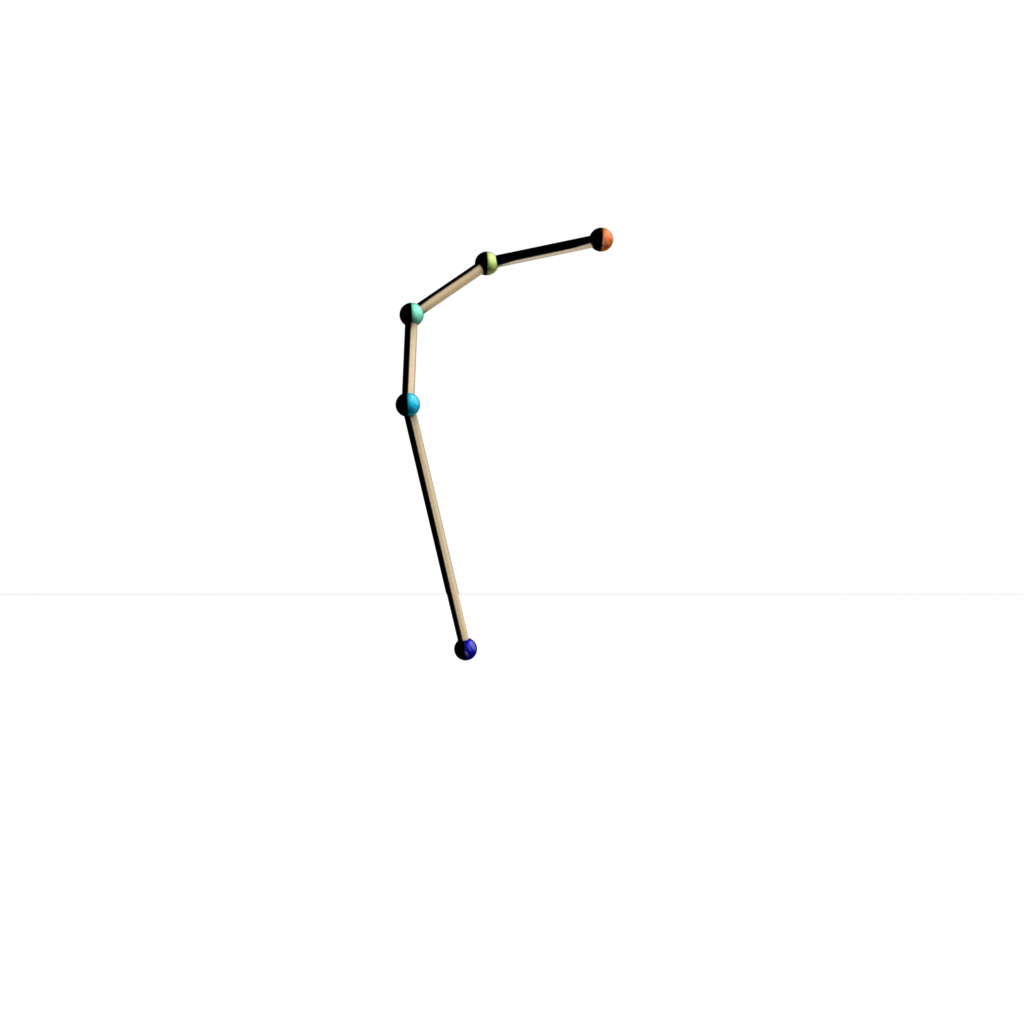}{0.15}{8cm 6cm 10cm 6cm}
         \robotvis{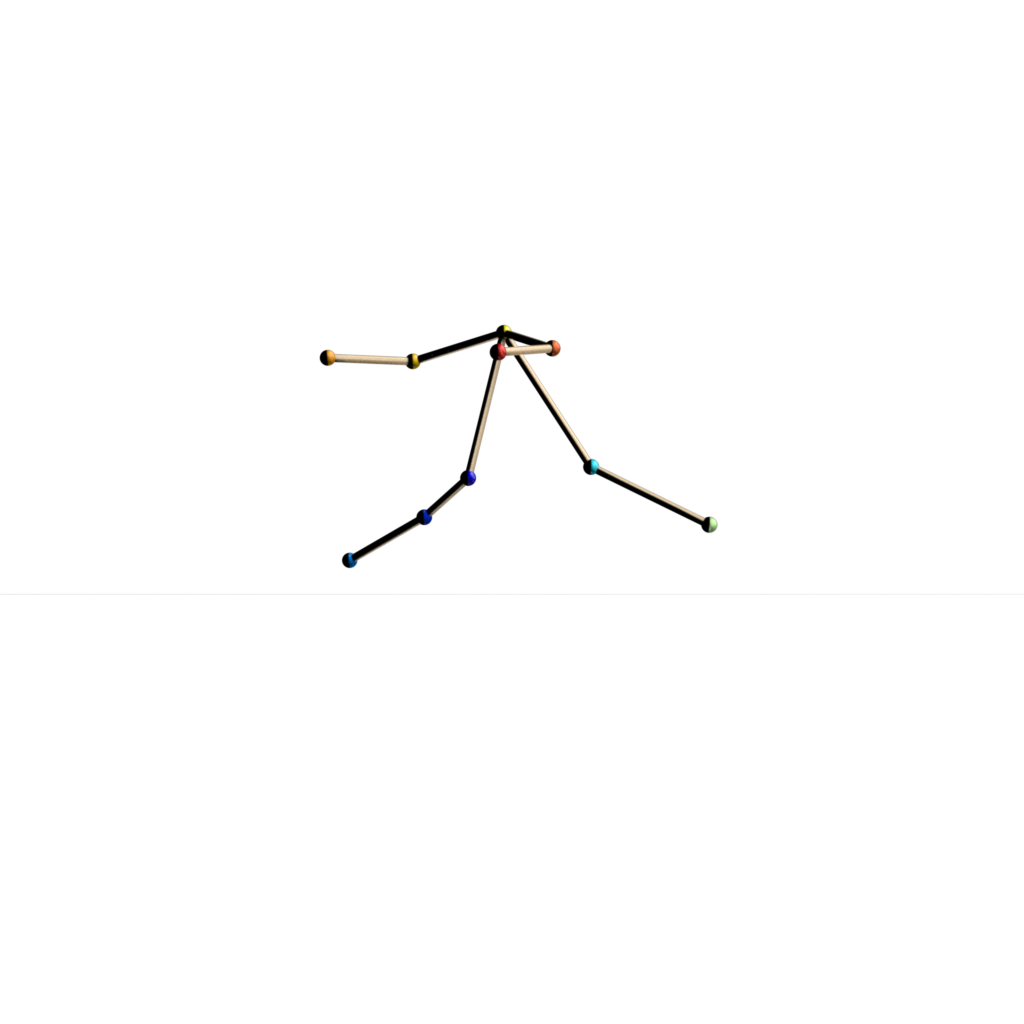}{0.15}{6cm 8cm 8cm 8cm}
         \robotvis{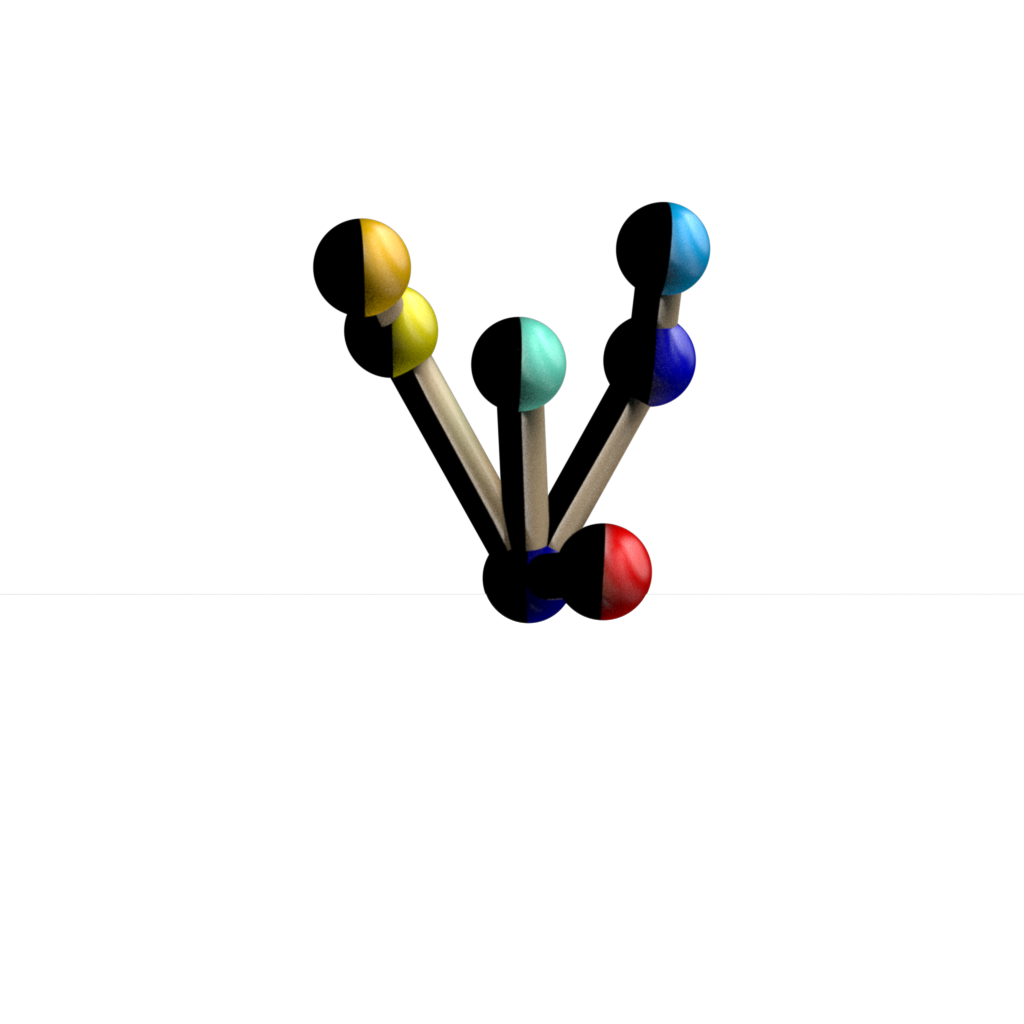}{0.15}{3cm 3cm 3cm 3cm} 
        (a) Input Point Clouds &
        (b) Part Segmentation &
        (c) Topology &
        (d) Flow &
        (e) Part Segmentation &
        (f) Topology &
        (g) Flow \\
    \end{tabular}
    }
\captionof{figure}{\textbf{Qualitative Results on \robotD Dataset (2/2).}} 
\label{fig:supp-robots-2}
\end{table*}

\begin{table*}[ht]
    \begin{subfigure}{.48\linewidth}
    \centering
    \small
    \resizebox{\linewidth}{!}{
    \setlength{\tabcolsep}{0.1em} %
    \begin{tabular}{cccc}
        \reanimatevis{bolt}{0.3}{8cm 10cm 10cm 7cm}{0}
        \reanimatevis{allegro}{0.3}{3cm 3cm 3cm 3cm}{0}
        (a) Input & 
        (b) Pred & 
        (d) GT 
    \end{tabular}
    }
    \end{subfigure}
    \hfill\vline\hfill
    \begin{subfigure}{.48\linewidth}
    \centering
    \small
    \resizebox{\linewidth}{!}{
    \setlength{\tabcolsep}{0.1em} %
    \begin{tabular}{cccc}
        \reanimatevis{barrett}{0.3}{5cm 5cm 5cm 5cm}{1}
        \reanimatevis{a1}{0.3}{8.0cm 11cm 9.0cm 6.0cm}{1}
        (a) Input & 
        (b) Pred & 
        (d) GT 
    \end{tabular}
    }
    \end{subfigure}
\captionof{figure}{\textbf{Reanimation Results on the \robotD Dataset (1/2).} Given new locations for a sparse set of points on the object(shown in (a)), our method (shown in (b)) is able to generate a reasonable reanimation to match the specified points.}
\label{fig:supp-animation-1}
\end{table*}

\begin{table*}[ht]
    \begin{minipage}{.48\linewidth}
    \centering
    \small
    \resizebox{\linewidth}{!}{
    \setlength{\tabcolsep}{0.1em} %
    \begin{tabular}{cccc}
        \reanimatevis{cassie}{0.3}{10cm 8cm 11cm 10cm}{0}
        \reanimatevis{nao}{0.3}{6cm 6cm 6cm 6cm}{2}
        \reanimatevis{panda}{0.3}{7cm 8cm 11cm 10cm}{2}
        \reanimatevis{spot}{0.3}{7cm 8cm 11cm 10cm}{0}
        (a) Input & 
        (b) Pred & 
        (d) GT 
    \end{tabular}
    }
    \end{minipage}
    \hfill\vline\hfill
    \begin{minipage}{.48\linewidth}
    \centering
    \small
    \resizebox{\linewidth}{!}{
    \setlength{\tabcolsep}{0.1em} %
    \begin{tabular}{cccc}
        \reanimatevis{icub}{0.3}{9.75cm 8cm 10cm 7cm}{1}
        \reanimatevis{kinova}{0.3}{8cm 6cm 10cm 6cm}{1}
        \reanimatevis{reachy}{0.3}{8cm 12cm 10cm 12cm}{1}
        \reanimatevis{ur5}{0.3}{5cm 8cm 10cm 8cm}{0}
        (a) Input & 
        (b) Pred & 
        (d) GT 
    \end{tabular}
    }
    \end{minipage}
\captionof{figure}{\textbf{Robot reanimation Results on the \robotD Dataset (2/2).}}
\label{fig:supp-animation-2}
\end{table*}

\begin{table*}[ht]
    \begin{subfigure}{.48\linewidth}
    \centering
    \footnotesize
    \resizebox{\linewidth}{!}{
    \setlength{\tabcolsep}{0.1em} %
    \begin{tabular}{ccc}
        \sapienvis{91}{2cm 6cm 7cm 3cm} \\
        \multicolumn{3}{c}{Faucet} \\
        \sapienvis{107}{2cm 6cm 7cm 3cm} \\
        \multicolumn{3}{c}{Laptop} \\
        \sapienvis{176}{2cm 6cm 7cm 3cm} \\
        \multicolumn{3}{c}{Knife} \\
        \sapienvis{242}{2cm 6cm 7cm 3cm} \\
        \multicolumn{3}{c}{Monitor} \\
        \sapienvis{287}{2cm 6cm 7cm 3cm} \\
        \multicolumn{3}{c}{Window} \\
        \sapienvis{238}{2cm 6cm 7cm 3cm} \\
        \multicolumn{3}{c}{Toilet} \\
        (a) Input & (b) Predicted Model & (c) Ground Truth
    \end{tabular}
    }
    \end{subfigure}
    \hfill\vline\hfill
    \begin{subfigure}{.48\linewidth}
    \centering
    \footnotesize
    \resizebox{\linewidth}{!}{
    \setlength{\tabcolsep}{0.1em} %
    \begin{tabular}{ccc}
        \sapienvis{183}{2cm 6cm 7cm 3cm} \\
        \multicolumn{3}{c}{Stapler} \\
        \sapienvis{201}{2cm 6cm 7cm 3cm} \\
        \multicolumn{3}{c}{Lighter} \\
        \sapienvis{231}{2cm 6cm 7cm 3cm} \\
        \multicolumn{3}{c}{Trash Can} \\
        \sapienvis{512}{2cm 6cm 7cm 3cm} \\
        \multicolumn{3}{c}{Washing Machine} \\
        \sapienvis{712}{2cm 6cm 7cm 3cm} \\
        \multicolumn{3}{c}{Kettle} \\
        \sapienvis{515}{2cm 6cm 7cm 3cm} \\
        \multicolumn{3}{c}{Pliers} \\
        (a) Input & (b) Predicted Model & (c) Ground Truth
    \end{tabular}
    }
    \end{subfigure}

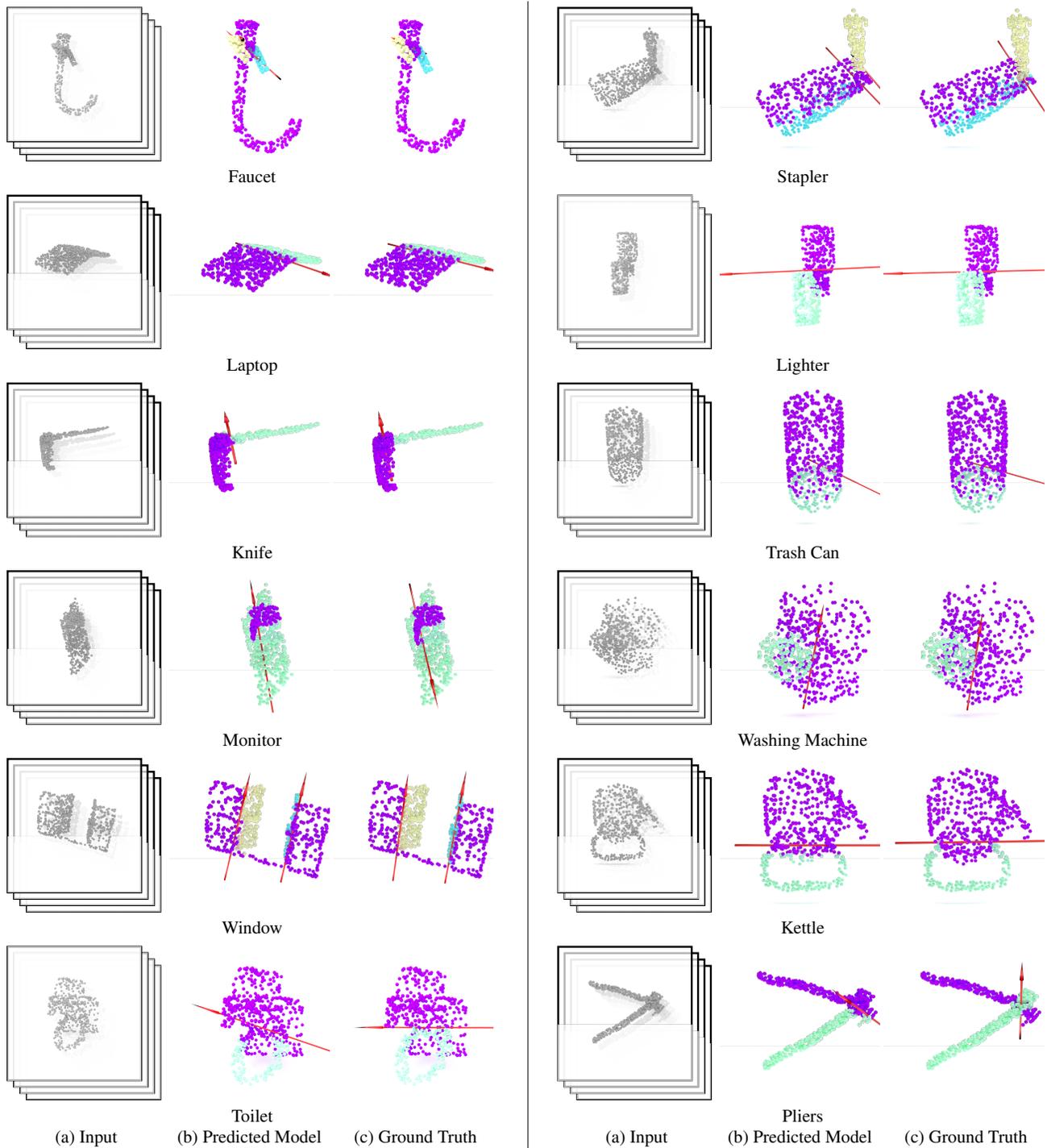
\captionof{figure}{\textbf{Qualitative results of common categories on \sapiensD dataset from~\cite{huang2021multibodysync}.} We visualize the predicted and ground truth articulated models. Different parts are in different colors, and we also show the screw parameters (in red) for the inferred joints. We use the provided flow estimation model~\cite{huang2021multibodysync}. Last row show some inaccurate results mostly caused by the noise in flow estimation.}
\label{fig:supp-sapien}
\end{table*}

\section{\robotD Dataset}
\label{sec:supp-dataset}
The \robotD dataset consists of 18 robots, including 6 different robot type: arms, bipeds, hands, mobile manipulators, humanoids, and quadrupeds. The robots, train-validation-test split are shown in~\cref{tab:robot-type} and ~\cref{tab:robot-split}. For each category, the points cloud sequence contain 10 frames with 4096 points independently sampled in each frame. Visualization of different categories are shown in~\cref{fig:supp-robot-dataset}.

\section{Additional Results}
\label{sec:supp-results}

\paragraph{Beyond 1DOF joints.} While our focus is on everyday objects, many of which have a piece-wise rigid structure with 1DOF joints, our framework is general and can tackle other joint types by modifying the {\it project} and {\it final fitting} steps. As a concrete example, spherical joints, which are a better model human and animals, can be tackled by a) replacing $E_\text{1-DOF}$ with $E_\text{3-DOF}$\footnote{
$E_\text{3-DOF}$ measures how well the child part motion (relative to the parent) is explained by rotation around a fixed center.} 
in $E_\text{project}$ in Eq. (11), and b) optimizing over spherical joint parameters (\vs screw params) during final fitting. We show results on a 10 time-step human point cloud sequence from D-FAUST dataset~\cite{bogo2017dynamic} in \cref{fig:supp-human}. This demonstrates that our framework can tackle objects with more general joints.

\paragraph{Per-category performance on \sapiensD dataset.} We report per-category (20 category in total) performance including both flow error and Multi-scan RI on \sapiensD daatset in \cref{tab:supp-sapien-cat}.

\paragraph{Comparison against \mbs and \wim on \robotD dataset.} We provide additional comparison results on remaining test set categories besides those shown in Fig. 5 in the main paper. We compare our method on all \robotD test set robot categories against \mbs~\cite{huang2021multibodysync} and \wim~\cite{noguchi2022watch}, the qualitative comparison is shown in \cref{fig:supp-comparison}. As can be seen \mbs severely suffer from missing parts and and \wim suffer from incorrect topology given it only takes spatial closeness into account when constructing the topology.  

\paragraph{Qualitative Results Visualization on \robotD dataset.} We provide additional qualitative visualization results in \cref{fig:supp-robots-1} and \cref{fig:supp-robots-2} on robot categories of validation set and remaining test set besides those shown in Fig. 4 in the main paper. We visualize part segmentation, topology and implied flow against ground-truth in each column. It can be seen our method work well for all robots with arbitrary topologies and geometries. 

\paragraph{Reanimation Results on \robotD dataset.} We provide additional reanimation results in \cref{fig:supp-animation-1} and \cref{fig:supp-animation-2} on robot categories of validation set and remaining test set besides those shown in Fig. 7 in the main paper. As shown in the figure, we observe the results looks reasonable and match to the sparse guidance input. This demonstrates our animatable models's rearticulation ability.

\paragraph{Qualitative Results Visualization on \sapiensD dataset.} We provided common Sapien categories prediction results in \cref{fig:supp-sapien} besides those shown in Fig. 6 in the main paper. As shown in the figure, our method works well on arbitrary daily articulated objects with different geometries and number of parts. We also show some inaccurate results in the last row of \cref{fig:supp-sapien}. We found out those inaccuracy are mostly caused by the noisy flow estimation provided by~\cite{huang2021multibodysync}.

\end{document}